\documentclass[11pt]{article}

\usepackage[preprint]{acl}

\usepackage{times}
\usepackage{latexsym}
\usepackage[T1]{fontenc}
\usepackage[utf8]{inputenc}
\usepackage{microtype}
\usepackage{inconsolata}
\usepackage{graphicx}
\usepackage{xspace}
\usepackage{tabularx}
\usepackage{multirow}
\usepackage{graphicx}
\usepackage{amssymb}
\usepackage{pifont}
\usepackage{float}
\usepackage{enumitem}
\usepackage{amsmath}
\usepackage{times}
\usepackage{latexsym}
\usepackage{booktabs}
\usepackage{arydshln}
\usepackage{algorithmicx}
\usepackage{algpseudocode}
\usepackage{algorithm}
\usepackage{placeins}
\usepackage{float}
\usepackage{cuted}
\usepackage[most]{tcolorbox}
\usepackage{colortbl}
\usepackage{tcolorbox}
\usepackage{xcolor}     
\usepackage{color}
\usepackage{colortbl,array}
\usepackage[rgb]{xcolor}

\newcommand{\ie}{\emph{i.e., }}
\newcommand{\eg}{\emph{e.g., }}

\title{AgentCVR: Active Multi-Agent Cross-Video Reasoning via Script-Simulated Reinforcement Learning}

\author{
 \textbf{Yilun Qiu\textsuperscript{1\thanks{Equal Contribution}}},
 \textbf{Jiahe Wang\textsuperscript{1,2\footnotemark[1]}},
\textbf{Cilin Yan\textsuperscript{1}},
\textbf{Jiayin Cai\textsuperscript{1}},
\textbf{Xiaolong Jiang\textsuperscript{1}},
\textbf{Yan Hu\textsuperscript{1}},
\textbf{Chun Yuan\textsuperscript{2}\thanks{Corresponding Author}}
\\
 \textsuperscript{1} Xiaohongshu Inc. \\
 \textsuperscript{2} Tsinghua Shenzhen International Graduate School, Tsinghua University \\
\small qiuyilun@u.nus.edu, wang-jh24@mails.tsinghua.edu.cn, clyanhh@gmail.com, caijy18@tsinghua.org.cn, \\
\small laige@xiaohongshu.com, yaoohu@gmail.com, yuanc@sz.tsinghua.edu.cn
}

\begin{document}
\maketitle
\begin{abstract}
Cross-Video Reasoning (CVR) has emerged as a critical frontier in multimodal intelligence, requiring models to retrieve, align, and aggregate evidence distributed across multiple videos.
Current Multimodal Large Language Models (MLLMs) often struggle with CVR, as simple single-pass strategies encode multiple videos into a shared compressed context, potentially obscuring rare but critical evidence.
In this paper, we propose \textbf{AgentCVR}, a multi-agent framework that treats CVR as an active evidence-acquisition task.
AgentCVR employs a Master Agent to iteratively coordinate specialized Visual and Audio Agents for targeted evidence extraction.
To ensure efficient training, we introduce Script-Simulated RL, which optimizes the agent’s policy with LLM-generated semantic scripts and a lightweight text-based simulator, bypassing costly multimodal inference during online exploration.
Experimental results on a comprehensive CVR benchmark show that AgentCVR outperforms single-pass baselines and achieves comparable performance to state-of-the-art closed-source systems, particularly in complex cross-video alignment and localization.
To ensure reproducibility, our code is available at \url{https://github.com/wang-jh24/AgentCVR}.
\end{abstract}

\section{Introduction}


With continuous efforts, Multimodal Large Language Models (MLLMs)~\cite{gemini,kimi25,qwen3vl} have significantly advanced artificial intelligence in complex vision-language tasks.
In the domain of video understanding~\cite{videosurvey2,videosurvey1}, these models have demonstrated strong capabilities on tasks such as question answering~\cite{flippedvqa,nextgqa} and temporal grounding~\cite{tgvid,numpro}.
However, most existing video understanding studies and benchmarks are limited to single-video analysis, and thus fail to adequately evaluate a model’s ability to reason across multiple videos. 
As real-world scenarios become more complex, processing isolated videos is no longer enough, driving a shift toward \emph{Cross-Video Reasoning} (CVR)~\cite{cvbench,crossvid}.


CVR requires models to answer queries whose evidence is distributed across multiple videos, often involving retrieval, alignment, comparison, and aggregation over temporally distant events.
This transition introduces an unprecedented challenge: 
critical evidence is often sparsely distributed within individual videos and temporally misaligned across them, requiring explicit cross-video comparison to derive meaningful insights~\cite{longvideobench,mvbench,videomme}.
Such properties make CVR particularly challenging for current MLLMs.

\begin{figure}[t]
    \centering
    \includegraphics[width=1 \linewidth]{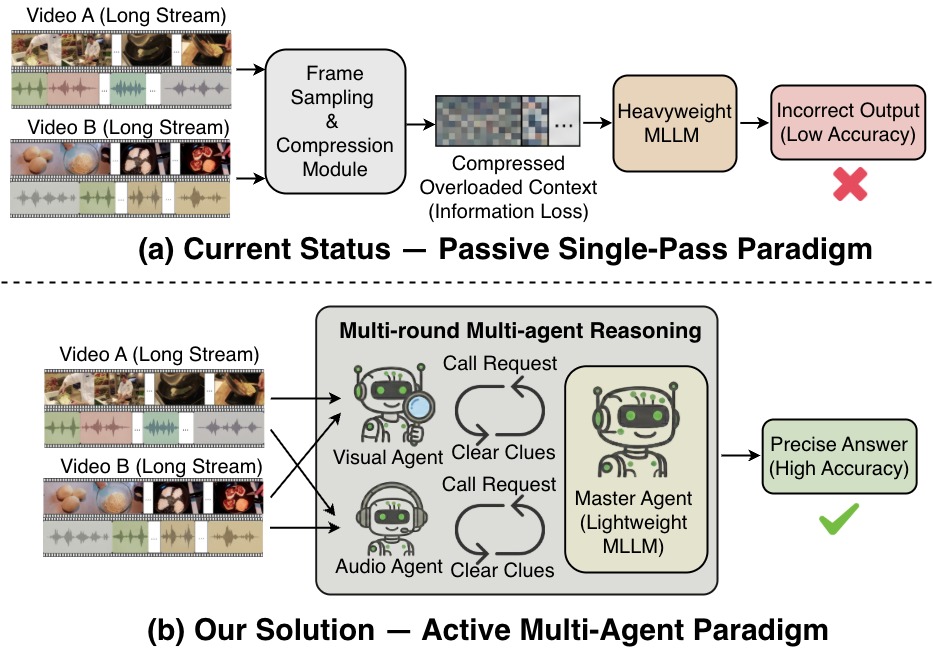}
    \vspace{-1.5em}
    \caption{Comparison between two formulations of Cross-Video Reasoning (CVR). (a) Current Status: passive single-pass paradigm. (b) Our Solution: active multi-agent paradigm.}
    \label{agentcvr_intro}
    \vspace{-1em}
\end{figure}


As illustrated in Figure~\ref{agentcvr_intro}(a), a common strategy for CVR is to encode all candidate videos into a shared compressed context and generate the answer in a single pass~\cite{internvideo2,timechat,longvideobench}.
While simple, this design compresses long videos into a fixed-size representation, which can obscure rare but critical cues and leave the model to reason from only partially grounded observations.
Consequently, models often exhibit weak evidence attribution and unreliable cross-video comparison, especially for temporal and comparative queries.


To address these limitations, we argue that CVR should be treated not only as a context-length problem but also as an evidence-acquisition problem.
Instead of reasoning over all videos in a single step, a framework should iteratively determine what evidence to inspect, which modality to query, and when sufficient support has been collected.
To this end, we propose \textbf{AgentCVR}, an active multi-agent framework for CVR.
As illustrated in Figure~\ref{agentcvr_intro}(b), a lightweight Master Agent~\cite{agentbench} iteratively coordinates specialized Visual and Audio Agents~\cite{hugginggpt,vipergpt,appagent} to extract relevant clues for high-level cross-video deduction.
This design keeps multimodal processing focused on targeted evidence and provides the reasoning module with more explicit, query-conditioned inputs.


To further boost AgentCVR's capability to make complex, long-horizon decisions for inter-agent coordination and evidence acquisition, we aim to employ Reinforcement Learning (RL) to optimize its multi-round policy.
However, naively applying online RL to CVR is prohibitively expensive, as repeatedly invoking video models incurs massive computational overhead in each step, exacerbated by the scarcity of annotated cross-video trajectories.
To address this, we introduce \emph{Script-Simulated RL}, an efficient training paradigm that replaces heavy raw-video interactions during online exploration with LLM-generated semantic scripts and a lightweight text-based simulator that approximates multi-round evidence acquisition.
This formulation preserves the core reasoning structure while drastically reducing training costs and the need for human-annotated trajectories.
The optimized policy is then directly transferred to the real inference pipeline, where the agent interacts with actual visual and audio agents over raw videos.


We evaluate AgentCVR on the CrossVid benchmark~\cite{crossvid}, a large-scale and comprehensive dataset specifically designed for CVR.
Experimental results demonstrate that our proposed framework outperforms single-pass baselines and achieves performance comparable to that of state-of-the-art closed-source systems, especially on tasks requiring precise evidence localization and fine-grained cross-video alignment.

Our main contributions are as follows:
\begin{itemize}[leftmargin=*, topsep=2pt, itemsep=0pt]
    \item We propose AgentCVR, a multi-agent framework for CVR that fundamentally shifts the paradigm from passive single-pass context compression to active, multi-round evidence acquisition.
    \item We introduce Script-Simulated RL, a paradigm for training AgentCVR with LLM-generated semantic scripts and a lightweight text-based simulator, avoiding costly multimodal inference during online exploration.
    \item Extensive experimental results demonstrate that AgentCVR achieves strong performance across comprehensive CVR tasks.
\end{itemize}

\section{Related Work}

We review related work along three dimensions in this section:
video understanding, multimodal agents, and reinforcement learning.

\vspace{0.5em}
\noindent
\textbf{Video Understanding.} 
The rapid evolution of MLLMs has significantly advanced video understanding~\cite{flamingo,videochatgpt,videollava,videomind}.
To efficiently process long videos, existing approaches typically adopt key-frame selection and temporal token compression~\cite{internvideo2,pllava,videollava}.
Meanwhile, benchmarks have progressed from short-clip QA~\cite{tvqa,activitynetqa,nextqa} to long-context evaluations~\cite{egoschema,longvideobench,lvbench}.
Beyond single-video settings, real-world applications increasingly require Cross-Video Reasoning, which integrates information across multiple videos~\cite{cvbench,youtuvl}.
CrossVid~\cite{crossvid} has recently emerged as a representative benchmark.
However, single-pass models struggle in CVR, as concatenating videos often obscures sparse and temporally dispersed evidence.
Our work focuses on CVR and explores an agentic alternative based on multi-round evidence acquisition.

\vspace{0.5em}
\noindent
\textbf{Multimodal Agents.}
LLM-based agents enable iterative decision making and tool use by treating the LLM as a controller for multi-step reasoning and interactions~\cite{react,reflexion,toolformer,webarena}.
In multimodal contexts, prior work explores modular tool use to facilitate perception and reasoning~\cite{vipergpt,visualprogramming,dart} under the paradigm of \emph{active perception}~\cite{activevision}, where sensory input is dynamically selected based on evolving hypotheses.
While active perception has shown promise in single-video scenarios~\cite{videoagent,traveler}, its application to CVR remains underexplored.
To the best of our knowledge, our work fills this gap by proposing a novel strategy specifically tailored for CVR.

\vspace{0.5em}
\noindent
\textbf{Reinforcement Learning.}
Reinforcement learning (RL) has significantly advanced LLM reasoning~\cite{ppo,deepseekmath,deepseekr1}.
However, optimizing multimodal agents via naive online exploration incurs prohibitive costs due to repeated heavyweight vision-language model inferences.
To mitigate this, policies are often trained within surrogate simulators before real-world deployment~\cite{textworld,alfworld,webgpt,voyager,tarl,rt2}.
Following this principle, we avoid expensive training by learning CVR tool-use policies in a lightweight script-based simulator, while deploying real visual and audio tools only during inference.
\section{Preliminary}

In this section, we formally define the CVR problem and introduce the Partially Observable Markov Decision Process (POMDP), which serves as the foundation for our proposed framework.

\vspace{0.5em}
\noindent
\textbf{Problem Formulation.}
We formulate Cross-Video Reasoning (CVR) as a generalized multimodal instruction-following task that requires retrieving, aligning, and aggregating information across multiple independent video streams to produce a final answer.
Formally, the system is provided with a user query $q$ and a candidate video set $\mathcal{V} = \{V_1, V_2, \dots, V_N\}$, where $N \ge 2$.
Each video $V_i \in \mathcal{V}$ represents an independent, unaligned spatiotemporal stream comprising heterogeneous multimodal signals (e.g., visual frames and audio tracks).
The objective is to learn an intelligent system $\mathcal{G}$ that outputs a response $\hat{a}$ matching the ground-truth correct answer $a$, denoted as:
\begin{equation}
    \hat{a}=\mathcal{G}(q, \mathcal{V}).
\end{equation}
Unlike single-video tasks with locally confined evidence, CVR involves sparse and temporally unaligned cues across $\mathcal{V}$.
Thus, deriving $\hat{a}$ requires explicit inter-video reasoning to align and aggregate fragmented multimodal evidence.

\vspace{0.5em}
\noindent
\textbf{Partially Observable Markov Decision Process.}
To systematically solve the CVR problem formulated above, we model the active multi-round evidence acquisition process as an MLLM-based POMDP defined by the tuple $(\mathcal{S}, \mathcal{A}, \mathcal{O}, \mathcal{T}, \mathcal{R})$.
In this formulation, the Master Agent acts as the central controller interacting with a multimodal environment. The components of the POMDP tuple are defined as follows:
\begin{itemize}[leftmargin=*, topsep=2pt, itemsep=0pt]
    \item \textbf{State ($\mathcal{S}$):} At step $t$, the state is defined as $s_t=(q, \mathcal{H}_t)$, where $q$ is the user query and $\mathcal{H}_t=(a_1, o_1, \dots, a_{t-1}, o_{t-1})$ encapsulates the entire interaction history of past actions and observations up to the current step.
    \item \textbf{Action ($\mathcal{A}$):} The agent samples an action $a_t \in \mathcal{A}$ from its parameterized policy $\pi_{\theta}(a_t|s_t)$, including invoking modality-specific tools (e.g., visual or audio) for evidence retrieval or executing a termination action to generate the final answer.
    \item \textbf{Observation ($\mathcal{O}$):} After executing $a_t$, the environment returns an observation $o_t \in \mathcal{O}$ (\eg textual clues from a video segment), and the state is updated accordingly.
    \item \textbf{Transition ($\mathcal{T}$):} The transition is deterministic, updating the state by appending the new pair $(a_t, o_t)$ to the history.
    \item \textbf{Reward ($\mathcal{R}$):} After the trajectory terminates at step $T$, an episodic reward $R \in \mathcal{R}$ is assigned to evaluate the complete interaction trajectory.
\end{itemize}
The objective of this MLLM-based POMDP is to optimize the policy $\pi$ that maximizes the expected cumulative reward:
\vspace{-0.2em}
\begin{equation}
    \max_{\theta} \ \mathbb{E}_{\pi_\theta}[R].
\end{equation}
\section{Methodology}

\begin{figure*}[t]
    \centering
    \includegraphics[width=\linewidth]{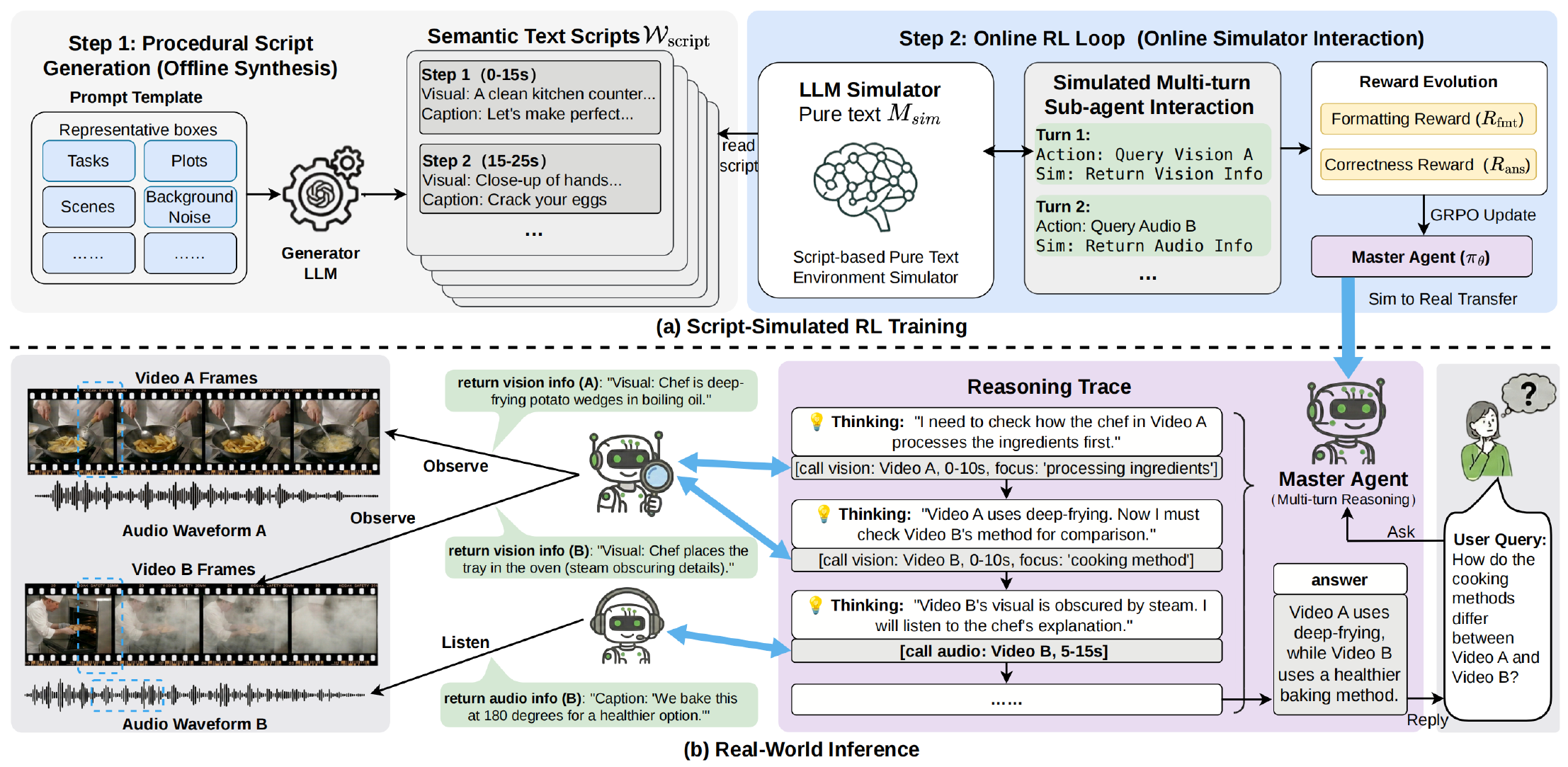}
    \vspace{-1.5em}
    \caption{
    \textbf{Overview of AgentCVR.} 
    (a) \emph{Script-Simulated RL Training:} An LLM generator produces semantic scripts ($\mathcal{W}_{\mathrm{script}}$), and a text-based simulator ($M_{\mathrm{sim}}$) provides feedback for policy optimization of the Master Agent ($\pi_\theta$) with GRPO.
    (b) \emph{Real-World Inference:} At inference time, the trained Master Agent interacts with visual and audio agents over raw videos to gather localized multimodal evidence for CVR.
    }
    \label{agentcvr_overview}
    \vspace{-1em}
\end{figure*}

In this section, we introduce our proposed \textbf{AgentCVR} framework, followed by its two main phases: \emph{Script-Simulated RL Training} and \emph{Real-World Inference}, as illustrated in Figure~\ref{agentcvr_overview}.

\subsection{AgentCVR}

AgentCVR adopts a multi-round evidence acquisition framework for CVR.
At its core, a lightweight Master Agent serves as the central controller.
At each step $t$, it reasons over the query and the accumulated interaction history $\mathcal{H}_t$ to dynamically determine which modality-specific agent to invoke, which temporal segment to inspect, and when to terminate the process and generate the final answer.

Formally, we decompose the action space into three subspaces: $\mathcal{A} = \mathcal{A}_{\mathrm{vis}} \cup \mathcal{A}_{\mathrm{aud}} \cup \mathcal{A}_{\mathrm{ans}}$.
\begin{itemize}[leftmargin=*, topsep=2pt, itemsep=0pt]
    \item \textbf{Visual Query ($a_{\mathrm{vis}} \in \mathcal{A}_{\mathrm{vis}}$):}
    A visual action is parameterized as $(vid, \tau_{\mathrm{start}}, \tau_{\mathrm{end}}, \mathcal{P}_{\mathrm{focus}})$, where the Master Agent specifies the target video identifier $vid$, the temporal interval $[\tau_{\mathrm{start}}, \tau_{\mathrm{end}}]$, and an optional focus prompt $\mathcal{P}_{\mathrm{focus}}$.
    The visual agent then processes this specific segment and returns localized textual observations.
    \item \textbf{Audio Query ($a_{\mathrm{aud}} \in \mathcal{A}_{\mathrm{aud}}$):}
    Recognizing that auditory signals provide complementary cues for cross-video alignment, an audio action is parameterized as $(vid, \tau_{\mathrm{start}}, \tau_{\mathrm{end}})$, where the Master Agent triggers the audio agent to retrieve speech or sound descriptions from the selected temporal interval $[\tau_{\mathrm{start}}, \tau_{\mathrm{end}}]$.
    \item \textbf{Answer Action ($a_{\mathrm{ans}} \in \mathcal{A}_{\mathrm{ans}}$):}
    Once the Master Agent determines that sufficient evidence has been collected across the videos, it executes the answer action to terminate the process and generate the final answer $\hat{a}$.
\end{itemize}

At each round, the Master Agent updates its decision based on the accumulated history and newly returned observations. 
In this way, \textsc{AgentCVR} separates evidence gathering from final reasoning: modality-specific agents are responsible for retrieving localized observations, while the Master Agent performs cross-video comparison and deduction over the collected evidence. 
This multi-round design is particularly useful for CVR, where the relevant cues may be sparse, temporally distant, or distributed across different videos.

\subsection{Script-Simulated RL Training}

To empower AgentCVR to make sophisticated decisions for inter-agent coordination and evidence acquisition, we employ Reinforcement Learning (RL) to optimize the Master Agent.
To make RL optimization practical for CVR, we introduce a lightweight script-simulated surrogate environment inspired by text-based environments.
The overall pipeline, including offline script construction and online RL training, is summarized in Algorithm~\ref{alg_training}.

\vspace{0.5em}
\noindent
\textbf{Synthetic Script Construction.}
We construct a script-simulated surrogate environment by generating synthetic semantic scripts $\mathcal{W}_{\mathrm{script}}$ with a strong LLM conditioned on structured schemas and task templates.
Each script defines (i) a set of videos, (ii) temporally grounded events and attributes, and (iii) cross-video relations required to answer queries such as comparisons and temporal constraints.
By sampling diverse templates and attribute configurations, we obtain a collection of lightweight textual environments that cover common CVR reasoning patterns.
We provide the script schema and representative examples of $\mathcal{W}_{\mathrm{script}}$ in Appendix~\ref{apd_prompt}.

\vspace{0.5em}
\noindent
\textbf{Online Text-Based Simulator.}
During online RL, we remove raw-video inference from the training loop and instead employ a lightweight language model as the surrogate simulator $M_{\mathrm{sim}}$.
Given a sub-agent invocation action $a_t$ (\eg querying a video segment) and the corresponding script slice, the simulator returns an observation:
\vspace{-0.2em}
\begin{equation}
o_t = M_{\mathrm{sim}}\!\left(a_t,\; \mathcal{W}_{\mathrm{script}}\!\left(vid,\, [\tau_{\mathrm{start}}, \tau_{\mathrm{end}}]\right)\right).
\end{equation}
Concretely, $M_{\mathrm{sim}}$ performs semantic lookup and composition over scripted events to generate natural-language responses that approximate feedback from visual and audio agents.
We further constrain the simulator via system-level instructions to ensure behavioral consistency.
In particular, when no relevant event occurs within the queried interval, it returns an explicit ``no evidence'' response instead of hallucinating content.
This design provides a stable and reliable environment feedback for policy optimization while reducing interaction cost, achieving up to $\sim$5$\times$ lower latency in our setup.
Moreover, representing videos as structured scripts facilitates scalable environment diversification through LLM-based augmentation without requiring additional video collection.

\vspace{0.5em}
\noindent
\textbf{RL Reward.}
To align the agent with task objectives and inter-agent interaction constraints, we define a trajectory-level reward consisting of an answer correctness term and an interaction formatting term~\cite{let}:
\begin{itemize}[leftmargin=*, topsep=2pt, itemsep=0pt]
    \item \textbf{Correctness Reward:} 
    The primary reward signal is the correctness of the final answer, implemented as a sparse reward:
    \vspace{-0.2em}
    \begin{equation}
        R_{\mathrm{ans}} =
        \begin{cases}
        1, & \text{if } \hat{a} = a,\\
        0, & \text{otherwise}.
        \end{cases}
    \end{equation}
    
    \item \textbf{Formatting Reward:}
    Since our framework relies on structured sub-agent interactions, invalid invocation actions may disrupt execution.
    We therefore add a lightweight auxiliary reward to encourage valid trajectories.
    Specifically, we set $R_{\mathrm{fmt}}=0.1$ if the invocation satisfy predefined format constraints, and $R_{\mathrm{fmt}}=0$ otherwise.
\end{itemize}
The total reward is defined as:
\vspace{-0.2em}
\begin{equation}
    R_{\mathrm{total}} = R_{\mathrm{ans}} + R_{\mathrm{fmt}}.
\end{equation}

\vspace{0.5em}
\noindent
\textbf{Policy Update.}
To optimize the Master Agent, we employ Group Relative Policy Optimization (GRPO)~\cite{deepseekmath}, an efficient RL algorithm.
For each query $q$, we sample a group of $G$ trajectories $\{ \mathcal{H}^{(1)}, \dots, \mathcal{H}^{(G)} \}$ from the current policy $\pi_{\text{old}}$ and optimize the following objective to obtain the updated policy $\pi_\theta$:
\vspace{-0.2em}
\begin{equation}
\small
\begin{gathered}
    \mathcal{J}_{\text{GRPO}}(\theta) = \mathbb{E}_{\mathcal{H}^{(i)} \sim \pi_{\theta_{\text{old}}}} \Bigg[ \frac{1}{G} \sum_{i=1}^G \min \bigg( r_i(\theta) A_i, \\
    \text{clip}\left(r_i(\theta), 1-\epsilon, 1+\epsilon\right) A_i \bigg) \Bigg] - \beta \mathbb{D}_{\text{KL}}(\pi_\theta || \pi_{\text{ref}}),
\end{gathered}
\end{equation}
where $r_i(\theta)=\frac{\pi_\theta(\mathcal{H}^{(i)})}{\pi_{\theta_{\mathrm{old}}}(\mathcal{H}^{(i)})}$ is the trajectory-level importance sampling ratio computed over the full sequence of actions and outputs in $\mathcal{H}^{(i)}$, $\epsilon$ and $\beta$ are hyperparameters, and $\mathbb{D}_{\text{KL}}$ is a KL term that regularizes the updated policy toward a reference policy.
The advantage $A_i$ is derived by normalizing the reward relative to the generated group:
\vspace{-0.2em}
\begin{equation}
    \hat{A}_i = \frac{R^{(i)}_{\mathrm{total}} - \mu_R}{\sigma_R},
\end{equation}
where $R^{(i)}_{\mathrm{total}}$ denotes the reward of trajectory $\mathcal{H}^{(i)}$, $\mu_R$ is the group mean reward, and $\sigma_R$ is the standard deviation of the rewards within the group.

\subsection{Real-World Inference}

A primary advantage of our Script-Simulated RL Training is its seamless Sim-to-Real transferability.
At inference time, the optimized policy $\pi_{\theta}$ is deployed directly to interact with the physical environment.

As outlined in Algorithm~\ref{alg_inference}, for each user query $q$ and a set of raw, uncompressed candidate videos $\mathcal{V}$, the Master Agent initiates an iterative reasoning loop.
At each step $t$, the agent samples an action $a_t \sim \pi_{\theta}(\cdot \mid s_t)$ based on its current state.
Rather than relying on textual simulation, the agent executes $a_t$ by invoking actual perception models such as a visual agent $T_{\mathrm{vis}}$ or an audio agent $T_{\mathrm{aud}}$ on the raw videos $\mathcal{V}$ to obtain the real multimodal observation $o_t$.
The interaction history is subsequently updated with the newly acquired $(a_t, o_t)$ pair, and the state is advanced accordingly.
This evidence-gathering process continues until the Master Agent determines that sufficient cross-video support has been collected, at which point it terminates the loop and outputs the final prediction $\hat{a}$.

\section{Experiments}

\subsection{Experimental Setup}

\noindent
\textbf{Dataset.} 
We evaluate the performance of our AgentCVR on CrossVid~\cite{crossvid}, a comprehensive large-scale benchmark designed for CVR.
Unlike conventional video QA datasets that focus on a single video per query, CrossVid adopts a setting where each query is associated with multiple videos.
CrossVid is organized along four dimensions: Comparative Analysis, Temporal Understanding, Multi-view Reasoning, and Free-form QA. 
It covers 10 tasks in total, with detailed descriptions provided in Appendix~\ref{apd_dataset}.

\definecolor{mygray1}{HTML}{8F8C8C}
\definecolor{mygray2}{HTML}{666565}

\begin{table*}[t]
  \caption{
  Performance comparison between baselines and our AgentCVR on the CrossVid benchmark. 
  The best results are highlighted in \textbf{bold}, and the second-best results are \underline{underlined} across open-source single-pass models, adapted single-video agents, and our proposed framework.
  The table reports accuracy (\%) across 10 tasks, along with dimension-level averages, including \textbf{C.Avg}, \textbf{T.Avg}, and \textbf{M.Avg}.
  \textbf{\texttt{O.Avg}} denotes the overall average.
  }
  \vspace{-0.5em}
  \label{main_table}
  \centering
  \renewcommand{\arraystretch}{1.2}
  
  \fontsize{8.3}{9}\selectfont
  
  \setlength{\tabcolsep}{1.5pt} 
  \newcolumntype{Y}{>{\centering\arraybackslash}X}
  
  \begin{tabularx}{\textwidth}{@{} l @{\hspace{6pt}} *{5}{Y} @{\hspace{6pt}} *{4}{Y} @{\hspace{6pt}} *{3}{Y} @{\hspace{6pt}} c @{\hspace{6pt}} Y @{}}
    \toprule
    \textbf{Task ($\rightarrow$)} & \multicolumn{5}{c}{Comparative Analysis} & \multicolumn{4}{c}{Temporal Understanding} & \multicolumn{3}{c}{Multi-view Reasoning} & Free-form QA & \multirow{2}{*}{\textbf{\texttt{O.Avg}}} \\
    \noalign{\vspace{-0.5pt}}
    \cmidrule(lr){2-6} \cmidrule(lr){7-10} \cmidrule(lr){11-13} \cmidrule(lr){14-14}
    \noalign{\vspace{-0.5pt}}
    \textbf{Method ($\downarrow$)} & BU & NC & CC & PEA & \textbf{C.Avg} & PI & FSA & PSS & \textbf{T.Avg} & MSR & MOC & \textbf{M.Avg} & CCQA & \\
    \midrule
    \textcolor{mygray1}{\textit{Human}} & \textcolor{mygray1}{85.60} & \textcolor{mygray1}{92.30} & \textcolor{mygray1}{90.70} & \textcolor{mygray1}{83.90} & \textcolor{mygray1}{88.13} & \textcolor{mygray1}{91.60} & \textcolor{mygray1}{85.20} & \textcolor{mygray1}{89.90} & \textcolor{mygray1}{88.90} & \textcolor{mygray1}{93.20} & \textcolor{mygray1}{94.20} & \textcolor{mygray1}{93.70} & \textcolor{mygray1}{85.20} & \textcolor{mygray1}{89.18} \\
    \midrule

    \multicolumn{15}{@{}l}{\textcolor{mygray2}{\textit{Closed-source Frontier Models}}} \\
    \textcolor{mygray2}{Doubao-1.5-VL-pro} & \textcolor{mygray2}{51.20} & \textcolor{mygray2}{58.10} & \textcolor{mygray2}{69.50} & \textcolor{mygray2}{36.40} & \textcolor{mygray2}{53.80} & \textcolor{mygray2}{66.90} & \textcolor{mygray2}{4.60} & \textcolor{mygray2}{36.80} & \textcolor{mygray2}{36.10} & \textcolor{mygray2}{37.40} & \textcolor{mygray2}{32.00} & \textcolor{mygray2}{34.70} & \textcolor{mygray2}{50.10} & \textcolor{mygray2}{44.30} \\
    \textcolor{mygray2}{GPT-4.1} & \textcolor{mygray2}{46.20} & \textcolor{mygray2}{34.60} & \textcolor{mygray2}{58.50} & \textcolor{mygray2}{51.20} & \textcolor{mygray2}{47.63} & \textcolor{mygray2}{70.90} & \textcolor{mygray2}{8.60} & \textcolor{mygray2}{60.50} & \textcolor{mygray2}{46.67} & \textcolor{mygray2}{38.60} & \textcolor{mygray2}{38.20} & \textcolor{mygray2}{38.40} & \textcolor{mygray2}{44.60} & \textcolor{mygray2}{45.19} \\
    \textcolor{mygray2}{Gemini-2.5-Pro} & \textcolor{mygray2}{54.20} & \textcolor{mygray2}{51.80} & \textcolor{mygray2}{68.70} & \textcolor{mygray2}{36.40} & \textcolor{mygray2}{52.78} & \textcolor{mygray2}{76.50} & \textcolor{mygray2}{13.40} & \textcolor{mygray2}{78.20} & \textcolor{mygray2}{56.03} & \textcolor{mygray2}{32.00} & \textcolor{mygray2}{25.30} & \textcolor{mygray2}{28.65} & \textcolor{mygray2}{59.80} & \textcolor{mygray2}{49.63} \\
    \midrule
    
    \multicolumn{15}{@{}l}{\textit{Open-source Single-pass Models}} \\
    Qwen3-VL-4B & 17.93 & 21.47 & 26.62 & 28.61 & 23.66 & 49.89 & 3.13 & 2.97 & 18.66 & 24.94 & 28.71 & 26.83 & 11.59 & 21.59 \\
    Qwen3-VL-8B & 23.48 & 27.35 & 40.72 & 34.94 & 31.60 & 62.54 & 5.71 & 7.22 & 25.16 & 34.51 & 27.15 & 30.83 & 31.18 & 29.47 \\
    Qwen2.5-VL-32B & 31.40 & 30.50 & 48.60 & 39.70 & 38.30 & 65.70 & 5.20 & 8.70 & 26.53 & 23.70 & \textbf{39.60} & \underline{31.65} & 41.20 & 33.78 \\
    Qwen3-VL-32B & \textbf{46.75} & 37.50 & \underline{59.54} & \underline{39.77} & \underline{45.89} & \textbf{73.00} & 4.70 & 7.50 & 28.40 & 26.62 & \underline{36.56} & 31.59 & \textbf{47.51} & \underline{38.37} \\
    \midrule
    
    \multicolumn{15}{@{}l}{\textit{Adapted Single-Video Agents}} \\
    VideoAgent-8B & 24.31 & 31.42 & 42.16 & 31.73 & 32.41 & 67.20 & 5.93 & 17.10 & 30.08 & 32.70 & 28.50 & 30.60 & 30.65 & 30.01 \\
    VCA-8B & 27.50 & 30.80 & 39.20 & 32.80 & 32.53 & 64.60 & 8.30 & \underline{21.60} & \underline{31.50} & \underline{34.40} & 26.90 & 30.65 & 33.60 & 30.71 \\
    \midrule
    
    \multicolumn{15}{@{}l}{\textit{Our Proposed Framework}} \\
    \textbf{AgentCVR-4B} & 26.41 & \underline{38.71} & 51.66 & 32.61 & 35.35 & 62.31 & \underline{12.31} & 19.42 & 31.35 & 28.32 & 31.43 & 29.88 & 26.43 & 32.06 \\
    \textbf{AgentCVR-8B} & \underline{33.88} & \textbf{43.65} & \textbf{70.71} & \textbf{41.23} & \textbf{47.35} & \underline{67.43} & \textbf{16.63} & \textbf{30.41} & \textbf{38.68} & \textbf{37.18} & 33.20 & \textbf{35.19} & \underline{41.85} & \textbf{42.03} \\
    \bottomrule
  \end{tabularx}
  \vspace{-1em}
\end{table*}

\noindent
\textbf{Baselines.}
We compare our proposed AgentCVR with the following baseline methods:
\begin{itemize}[leftmargin=*, topsep=2pt, itemsep=0pt]
    \item \textit{Closed-source models:} 
    We include Doubao-1.5-VL-pro~\cite{doubao15}, GPT-4.1~\cite{gpt}, and Gemini-2.5-Pro~\cite{gemini} as representative proprietary systems.
    \item \textit{Open-source single-pass models:}
    We evaluate Qwen2.5-VL-32B-Instruct and the Qwen3-VL series models (4B, 8B, and 32B in thinking mode) in a single-pass setting, where frames are uniformly sampled from all videos and fed to the model along with the prompt.
    \item \textit{Adapted single-video agents:}
    To assess the necessity of cross-video-specific design, we reproduce two recent single-video agents, VideoAgent~\cite{videoagent} and VCA~\cite{vca}, and adapt them to the CVR setting by concatenating the input videos with explicit video-level timestamp boundaries in the prompt.
    For fair comparison, both agents use the same 8B-scale backbone model as AgentCVR.
\end{itemize}

\vspace{0.5em}
\noindent
\textbf{Implementation details.}
We use Qwen3-VL-4B and Qwen3-VL-8B models~\cite{qwen3vl} in thinking mode as the Master Agent for AgentCVR.
The visual agent matches the Master Agent scale, while the audio agent is based on Whisper~\cite{whisper}.
For Script-Simulated RL, we use an additional Qwen3-4B~\cite{qwen3} model as the environment simulator, and GRPO training uses a maximum turn limit $T_{\max}=20$~\cite{deepseekr1}.
Detailed hyperparameter configurations and implementation settings are described in Appendix~\ref{apd_implementation}, and the prompts employed in AgentCVR are presented in Appendix~\ref{apd_prompt}.

\subsection{Main Results}

We first present an overall comparison of all methods.
The main results on the CrossVid benchmark are shown in Table~\ref{main_table}, from which we draw the following observations:


\begin{itemize}[leftmargin=*, topsep=2pt, itemsep=0pt]
    \item \textbf{Existing models face a strict trade-off between scale and practicality.}
    Open-source single-pass models show limited performance at smaller scales, only becoming competitive when scaled up to 32B-sized.
    Conversely, although closed-source frontier models achieve stronger results, their high inference cost and closed-source nature limit their applicability, especially in privacy-sensitive or local deployment scenarios.
    \item \textbf{Current single-video agent paradigms provide limited benefits for CVR.}
    Although adapted single-video agents yield slight improvements on specific tasks, their overall performance gains remain marginal.
    This limitation underscores the necessity for a more advanced, tailored framework specifically designed to optimize performance in complex CVR scenarios.
    \item \textbf{Our AgentCVR framework establishes a new standard for CVR.}
    Both AgentCVR-4B and AgentCVR-8B effectively unlock the potential of smaller-scale models, achieving substantial gains over their corresponding base models.
    Notably, AgentCVR-8B achieves optimal results across multiple tasks when evaluated against open-source single-pass models and adapted single-video agents, while delivering performance highly comparable to that of powerful closed-source frontier models.
\end{itemize}

\subsection{Ablation Studies}

To better understand the contribution of different components in AgentCVR, we conduct extensive ablation studies from two perspectives: (1) architectural and training configurations, and (2) multimodal agent synergy.

\begin{table}[t]
      \caption{
        Ablation studies of (i) architectural and training configurations, and (ii) multimodal tool synergy.
      }
      \vspace{-0.5em}
      \label{tab_ablation}
      \centering
      \resizebox{0.96\linewidth}{!}{
        \begin{tabular}{@{}c c | c c@{}}
        \toprule
        \multicolumn{2}{c|}{AgentCVR Components} & \multicolumn{2}{c}{\texttt{\textbf{O.Avg}} (\%)} \\
        \cmidrule{1-4}
        Multi-Agent & Script-Simulated RL & 4B & 8B \\
        \midrule
        \ding{55} & \ding{55} & 21.59 & 29.47 \\
        \ding{51} & \ding{55} & 30.18 & 37.20 \\
        \ding{51} & \ding{51} & \textbf{32.06} & \textbf{42.03} \\
        \midrule
        Vision Tool & Audio Tool & \multicolumn{2}{c}{8B} \\
        \midrule
        \ding{55} & \ding{51} & \multicolumn{2}{c}{14.56$^{\dagger}$} \\ 
        \ding{51} & \ding{55} & \multicolumn{2}{c}{36.07} \\ 
        \ding{51} & \ding{51} & \multicolumn{2}{c}{\textbf{42.03}} \\
        \bottomrule
        \addlinespace[0.2em]
\multicolumn{4}{p{0.98\linewidth}}{
$^{\dagger}$ Note that the overall average for the Audio-only configuration is computed exclusively over the subset of tasks that contain audio tracks.}
      \end{tabular}
      }
      \vspace{-1em}
\end{table}

\vspace{0.5em}
\noindent
\textbf{Architectural and training configurations.}
We first evaluate the individual contributions of our Multi-Agent architecture and the Script-Simulated RL paradigm by progressively adding them to the base model.
As shown in the upper part of Table~\ref{tab_ablation}, replacing the naive single-pass baseline with our multi-agent architecture yields a clear improvement at both the 4B and 8B scales.
This suggests that separating localized evidence gathering from high-level reasoning effectively overcomes the context-overload limitations of single-pass models.
Notably, introducing Script-Simulated RL further improves performance, validating the effectiveness of our training paradigm and highlighting the importance of explicit policy optimization for mastering long-horizon sub-agent interaction decisions and reliable termination behaviors beyond the capability of zero-shot prompting alone.

\vspace{0.5em}
\noindent
\textbf{Multimodal agent synergy.}
Next, we investigate the individual contributions and the synergistic effect of the visual and audio agents in AgentCVR.
As shown in the lower part of Table~\ref{tab_ablation}, relying exclusively on the audio agent leads to poor overall performance.
While audio provides useful contextual cues, the results highlight the inherent difficulty of CVR tasks in the absence of fundamental visual grounding.
Conversely, using the visual agent solely achieves relatively strong performance, indicating that visual perception is the key factor in extracting cross-video evidence.

To provide a more fine-grained analysis, we provide detailed task-level ablation breakdowns in Appendix~\ref{apd_ablation}.

\subsection{In-depth Analysis}

In this section, we conduct additional experiments to further study the design and effectiveness of our AgentCVR framework.

\subsubsection{Analysis of the Simulated Environment}

\begin{table}[t]
\caption{
Comparison of decision alignment and average per-call latency between simulated (Sim) and real (Real) environments.
}
\vspace{-0.5em}
\label{tab_sim_to_real}
\centering
\small
\setlength{\tabcolsep}{4pt}

\begin{tabularx}{\columnwidth}{@{}l c c c@{}}
\toprule
Task Category & Alignment (\%) & Sim Lat. & Real Lat. \\
\midrule
Reasoning (BU)     & 83.9$^\dagger$ & \textbf{1.6s} & 7.3s \\
Localization (FSA) & 87.4$^\ast$    & \textbf{1.7s} & 8.6s \\
\bottomrule
\addlinespace[0.2em]
\multicolumn{4}{l}{\footnotesize 
$^\ast$Measured by Sim-to-Real IoU. $^\dagger$Decision Overlap Rate.}
\end{tabularx}
\vspace{-1em}
\end{table}

A key requirement of \textit{Script-Simulated RL} is that the simulator $M_{\mathrm{sim}}$ provides feedback consistent with the real execution environment.
We evaluate Sim-to-Real alignment on two representative tasks: Functional Step Alignment (FSA), which involves predicting temporal intervals, and Behavioral Understanding (BU), formulated as multiple-choice reasoning.
To assess simulator fidelity, we construct an evaluation subset by transcribing the visual and audio content into semantic scripts $\mathcal{W}_{\mathrm{trans}}$.
We then run AgentCVR in both (i) the script-based simulator and (ii) the real execution environment, and measure consistency between these two settings, rather than accuracy against ground truth.

\vspace{0.5em}
\noindent
\textbf{Cross-environment consistency.}
As shown in Table~\ref{tab_sim_to_real}, the simulator exhibits strong alignment with the real environment.
For FSA, we measure IoU between predicted temporal intervals, showing strong agreement across the two settings and indicating that $\mathcal{W}_{\mathrm{trans}}$ preserves key temporal structure.
For BU, we report the Decision Overlap Rate, \ie the fraction of identical answer choices selected in both environments, which remains high.
Overall, the simulator provides consistent feedback for both continuous and discrete decision settings.

\vspace{0.5em}
\noindent
\textbf{Efficiency.}
We further measure the average per-call latency as reported in Table~\ref{tab_sim_to_real}.
For FSA, the simulator reduces latency from 8.6s to 1.7s per call, achieving an approximately $5\times$ speedup.
For BU, latency is reduced from 7.3s per turn in the real environment to 1.6s in simulation.
These substantial reductions significantly improve the efficiency of online RL exploration, while preserving strong alignment with real-world feedback.

\begin{figure}[!t]
  \centering
  \includegraphics[width=0.94\linewidth]{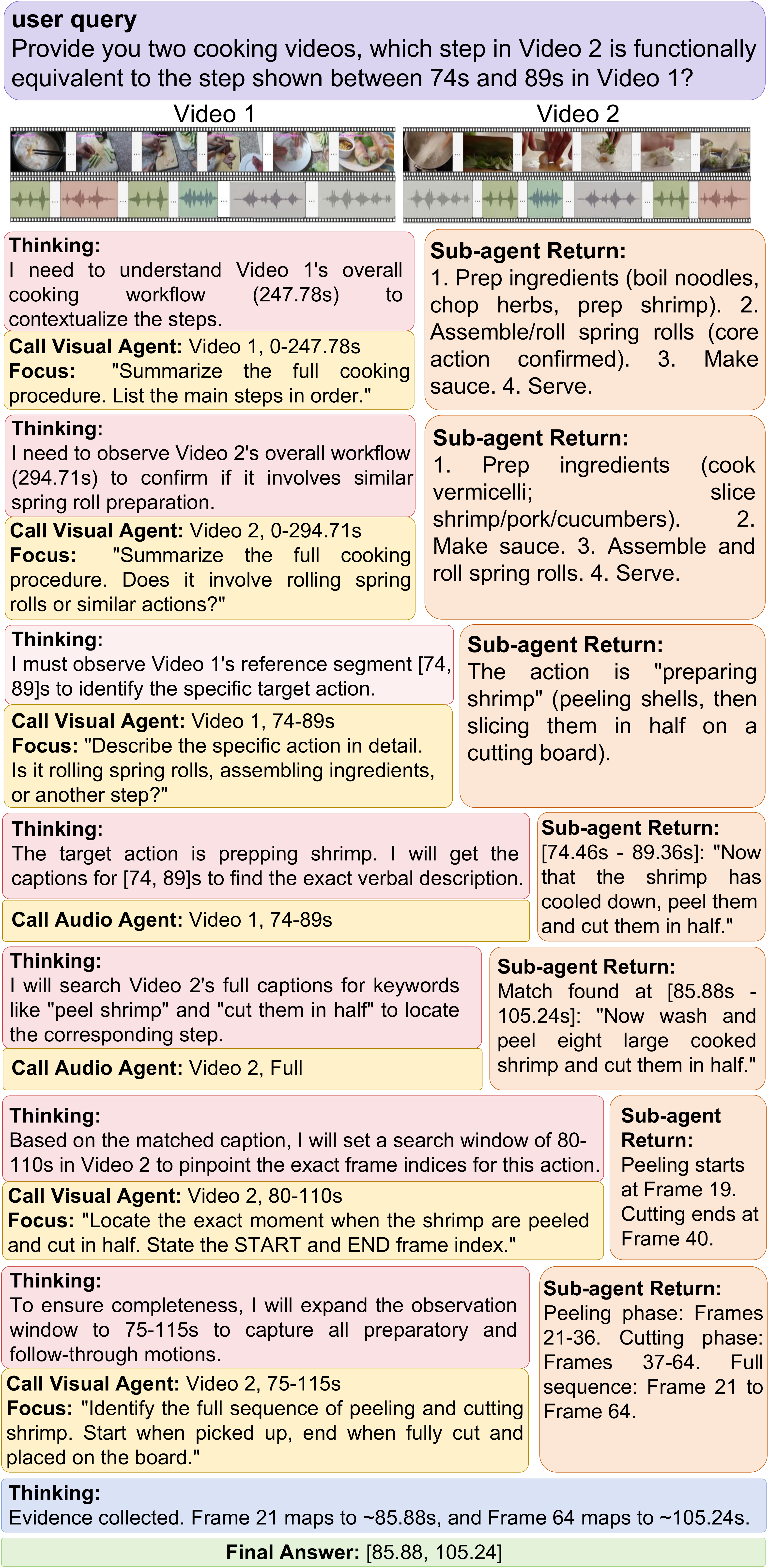}
  \vspace{-0.5em}
  \caption{A case study of AgentCVR multi-turn reasoning trace on an FSA task.}
  \label{qualitative}
  \vspace{-1em}
\end{figure}

\subsubsection{Qualitative Analysis}

Figure~\ref{qualitative} shows a representative reasoning trace of AgentCVR on a Functional Step Alignment (FSA) instance, where the goal is to localize in Video~2 a segment functionally equivalent to the reference interval in Video~1 (74s--89s).
More qualitative cases for AgentCVR are provided in Appendix~\ref{case_studies}.

The Master Agent first queries the visual agent for high-level procedure summaries of both videos and inspects the reference interval in Video~1 to identify the key action.
It then leverages complementary audio cues, querying the audio agent for corresponding narration or captions to corroborate the visual observation and narrow the search space in Video~2.
Guided by this cue, the Master Agent focuses subsequent queries on a candidate window in Video~2 (85.88s--105.24s), avoiding exhaustive scanning of the full video.
Finally, it refines the temporal boundaries through denser visual queries within the candidate window and aligns the complete action sequence.
The resulting prediction (85.88s--105.24s) achieves an IoU of $0.93$ with the ground-truth interval (86.0s--104.0s). 
This case demonstrates how iterative multi-agent interaction in AgentCVR supports localized evidence retrieval and precise temporal grounding for CVR.

\vspace{0.5em}
\noindent
\textbf{Failure modes.}
AgentCVR can still fail in long interaction traces due to instruction or format drift, and may prematurely commit to a single label in ambiguous multi-label scenarios when the available evidence is insufficient.
We provide a taxonomy of failure modes along with representative cases and detailed analyses in Appendix~\ref{failure_case}.
\section{Conclusion}

In this work, we propose AgentCVR, a novel framework that fundamentally shifts the paradigm of Cross-Video Reasoning from passive, single-pass context compression to dynamic, multi-round evidence acquisition.
By utilizing a lightweight Master Agent to iteratively coordinate specialized visual and audio tools, our approach ensures that cross-video analysis is explicitly grounded in targeted, query-relevant clues rather than an overloaded context window.
To overcome the prohibitive cost and data scarcity of online multimodal exploration, we introduce Script-Simulated RL, which replaces raw-video interactions with LLM-generated semantic scripts and a lightweight text simulator for scalable long-horizon policy learning.
Extensive experiments on the challenging CrossVid benchmark demonstrate the superiority of our approach.
Ultimately, AgentCVR establishes a new, efficient, and highly interpretable path for multimodal agents to actively reason across complex, real-world video streams.

\clearpage
\section*{Limitations}

Although AgentCVR generally outperforms open-source single-pass models and adaptive single-video agents, a marginal performance gap remains when compared to state-of-the-art closed-source frontier models, highlighting the need for further optimization.
Our failure case analysis also shows that multi-turn reasoning may suffer from instruction drift, premature cognitive closure, and over-reliance on internal parametric priors.
In addition, the current framework only optimizes the Master Agent, leaving modality-specific agents frozen without task-specific adaptation.
While the proposed Script-Simulated RL paradigm is effective, slight misalignments between the simulated and real-world environments persist, necessitating higher-fidelity simulators to ensure absolute environmental reliability.
Finally, the active multi-round interaction paradigm inevitably introduces higher inference latency than single-pass methods, indicating that future work must focus on optimizing decision efficiency and minimizing interaction turns to facilitate seamless real-world deployment.

\bibliography{custom}

\clearpage
\newtcolorbox{dialoguebox}[1][]{
  enhanced,
  breakable, 
  colback=blue!3!white, 
  colframe=blue!60!black, 
  fonttitle=\bfseries, 
  coltitle=white, 
  title=#1, 
  boxrule=0.5pt,
  arc=3pt, 
  left=5pt, right=5pt, top=5pt, bottom=5pt,
  before upper={\scriptsize} 
}

\newtcolorbox{promptbox}[1][]{
  enhanced,
  breakable,              
  colback=green!3!white,  
  colframe=green!50!black,
  fonttitle=\bfseries,    
  coltitle=white,         
  title=#1,               
  boxrule=0.5pt,
  arc=3pt,                
  left=5pt, right=5pt, top=5pt, bottom=5pt,
  before upper={\tiny} 
}

\begin{strip}
\noindent\begin{minipage}{\textwidth}

\begin{algorithm}[H]
\caption{\textbf{AgentCVR}: Training Pipeline}
\label{alg_training}
\textbf{Require:} Policy $\pi_\theta$, simulator $M_{\mathrm{sim}}$
\begin{algorithmic}[1]
\Statex \textbf{Offline script construction}
\State $\mathcal{W}_{\mathrm{script}} \leftarrow$ Generate semantic scripts for cross-video scenarios

\Statex \textbf{Online RL training}
\For{each training episode}
    \While{the agent has not produced a final answer $\hat{a}$}
        \State Sample action $a_t \sim \pi_\theta(\cdot \mid s_t)$
        \State $o_t \leftarrow M_{\mathrm{sim}}(a_t, \mathcal{W}_{\mathrm{script}})$
        \State Update the interaction history with $(a_t, o_t)$
    \EndWhile
    \State Compute the trajectory reward and update $\pi_\theta$ with GRPO
\EndFor
\end{algorithmic}
\end{algorithm}

\vspace{-1.5em}

\begin{algorithm}[H]
\caption{\textbf{AgentCVR}: Inference Pipeline}
\label{alg_inference}
\textbf{Require:} Policy $\pi_{\theta^*}$, visual tool $T_{\mathrm{vis}}$, audio tool $T_{\mathrm{aud}}$
\begin{algorithmic}[1]
\Statex \textbf{Inference on raw videos}
\For{each query $q$ and input videos $\mathcal{V}$}
    \While{the agent has not produced a final answer $\hat{a}$}
        \State Sample action $a_t \sim \pi_{\theta^*}(\cdot \mid s_t)$
        \State $o_t \leftarrow$ Execute $a_t$ using $T_{\mathrm{vis}}$ or $T_{\mathrm{aud}}$ on $\mathcal{V}$
        \State Update the interaction history with $(a_t, o_t)$
    \EndWhile
\EndFor
\end{algorithmic}
\end{algorithm}

\end{minipage}
\end{strip}

\appendix
\section{Dataset Details}\label{apd_dataset}

In this section, we provide a detailed introduction to CrossVid~\cite{crossvid}, a comprehensive large-scale benchmark specifically designed for Cross-Video Reasoning (CVR) settings.
Curated from diverse publicly available datasets, CrossVid consists of 5,331 videos and 9,015 high-quality QA pairs.
These queries cover diverse formats, including single-choice, multiple-choice, and open-ended generation, requiring models to process and reason over an average of 770 seconds of video content per query.
It encompasses 10 distinct tasks grouped into four high-level dimensions. 
We introduce each of these dimensions in the following sections.

\subsection{Comparative Analysis}

This dimension evaluates the ability of intelligent systems to extract task-relevant information from multiple videos and perform cross-video comparisons. 
It consists of the following four tasks:
\begin{itemize}[leftmargin=*, topsep=2pt, itemsep=0pt]
    \item \textbf{Behavioral Understanding (BU):} 
    Given a set of videos depicting either wildlife behaviors or everyday human activities, models are required to recognize specific actions, understand their aims and purposes, or accurately identify whether each video contains the queried action.
    \item \textbf{Narrative Comprehension (NC):} 
    Given four film clips sharing the same genre, models are required to analyze and contrast the plot, characters, environment, and underlying themes across the clips.
    \item \textbf{Culinary Comparison (CC):} 
    Given a group of videos showing the preparation of the same dishes, models must compare ingredient processing methods, utensil usage, procedural sequences, and flavor profiles across the videos.
    \item \textbf{Procedural Error Analysis (PEA):} Given videos accompanied by descriptions of possible errors, models are required to identify specific errors mentioned in the query and trace the reasons for these mistakes.
\end{itemize}

\subsection{Temporal Understanding}

This dimension assesses the capability of intelligent systems to perform temporal localization and chronological reasoning across multiple video streams. 
It contains three tasks:
\begin{itemize}[leftmargin=*, topsep=2pt, itemsep=0pt]
    \item \textbf{Plot Inference (PI):} 
    Given the beginning and ending segments of a film, the model is asked to infer the missing plot in the middle part.
    \item \textbf{Functional Step Alignment (FSA):} 
    Given two different videos, models are asked to locate a specific temporal segment in one video that corresponds to a specified time interval in the other, requiring the alignment of corresponding steps based on semantic and functional equivalence.
    \item \textbf{Procedural Step Sequencing (PSS):} 
    A single video is segmented at the step level, and the clips are randomly shuffled. 
    Models must reconstruct the correct temporal sequence, which evaluates their causal reasoning and temporal inference capabilities.
\end{itemize}

\subsection{Multi-view Reasoning}

This dimension provides intelligent systems with two temporally synchronized road videos, each captured from a different aerial drone perspective. 
It consists of two tasks:
\begin{itemize}[leftmargin=*, topsep=2pt, itemsep=0pt]
    \item \textbf{Multi-view Spatial Reasoning (MSR):} 
    Models are queried about spatial relationships, such as the relative distances and positions of specific objects at a given moment across different views.
    \item \textbf{Multi-view Object Counting (MOC):} 
    Models are required to count specific objects at a certain moment or over a defined time interval, which requires the integration of multi-perspective information for precise counting.
\end{itemize}

\subsection{Free-form QA}

This dimension evaluates the ability of intelligent systems to perform comparative analysis and answer open-ended questions comprehensively:
\begin{itemize}[leftmargin=*, topsep=2pt, itemsep=0pt]
    \item \textbf{Comparative Culinary QA (CCQA):} 
    Two videos featuring the same items are provided. 
    Models are required to compare them and generate a detailed textual response identifying the differences in procedures, assessing their capability to compare fine-grained details without predefined options.
\end{itemize}

\section{Implementation Details}\label{apd_implementation}

For Script-Simulated RL training, we use the open-source \texttt{verl}~\cite{verl} framework with a GRPO-based~\cite{deepseekmath} optimization objective.
For each query, we sample $G=8$ complete trajectories and constrain the multi-turn interaction with a maximum turn limit $T_{\max} = 20$ and a tolerance zone $T_{\mathrm{tol}} = 10$ to prevent infinite looping.
The optimizer used in RL is AdamW~\cite{adamw}.
The detailed hyperparameters are summarized in Table~\ref{rl_hyperparams}.
During RL rollout and inference, we use \texttt{vLLM}~\cite{vllm} engine to accelerate generation.
We configure the model with a maximum length of 8192 tokens and set the temperature to 0.0. 
All experiments are conducted on a single compute node equipped with 8 NVIDIA H800 (80GB) GPUs.

\begin{table}[!t]
  \caption{Detailed configurations used for training AgentCVR via GRPO.}
  \vspace{-0.5em}
  \label{rl_hyperparams}
  \centering
  \resizebox{0.48\textwidth}{!}{
  \begin{tabular}{@{}lcc@{}}
    \toprule
    \textbf{Hyperparameter} & \textbf{AgentCVR-4B} & \textbf{AgentCVR-8B} \\
    \midrule
    Base Model & \texttt{Qwen3-VL-4B} & \texttt{Qwen3-VL-8B} \\
    Learning Rate & 1e-7 & 5e-7 \\
    KL Coefficient ($\epsilon$) & 0.005 & 0.005 \\
    Samples per Prompt ($G$) & 8 & 8 \\
    Training Batch Size & 32 & 32 \\
    PPO Mini-Batch Size & 16 & 16 \\
    Micro Batch Size per GPU & 1 & 1 \\
    Max Prompt Length & 4096 & 4096 \\
    Max Response Length & 4096 & 4096 \\
    Epochs & 2 & 3 \\
    \bottomrule
  \end{tabular}
  }
  \vspace{-1em}
\end{table}

\section{Sub-Agents Details}

\subsection{Sub-Agent Configurations}
During the real-world zero-shot inference phase, AgentCVR strictly interacts with physical environments using the following specialized agents:
\begin{itemize}[leftmargin=*, topsep=2pt, itemsep=0pt]
    \item \textbf{Visual Agent:} 
    We employ the \texttt{Qwen3-VL-8B} model in thinking mode as our dedicated visual perception module. 
    When the Master Agent dispatches a visual observation request, frames from the targeted video segment are uniformly sampled. 
    To balance visual fidelity and computational efficiency, the extracted frames are resized so that the longer side is 360 pixels, while maintaining the original aspect ratio.
    \item \textbf{Audio Agent:}
    We utilize the \texttt{Whisper large-v3} model to execute selective auditory extraction. 
    To ensure robust background noise filtering and precise transcription, the model uses a beam size of 5, a decoding temperature of $0.0$, a log-probability threshold of $-1.0$, disables conditioning on previous text, and applies a no-speech threshold of $0.6$.
\end{itemize}

\subsection{Agent Fidelity Cases}

A fundamental prerequisite for our Script-Simulated RL is that the surrogate text simulator ($\mathcal{S}_{text}$) must faithfully replicate the semantic feedback of the real physical agents ($\mathcal{T}_{real}$). 
This ensures that the reasoning meta-policy learned by the Master Agent during the offline training phase can seamlessly transfer to the online real-world inference phase without semantic drift. 

To intuitively demonstrate this high Sim-to-Real fidelity, we present a comparative case study on a Functional Step Alignment (FSA) task. 
The Master Agent is tasked with finding a step in Video 2 that is functionally equivalent to the reference segment (55s--65s) in Video 1. 

As shown in the comparative dialogue boxes below, the Master Agent exhibits a remarkably consistent strategic meta-policy across both the simulated and real environments. 
In both settings, the agent autonomously develops a sophisticated cross-modal verification strategy: 
It first dispatches the visual agent to understand the physical action of the reference segment (\textit{pressing with a cloth to extract liquid}). 
It then invokes the audio agent to extract specific verbal anchors (\textit{"extract the water"}). 
Next, it efficiently searches the target video's audio track to locate functional synonyms (\textit{"press... dry them off"}). 
Finally, it conducts a visual verification to establish precise temporal boundaries.

\begin{dialoguebox}[Simulated Environment Training Trace (Using $\mathcal{S}_{text}$)]
\textbf{Task:} Functional Step Alignment (FSA) \\
\textbf{Query:} \textit{Provide you two cooking videos, which step in Video 2 is functionally equivalent to the step shown between 55s and 65s in Video 1?}

\textbf{Turn 1 \& 2 (Visual Exploration)} \\
\textbf{Thinking:} I need to observe Video 1's reference segment [55, 65]s to confirm the exact action. What is the person doing with hands/tools? \\
\textbf{Action:} {[call vision: Video 1, 55-65s, focus: "Describe the exact action..."]} \\
\textbf{Observation:} Simulated Tool Return: The chef uses a \textbf{white cloth to press} the ingredient. The action appears to extract any remaining juice.

\textbf{Turn 3 (Audio Grounding)} \\
\textbf{Thinking:} The reference segment describes extracting liquid. I need to get the captions to confirm the specific verbal description. \\
\textbf{Action:} {[call get\_caption: Video 1, 55-65s]} \\
\textbf{Observation:} Simulated Tool Return: [55.54s - 57.54s]: "Now we \textbf{extract the water}." [59.00s - 66.60s]: "A potato pulp, nice and dry..."

\textbf{Turn 4 (Cross-Video Audio Search)} \\
\textbf{Thinking:} Video 1 mentions "extract the water". I will search Video 2's full captions for similar functional keywords like "squeeze", "extract juice", or "dry". \\
\textbf{Action:} {[call get\_caption: Video 2, Full]} \\
\textbf{Observation:} Simulated Tool Return: ...[43.96s - 51.54s]: "get a nice clean tea towel." [55.56s - 60.76s]: "Fold the tea towel over and just give it a little wiggle and a \textbf{press. Just to dry them} off."...

\textbf{Turn 5 (Final Visual Verification)} \\
\textbf{Thinking:} The caption [55.56s-60.76s] matches the functional purpose. I need to verify if this visual action is removing moisture. \\
\textbf{Action:} {[call vision: Video 2, 55.56-60.76s, focus: "What is the purpose of pressing the tea towel?"]} \\
\textbf{Observation:} Simulated Tool Return: The purpose of pressing the tea towel on the potatoes is to \textbf{remove excess moisture}.

\textbf{Turn 6 (Final)} \\
\textbf{Thinking:} Evidence collected. The functional equivalence is perfectly aligned. \\
\textbf{Action:} {[Answer]} \\
\textbf{Final Answer:} [55.56, 60.76]
\end{dialoguebox}

\vspace{1em} 

\begin{dialoguebox}[Real-World Inference Trace (Using Physical $\mathcal{T}_{real}$)]
\textbf{Task:} Functional Step Alignment (FSA) \\
\textbf{Query:} \textit{Provide you two cooking videos, which step in Video 2 is functionally equivalent to the step shown between 55s and 65s in Video 1?}

\textbf{Turn 1 \& 2 (Visual Exploration)} \\
\textbf{Thinking:} I need to precisely observe Video 1's reference segment [55, 65]s to determine the specific visual features of the action. \\
\textbf{Action:} {[call vision: Video 1, 55-65s, focus: "Describe the specific action... What is the tool used?"]} \\
\textbf{Observation:} Real Vision Tool Return: Tool used: A \textbf{white cloth}. Action: The chef is \textbf{pressing/squeezing} the vegetable to extract juice into a metal bowl.

\textbf{Turn 3 (Audio Grounding)} \\
\textbf{Thinking:} The action is 'pressing vegetable to extract juice'. I will get the captions nearby to find text keywords like 'squeezing' or 'pressing'. \\
\textbf{Action:} {[call get\_caption: Video 1, 55-65s]} \\
\textbf{Observation:} Real Audio Tool Return: [55.54s - 57.54s]: "Now we \textbf{extract the water}." [59.00s - 66.60s]: "A potato pulp, nice and dry..."

\textbf{Turn 4 (Cross-Video Audio Search)} \\
\textbf{Thinking:} Video 1's caption mentions 'extract the water'. I will retrieve Video 2's captions to find corresponding keywords for extracting liquid. \\
\textbf{Action:} {[call get\_caption: Video 2, Full]} \\
\textbf{Observation:} Real Audio Tool Return: ...[43.96s - 51.54s]: "get a nice clean tea towel." [55.56s - 60.76s]: "Fold the tea towel over and just give it a little wiggle and a \textbf{press. Just to dry them} off."...

\textbf{Turn 5 (Final Visual Verification)} \\
\textbf{Thinking:} The caption at 55.56s-60.76s explicitly mentions pressing to dry. I need to observe this exact window (expanded slightly to 53-63s) to locate the precise frame boundaries of the pressing action. \\
\textbf{Action:} {[call vision: Video 2, 53-63s, focus: "Identify the full pressing action of the tea towel... Provide START and END frame indices."]} \\
\textbf{Observation:} Real Vision Tool Return: The full pressing action spans from \textbf{START frame 27} to \textbf{END frame 34}.

\textbf{Turn 6 (Final)} \\
\textbf{Thinking:} Evidence collected. The observation window is 53s - 63s (10s duration). Start frame 27 translates to 53.00s, and End frame 34 translates to 63.00s. \\
\textbf{Action:} {[Answer]} \\
\textbf{Final Answer:} [53.00, 63.00]
\end{dialoguebox}

This compelling alignment proves that the mock observations generated by $\mathcal{S}_{text}$ during RL training are robust enough to cultivate advanced, generalized multimodal agent interaction behaviors. 
Consequently, the agent can seamlessly replace the text simulator with heavy physical agents during the zero-shot inference phase without requiring any real-video fine-tuning, further highlighting the effectiveness of our Script-Simulated RL paradigm in AgentCVR.

\section{Training Dynamics}

\begin{figure*}[!t]
      \centering
      \begin{minipage}{0.49\linewidth}
        \includegraphics[width=\linewidth]{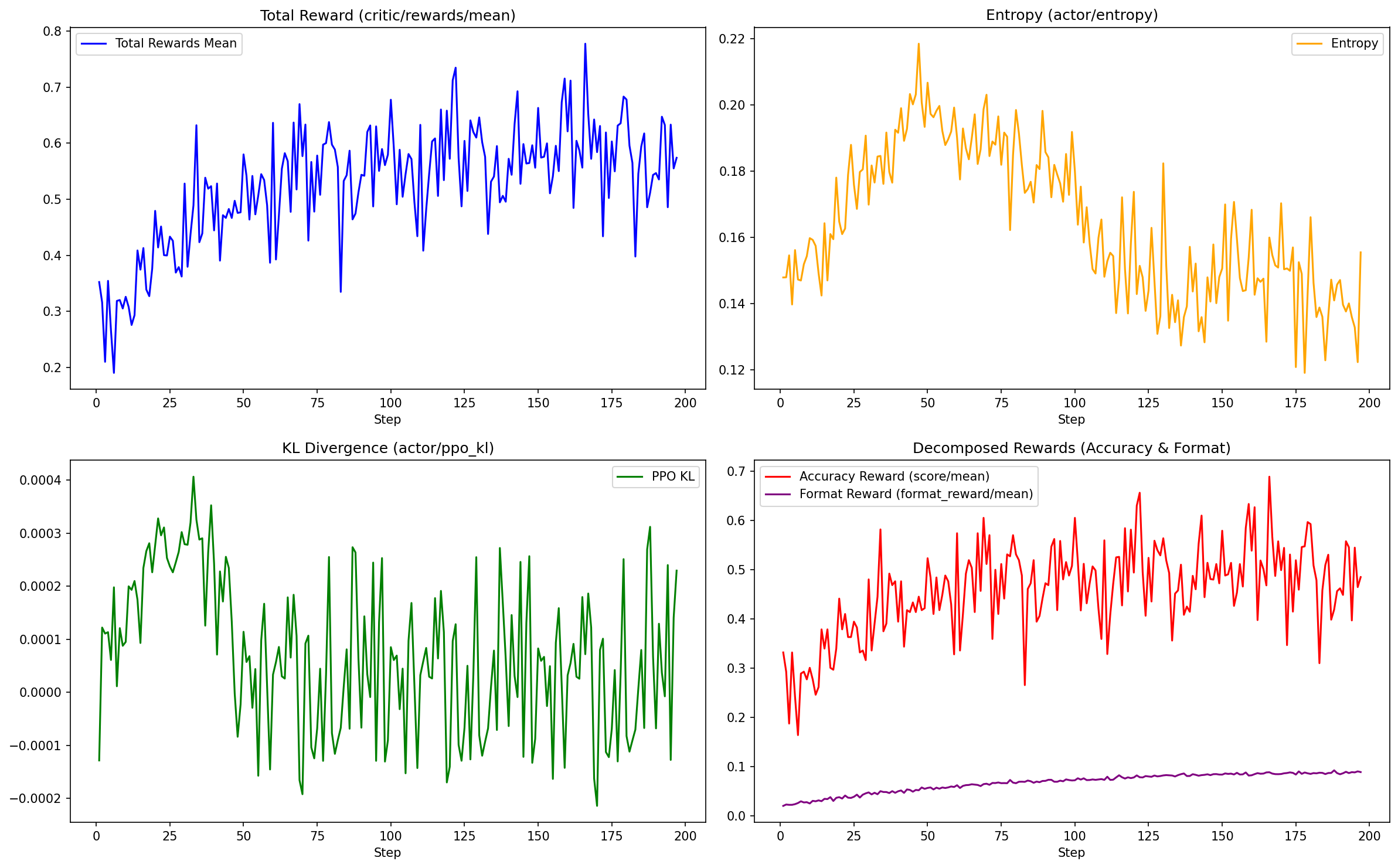}
        \centerline{(a) AgentCVR-4B}
      \end{minipage}\hfill
      \begin{minipage}{0.49\linewidth}
        \includegraphics[width=\linewidth]{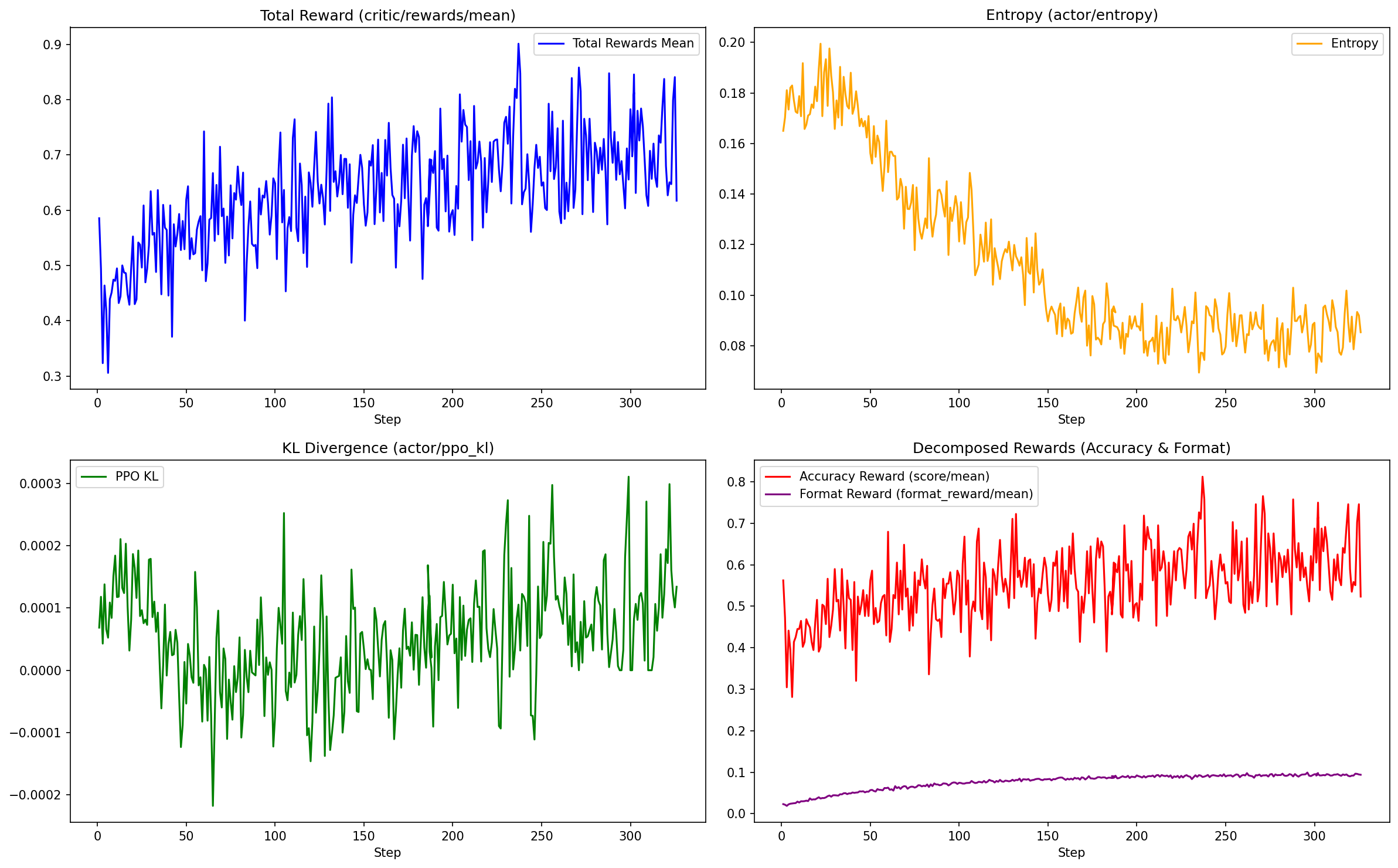}
        \centerline{(b) AgentCVR-8B}
      \end{minipage}
      \vspace{-0.5em}
      \caption{The RL training dynamics for (a) AgentCVR-4B and (b) AgentCVR-8B during the GRPO training phase, illustrating the convergence of Total Reward, Entropy, KL Divergence, and Decomposed Rewards.}
      \label{rl_curves}
    \vspace{-0.5em}
\end{figure*}

\vspace{1em}

\definecolor{mygray1}{HTML}{8F8C8C}
\definecolor{mygray2}{HTML}{666565}

\begin{table*}[!t]
  \caption{
  Experimental results of detailed task-level ablation studies. 
  The best results are highlighted in \textbf{bold}, the second-best results are \underline{underlined}, and \textbf{\texttt{O.Avg}} denotes the overall average.
  }
  \vspace{-0.5em}
  \label{tab_apd_abaltion}
  \centering
  \renewcommand{\arraystretch}{1.15}

  \fontsize{8.3}{9}\selectfont
  \setlength{\tabcolsep}{1.5pt}

  \newcolumntype{Y}{>{\centering\arraybackslash}X}

  \begin{tabularx}{\textwidth}{@{} l @{\hspace{6pt}} *{4}{Y} @{\hspace{6pt}} *{3}{Y} @{\hspace{6pt}} *{2}{Y} @{\hspace{6pt}} c @{\hspace{6pt}} Y @{}}
    \toprule

    \textbf{Task ($\rightarrow$)} & \multicolumn{4}{c}{Comparative Analysis} & \multicolumn{3}{c}{Temporal Understanding} & \multicolumn{2}{c}{Multi-view Reasoning} & Free-form QA & \multirow{2}{*}{\textbf{\texttt{O.Avg}}} \\

    \noalign{\vspace{-0.5pt}}
    \cmidrule(lr){2-5} \cmidrule(lr){6-8} \cmidrule(lr){9-10} \cmidrule(lr){11-11}
    \noalign{\vspace{-0.5pt}}

    \textbf{Method ($\downarrow$)} & BU & NC & CC & PEA & PI & FSA & PSS & \makebox[6em]{MSR} & \makebox[6em]{MOC} & CCQA & \\

    \midrule

    \multicolumn{12}{@{}l}{\textit{Open-source Single-pass Models}} \\
    Qwen3-VL-4B & 17.93 & 21.47 & 26.62 & 28.61 & 49.89 & 3.13 & 2.97 & \makebox[6em]{24.94} & \makebox[6em]{28.71} & 11.59 & 21.59 \\
    Qwen3-VL-8B & 23.48 & 27.35 & 40.72 & 34.94 & 62.54 & 5.71 & 7.22 & \makebox[6em]{\underline{34.51}} & \makebox[6em]{27.15} & 31.18 & 29.47 \\
    \midrule

    \multicolumn{12}{@{}l}{\textit{Multi-Agent Framework Only}} \\
    AgentCVR-4B (Zero-Shot) & 23.70 & 28.57 & 50.95 & 32.17 & 59.84 & 9.11 & 16.30 & \makebox[6em]{28.42} & \makebox[6em]{27.63} & 29.60 & 30.18 \\
    AgentCVR-8B (Zero-Shot) & 24.29 & \underline{42.92} & \underline{67.66} & \underline{40.71} & 62.94 & \textbf{16.84} & 21.35 & \makebox[6em]{33.71} & \makebox[6em]{29.37} & 33.80 & \underline{37.20} \\
    \midrule

    \multicolumn{12}{@{}l}{\textit{Multimodal Tool Synergy}} \\
    Audio-only & 14.70 & 26.30 & 23.72 & - & 32.20 & 0.70 & 0.40 & \makebox[6em]{-} & \makebox[6em]{-} & - & 14.56$^{\dagger}$ \\
    Visual-only & \underline{27.26} & 33.07 & 53.58 & \textbf{41.23} & \underline{62.95} & 6.21 & \underline{28.42} & \makebox[6em]{32.53} & \makebox[6em]{\textbf{37.18}} & \underline{35.20} & 36.07 \\
    \midrule

    \multicolumn{12}{@{}l}{\textit{Our Proposed Framework}} \\
    \textbf{AgentCVR-4B} & 26.41 & 38.71 & 51.66 & 32.61 & 62.31 & 12.31 & 19.42 & \makebox[6em]{28.32} & \makebox[6em]{31.43} & 26.43 & 32.06 \\
    \textbf{AgentCVR-8B} & \textbf{33.88} & \textbf{43.65} & \textbf{70.71} & \textbf{41.23} & \textbf{67.43} & \underline{16.63} & \textbf{30.41} & \makebox[6em]{\textbf{37.18}} & \makebox[6em]{\underline{33.20}} & \textbf{41.85} & \textbf{42.03} \\

    \bottomrule
\addlinespace[0.2em]
\multicolumn{12}{p{0.96\linewidth}}{$^{\dagger}$ Note that the overall average for the Audio-only configuration is computed exclusively over the subset of tasks that contain audio tracks.}
  \end{tabularx}
  \vspace{-1em}
\end{table*}

As illustrated in Figure~\ref{rl_curves}, both AgentCVR-4B and AgentCVR-8B demonstrate highly stable and effective policy optimization during the Script-Simulated RL phase.

First, regarding reward convergence and task decomposition, the \textit{Total Reward} for both models shows a consistent upward trend. 
By examining the \textit{Decomposed Rewards}, we observe a distinct two-stage learning process. 
The formatting reward ($R_{\mathrm{fmt}}$) converges rapidly in the early steps, indicating that the agent quickly learns to produce valid API calls and to strictly follow the reasoning-trace constraints.
Once the format is mastered, the correctness reward ($R_{\mathrm{ans}}$) becomes the primary driver of optimization, steadily climbing as the agent learns optimal active perception strategies to navigate the cross-video environment and identify correct answers.

Furthermore, the \textit{Entropy} curves reflect the exploration-exploitation trade-off and the emergence of policy confidence. 
Notably, AgentCVR-8B exhibits a highly pronounced and smooth entropy decay, dropping from approximately 0.18 to 0.08. 
This signifies that the larger model effectively transitions from broad exploration to a highly confident, deterministic agent interaction policy. 
While AgentCVR-4B also shows a general downward trend, it exhibits higher variance, reflecting the typical capacity limits of smaller models when establishing rigid cognitive closure criteria.

Finally, we evaluate training stability and capability retention through the \textit{KL Divergence}. 
Notably, the \textit{KL Divergence} for both variants remains consistently bounded within a very small range, on the order of $10^{-4}$, indicating stable optimization without significant capability drift.
This result suggests that our method improves multi-agent reasoning and interaction capabilities while largely preserving the base model's inherent linguistic and cognitive abilities.

\section{Detailed Ablation Studies}\label{apd_ablation}

In this section, we provide a comprehensive per-task breakdown of the ablation results in Table~\ref{tab_apd_abaltion}.
The results show that the full AgentCVR configuration equipped with both visual and audio agents achieves the most balanced performance across all four high-level dimensions.

\section{Turns \& Frames}

In this section, we detail the maximum allowable number of frames and average interaction turns for AgentCVR and the compared methods as shown in Table~\ref{tab_frame_turn}.

Compared with single-pass models that process all inputs in a single forward pass, agent-based methods perform multi-turn interactions to progressively retrieve and reason over relevant video content.
We observe that AgentCVR does not reach the pre-defined maximum limit of 20 interaction turns in practice.
Instead, its average number of turns remains comparable to other agent-based methods, indicating that the framework does not incur additional interaction overhead.
Meanwhile, AgentCVR achieves substantially higher frame coverage than existing baselines, demonstrating that it can efficiently utilize each interaction step to access a broader visual context without increasing the overall interaction cost.
This shows that AgentCVR maintains strong efficiency while improving effective visual exploration for CVR tasks.

\begin{table}[t]
  \caption{
Maximum allowable number of frames and average interaction turns for AgentCVR and the compared methods.
  }
  \vspace{-0.5em}
  \label{tab_frame_turn}
  \centering
  \renewcommand{\arraystretch}{1.15}
  
  \fontsize{8.5}{9}\selectfont
  
  \setlength{\tabcolsep}{6pt} 
  
  \begin{tabular}{@{}lcc@{}}
    \toprule
    \textbf{Method} & \textbf{\#Frame} & \textbf{\#Turn} \\
    \midrule
    
    \multicolumn{3}{@{}l}{\textit{Closed-source Frontier Models}} \\
    
    Doubao-1.5-VL-pro & 256 & 1 \\
    GPT-4.1 & 48 & 1 \\
    Gemini-2.5-Pro & 128 & 1 \\
    
    \midrule
    \multicolumn{3}{@{}l}{\textit{Open-source Single-pass Models}} \\
    
    Qwen3-VL-4B & 256 & 1 \\
    Qwen3-VL-8B & 256 & 1 \\
    Qwen2.5-VL-32B & 256 & 1 \\
    Qwen3-VL-32B & 256 & 1 \\
    
    \midrule
    \multicolumn{3}{@{}l}{\textit{Adapted Single-Video Agents}} \\
    
    VideoAgent-8B & 37.3 & 11.6 \\
    VCA-8B & 70.5 & 16.6 \\
    
    \midrule
    \multicolumn{3}{@{}l}{\textit{Our Proposed Framework}} \\
    
    \textbf{AgentCVR-4B} & 306.8 & 14.4 \\
    \textbf{AgentCVR-8B} & 355.2 & 16.7 \\
    
    \bottomrule
  \end{tabular}
  
  \vspace{-1em}
\end{table}

\section{Case Studies}\label{case_studies}

\subsection{Successful Cases}\label{success_case}

In this section, we provide a detailed qualitative analysis of AgentCVR's inference process to illustrate its robust multi-agent reasoning capabilities. 
To demonstrate the versatility of our framework, we select representative success cases across all four evaluation dimensions of the CrossVid benchmark: Comparative Analysis, Temporal Understanding, Multi-view Reasoning, and Free-form QA. 
The following interaction transcripts detail the multi-turn cognitive processes, highlighting how the Master Agent dynamically formulates active perception strategies, precisely invokes specialized visual and audio agents to extract fine-grained multimodal cues, and systematically synthesizes cross-video evidence to deduce the final answers.

\subsubsection{Dimension: Comparative Analysis} 
\begin{dialoguebox}[Task: Behavioral Understanding (BU)] 
\textbf{Query:} \textit{Which behavior in the following videos is most likely to help the animal avoid predators?} \\
\textbf{Options:} A. Video one. B. Video two. C. Video three. D. Video four. \\
\textbf{Ground Truth:} D

\textbf{Turn 1} \\
\textbf{Thinking:} Task initialization. I need to perform a comprehensive overview of Video 1 to identify the main character, setting, and primary action to determine if it relates to the query. \\
\textbf{Action:} [call vision: Video 1, 0-3.33s, focus: "Describe the main character and their primary action in detail. What is the setting? What objects are present?"] \\
\textbf{Observation:} Visual Agent Return: The main character is a single bird moving through dry grass. The setting is an open, natural field. It does not exhibit defensive behaviors, showing only natural movement.

\textbf{Turn 2} \\
\textbf{Thinking:} Continuing the scan. Now I will analyze Video 2 to check for any defensive behaviors, such as freezing, camouflaging, or rapid escape. \\
\textbf{Action:} [call vision: Video 2, 0-3.41s, focus: "Identify if the subject exhibits defensive behaviors like freezing, camouflaging, or rapid escape. Are there any signs of predator presence?"] \\
\textbf{Observation:} Visual Agent Return: The bird is flying normally. There is no freezing, camouflaging, or rapid escape, nor are there any signs of predator presence.

\textbf{Turn 3} \\
\textbf{Thinking:} Continuing the scan. I will analyze Video 3 for defensive behaviors or signs of predators. \\
\textbf{Action:} {[call vision: Video 3, 0-8.0s, focus: "Identify if the subject exhibits defensive behaviors like freezing, camouflaging, or rapid escape. Are there any signs of predator presence?"]} \\
\textbf{Observation:} Visual Agent Return: The birds exhibit calm, natural movements like walking and preening. No defensive behaviors or signs of predator presence are observed.

\textbf{Turn 4} \\
\textbf{Thinking:} Finally, I will analyze Video 4 to confirm if there are defensive behaviors or predator threats. \\
\textbf{Action:} {[call vision: Video 4, 0-3.20s, focus: "Identify if the subject exhibits defensive behaviors like freezing, camouflaging, or rapid escape. Are there any signs of predator presence?"]} \\
\textbf{Observation:} Visual Agent Return: The animal remains stationary and clings to the tree trunk. Its coloration and texture blend with the bark, indicating camouflage to avoid detection. This clearly exhibits defensive behavior.

\textbf{Turn 5 (Final)} \\
\textbf{Thinking:} After a comprehensive analysis of all four videos: Video 1 shows a bird moving in dry grass; Video 2 shows a bird flying normally; Video 3 shows birds moving calmly; Video 4 shows an animal actively avoiding detection through freezing and camouflage. The conclusion is option D. \\
\textbf{Action:} {[Answer]} \\
\textbf{Final Answer:} D
\end{dialoguebox}

\begin{dialoguebox}[Task: Narrative Comprehension (NC)]
\textbf{Query:} \textit{Which two videos feature home invasions as key plot points?} \\
\textbf{Options:} A. Video1 and Video2 \quad B. Video2 and Video3 \quad C. Video3 and Video4 \quad D. Video1 and Video4 \\
\textbf{Ground Truth:} C

\textbf{Turn 1} \\
\textbf{Thinking:} Task initiation. As a movie analysis expert, I first need to get the subtitles for Video 1 to understand the plot context. Captions will help me quickly locate keywords like 'home invasion', 'break in', or 'forced entry'. \\
\textbf{Action:} {[call get\_caption: Video 1, Full]} \\
\textbf{Observation:} Audio Agent Return: [Subtitles retrieved. Mentions witness protection, but no direct home invasion keywords are found.]

\textbf{Turn 2} \\
\textbf{Thinking:} Continuing the process, I will fetch the full captions for Video 2 to check for similar keywords that might imply a home invasion plot. \\
\textbf{Action:} {[call get\_caption: Video 2, Full]} \\
\textbf{Observation:} Audio Agent Return: [Subtitles retrieved. Mentions "Eastern Road, main house" and "search and rescue force".]

\textbf{Turn 3} \\
\textbf{Thinking:} Next, I will get the full captions for Video 3 to check for related keywords. \\
\textbf{Action:} {[call get\_caption: Video 3, Full]} \\
\textbf{Observation:} Audio Agent Return: [Subtitles retrieved. Mentions "search all informants who recently entered witness protection".]

\textbf{Turn 4} \\
\textbf{Thinking:} Finally, I will get the captions for Video 4 to locate key dialogue segments. \\
\textbf{Action:} {[call get\_caption: Video 4, Full]} \\
\textbf{Observation:} Audio Agent Return: [Subtitles retrieved. Mentions "Search everywhere! We need them alive!".]

\textbf{Turn 5} \\
\textbf{Thinking:} Subtitle analysis suggests Videos 1 and 2 might involve entering a house based on certain phrases. I need to use the Visual Agent to confirm if the specific timestamps corresponding to these phrases show actual home invasion scenes. \\
\textbf{Action:} {[call vision: Video 1 \& Video 2, specific timestamps, focus: "Check for house checking or forced entry. Note any signs of family members being invaded."]} \\
\textbf{Observation:} Visual Agent Return: Video 1 shows a calm setting with no signs of forced entry or search actions. Video 2 shows an outdoor street combat scene, not a house interior. Neither depicts a home invasion.

\textbf{Turn 6} \\
\textbf{Thinking:} Now I need to visually check the suspicious timestamps in Video 3 and Video 4 to see if they depict home invasions or forced entries. \\
\textbf{Action:} {[call vision: Video 3 \& Video 4, specific timestamps, focus: "Check if Video 3 involves searching a witness's home. Check if Video 4 shows family members being searched or invaded. Look for violent or forced entry."]} \\
\textbf{Observation:} Visual Agent Return: Video 3 shows a man on a phone in a normal home setting with no forced entry. Video 4 shows armed tactical personnel forcibly entering a residence with a scared woman and child present, indicating a clear home invasion.

\textbf{Turn 7 (Final)} \\
\textbf{Thinking:} After multiple rounds of analysis: Video 1 and Video 2 have no home invasion scenes. Video 4 clearly shows an armed home invasion targeting family members. Video 3 involves investigating/searching for a witness, which aligns with the plot point context. Through elimination and matching the available pairs in the options, the correct choice is C (Video 3 and Video 4). \\
\textbf{Action:} {[Answer]} \\
\textbf{Final Answer:} C
\end{dialoguebox}

\begin{dialoguebox}[Task: Culinary Comparison (CC)]
\textbf{Query:} \textit{What distinguishes the moisture removal method in Video 3 from the others? Options: A. Uses a metal press, B. Relies on sunlight drying, C. Employs a sieve, D. Skips moisture removal}  \\
\textbf{Ground Truth:} C 

\textbf{Turn 1} \\
\textbf{Thinking:} Task started; as an initialization step, I must first get the captions for Video 1, which is a critical step for building a global index. \\
\textbf{Action:} {[call get\_caption: Video 1, Full]}  \\
\textbf{Observation:} Audio Agent Return: [94.00s - 100.84s] "...take all the starchy moisture out of this as you can...".

\textbf{Turn 2} \\
\textbf{Thinking:} Continuing the initialization process, I now need to get the captions for Video 2 to ensure all text information is gathered. \\
\textbf{Action:} {[call get\_caption: Video 2, Full]}  \\
\textbf{Observation:} Audio Agent Return: [72.82s - 77.80s] "Okay now we're going to try to squeeze some of the moisture out of these. You can press it through a colander...".

\textbf{Turn 3} \\
\textbf{Thinking:} Continuing the initialization workflow, I will retrieve the captions for Video 3. \\
\textbf{Action:} {[call get\_caption: Video 3, Full]}  \\
\textbf{Observation:} Audio Agent Return: [114.82s - 120.68s] "we're gonna strain this right so let's get all of this right into our strainer...".

\textbf{Turn 4} \\
\textbf{Thinking:} Completing the initialization by getting the captions for Video 4. \\
\textbf{Action:} {[call get\_caption: Video 4, Full]}  \\
\textbf{Observation:} Audio Agent Return: [101.66s - 109.58s] "wrap up the towel tightly and squeeze out all the liquids...".

\textbf{Turn 5} \\
\textbf{Thinking:} Based on the captions, Video 3 mentions using a strainer, while Video 2 mentions a colander. I need to visually compare the moisture removal methods across all four videos to verify these descriptions. \\
\textbf{Action:} {[call vision: Videos 1, 2, 3, 4, focus: "Compare moisture removal methods: Check if Video 3 uses a sieve (strainer)..."]}  \\
\textbf{Observation:} Visual Agent Return: Video 3 uses a sieve (Correct). Video 2 uses manual squeezing, not a colander (Incorrect). Video 1 uses hands without a tea cloth (Incorrect). Video 4 uses a dishcloth (Correct).

\textbf{Turn 6} \\
\textbf{Thinking:} Video 3 definitely uses a sieve. I will double-check if Video 4 skips moisture removal or if Video 1 matches any other drying methods. \\
\textbf{Action:} {[call vision: Videos 1, 4, focus: "Check if Video 1 uses a tea cloth and if Video 4 skips moisture removal."]}  \\
\textbf{Observation:} Visual Agent Return: Video 1 does not use a tea cloth. Video 4 does not skip moisture removal.

\textbf{Turn 7 (Final)} \\
\textbf{Thinking:} Evidence collected over the turns confirms that Video 3's method of using a sieve is distinctly different from the others, aligning with Option C. \\
\textbf{Action:} {[Answer]}  \\
\textbf{Final Answer:} C
\end{dialoguebox}

\begin{dialoguebox}[Task: Procedural Error Analysis (PEA)]
\textbf{Query:} \textit{Which action is incorrectly performed in exactly two videos?} \\
\textbf{Options:} A. detach the base and the chassis \quad B. attach the sound module and the chassis \quad C. detach the bumper and the cabin \quad D. attach the rear roof and the transport cabin \\
\textbf{Ground Truth:} A

\textbf{Turn 1} \\
\textbf{Thinking:} The question asks to identify an action that is incorrectly performed in exactly two videos. Incorrect actions usually involve detaching rather than assembling (attaching). I need to observe the key segments of all videos to identify the incorrect actions and count their occurrences. \\
\textbf{Action:} {[call vision: Video 1 (0-218s), Video 2 (0-259s), Video 3 (0-170s), focus: "Describe the incorrect action in each video (e.g., detachment/abnormal operation), paying special attention to whether it involves the Detach or Attach actions mentioned in the options, and the names of the parts involved."]} \\
\textbf{Observation:} Visual Agent Return: Video 1: Detach wheel. The person incorrectly removes the wheel from the vehicle body. Video 2: Detach chassis. The person splits the vehicle body into two parts (the main body and the base/chassis), which is an incorrect detachment. Video 3: Attach cargo box. The person incorrectly installs the cargo box onto the vehicle body.

\textbf{Turn 2 (Final)} \\
\textbf{Thinking:} Based on the observations, both Video 1 and Video 2 contain a detachment action (Detach), while Video 3 involves an incorrect attachment. Option A is "detach the base and the chassis". The Visual Agent explicitly noted that the chassis was detached in Video 2. Video 1 and Video 2 are the only ones sharing the incorrect "Detach" action type. Based on the available options and the explicit identification of the chassis, Option A is the logical match for an incorrect detachment action occurring across the two videos. \\
\textbf{Action:} {[Answer]} \\
\textbf{Final Answer:} A
\end{dialoguebox}

\subsubsection{Dimension: Temporal Understanding}

\begin{dialoguebox}[Task: Plot Inference (PI)]
\textbf{Query:} \textit{Given the Beginning and Ending clips of a movie scene, infer what most likely happened in the middle.} \\
\textbf{Options:} A. The banker conspired with robbers to stage a fake heist. \quad B. Security systems malfunctioned during routine maintenance. \quad C. Robbers violently breached the vault and initiated a car chase. \quad D. The banker discovered embezzlement and confronted executives. \quad E. Protesters stormed the bank during a financial crisis. \quad F. An earthquake damaged vault security infrastructure. \\
\textbf{Ground Truth:} C

\textbf{Turn 1} \\
\textbf{Thinking:} Task initiation. I need to first get the captions for Video 1 (Beginning) to understand the premise of the story. \\
\textbf{Action:} {[call get\_caption: Video 1, Full]} \\
\textbf{Observation:} Audio Agent Return: The dialogue reveals a father talking to his son, threatening to kill "Dominic Toretto" because he "burn what is mine" regarding their "cash houses". This establishes a motive related to stolen/destroyed money and a clear conflict.

\textbf{Turn 2} \\
\textbf{Thinking:} Now I need to fetch the captions for Video 2 (Ending) to understand the narrative outcome. \\
\textbf{Action:} {[call get\_caption: Video 2, Full]} \\
\textbf{Observation:} Audio Agent Return: The captions contain repeated screams of "Oh, my God!" followed by dialogue about driving: "Find the line, feel the car, and let it fly." This suggests a high-stakes car-related sequence.

\textbf{Turn 3} \\
\textbf{Thinking:} The ending audio strongly suggests a car chase. I must visually verify the first 10 seconds of Video 2 to check for car wreckage, damage, or active pursuit. \\
\textbf{Action:} {[call vision: Video 2, 0-10s, focus: "Check for car wreckage, damage, or chase scenes. Note ground marks, smoke, or debris."]} \\
\textbf{Observation:} Visual Agent Return: The scene confirms a violent vehicle pursuit. There are dark SUVs, a large vehicle crashing into objects with smoke and debris, and a "POLICIA CIVIL" police car rolling over with severe body deformation. 

\textbf{Turn 4} \\
\textbf{Thinking:} The ending is a police car chase. I need to briefly check if Video 1 shows any robbers or vault visuals during the threat about burning property. \\
\textbf{Action:} {[call vision: Video 1, 90-105s, focus: "Check for visual clues of burned property, robbers, or cash houses."]} \\
\textbf{Observation:} Visual Agent Return: The scene is purely an indoor dialogue. No direct visual evidence of robbers, fire, or cash houses is shown yet.

\textbf{Turn 5 (Final)} \\
\textbf{Thinking:} Based on the collected evidence: Video 1 establishes a violent conflict involving stolen/burned cash houses. Video 2 depicts a destructive car chase involving dark SUVs (often used by robbers) and flipping police cars. The logical missing link between "threatening revenge over stolen cash" and an "intense police car chase" aligns perfectly with a violent heist and getaway. Options A, B, D, E, and F lack any visual or audio support (no protesters, no earthquake, no banker conspiracies). Option C is the only logical bridge connecting the plot. \\
\textbf{Action:} {[Answer]} \\
\textbf{Final Answer:} C
\end{dialoguebox}

\begin{dialoguebox}[Task: Functional Step Alignment (FSA)]
\textbf{Query:} \textit{Provide you two cooking videos, which step in Video 2 is functionally equivalent to the step shown between 74s and 89s in Video 1?} \\
\textbf{Ground Truth:} [86.0, 104.0]

\textbf{Turn 1} \\
\textbf{Thinking:} I need to first understand the overall cooking workflow of Video 1 (duration: 247.78s) to contextualize the steps. \\
\textbf{Action:} {[call vision: Video 1, 0-247.78s, focus: "Summarize the full cooking procedure. List the main steps in order."]} \\
\textbf{Observation:} Visual Agent Return: To summarize the full cooking procedure for Spring Rolls: 1. Gather Ingredients. 2. Cook Vermicelli Noodles. 3. Prepare Shrimp (peeling/slicing). 4. Prepare Char Siu Pork... 9. Assemble Spring Rolls. 10. Serve.

\textbf{Turn 2} \\
\textbf{Thinking:} Now I need to observe Video 2's overall workflow (294.71s) to confirm if it involves similar spring roll preparation and rolling actions. \\
\textbf{Action:} {[call vision: Video 2, 0-294.71s, focus: "Summarize the full cooking procedure. Does it involve rolling spring rolls or similar actions?"]} \\
\textbf{Observation:} Visual Agent Return: Step 1: Ingredient Preparation (boil noodles, chop herbs, prepare shrimp). Step 2: Assembling and Rolling Spring Rolls... Key Action: Yes, rolling spring rolls is a central action.

\textbf{Turn 3} \\
\textbf{Thinking:} I must observe Video 1's reference segment [74, 89]s to identify the specific target action. Based on the workflow, this might be the assembly or rolling step. \\
\textbf{Action:} {[call vision: Video 1, 74-89s, focus: "Describe the specific action in detail. Is it rolling spring rolls, assembling ingredients, or another step?"]} \\
\textbf{Observation:} Visual Agent Return: The specific action is preparing shrimp through two key steps: 1. Peeling shrimp. 2. Slicing peeled shrimp. This sequence is part of ingredient preparation, not rolling.

\textbf{Turn 4} \\
\textbf{Thinking:} The reference segment in Video 1 is about prepping shrimp. I will get the captions for this timeframe to find the exact verbal description. \\
\textbf{Action:} {[call get\_caption: Video 1, 74-89s]} \\
\textbf{Observation:} Audio Agent Return: [74.46s - 89.36s]: "Now that the shrimp has cooled down, peel them and cut them in half."

\textbf{Turn 5} \\
\textbf{Thinking:} The target action is "peel shrimp" and "cut them in half". I will search Video 2's full captions for these keywords to locate the corresponding step. \\
\textbf{Action:} {[call get\_caption: Video 2, Full]} \\
\textbf{Observation:} Audio Agent Return: ...[67.76s - 85.88s]: "of fresh cilantro, and two clean lettuce leaves." [85.88s - 105.24s]: "Now wash and peel eight large cooked shrimp and cut them in half." [105.24s - 109.60s]...

\textbf{Turn 6} \\
\textbf{Thinking:} The caption precisely matches the interval [85.88s - 105.24s] in Video 2. I will set a visual search window of 80-110s to pinpoint the exact start and end frame indices for this action. \\
\textbf{Action:} [call vision: Video 2, 80-110s, focus: "Locate the exact moment when the shrimp are peeled and cut in half. State the START frame index and END frame index."] \\
\textbf{Observation:} Visual Agent Return: START frame index: 19. END frame index: 40.

\textbf{Turn 7} \\
\textbf{Thinking:} To ensure completeness, I will expand the observation window slightly to 75-115s (using 64 frames) to capture all preparatory and follow-through motions. \\
\textbf{Action:} {[call vision: Video 2, 75-115s, focus: "Identify the full sequence of peeling and cutting shrimp. Start when picked up, end when fully cut and placed on the board."]} \\
\textbf{Observation:} Visual Agent Return: Peeling phase: Frames 21-36. Cutting phase: Frames 37-64. The full sequence spans frames 21 to 64.

\textbf{Turn 8 (Final)} \\
\textbf{Thinking:} Evidence collected. The observation window is 75s - 115s (40s span). Frame 21 corresponds to $\sim$85.88s, and Frame 64 corresponds to $\sim$105.24s. The alignment is fully confirmed by both audio and Visual Agents. \\
\textbf{Action:} {[Answer]} \\
\textbf{Final Answer:} [85.88, 105.24]
\end{dialoguebox}

\begin{dialoguebox}[Task: Procedural Step Sequencing (PSS)]
\textbf{Query:} \textit{Review these video clips. They are shuffled segments from a single cooking video. Please determine their correct chronological order.} \\
\textbf{Ground Truth:} 3->5->4->2->1

\textbf{Turn 1} \\
\textbf{Thinking:} Task started. I need to first get the captions for Video 1 to get a preliminary understanding of the action. \\
\textbf{Action:} {[call get\_caption: Video 1]} \\
\textbf{Observation:} Audio Agent Return: Mentions a little bit of feta cheese and that feta has a lot of salt in it.

\textbf{Turn 2} \\
\textbf{Thinking:} Continuing to get captions. Now I need to get the captions for Video 2 to understand its action. \\
\textbf{Action:} {[call get\_caption: Video 2]} \\
\textbf{Observation:} Audio Agent Return: Mentions going to use udon noodles, adding them to the pan, and letting the spinach wilt just a little bit.

\textbf{Turn 3} \\
\textbf{Thinking:} Continuing to get captions. Now I need to get the captions for Video 3 to understand its action. \\
\textbf{Action:} {[call get\_caption: Video 3]} \\
\textbf{Observation:} Audio Agent Return: Mentions putting noodles into some hot water.

\textbf{Turn 4} \\
\textbf{Thinking:} Continuing to get captions. Now I need to get the captions for Video 4 to understand its action. \\
\textbf{Action:} {[call get\_caption: Video 4]} \\
\textbf{Observation:} Audio Agent Return: Mentions adding fresh spinach.

\textbf{Turn 5} \\
\textbf{Thinking:} Continuing to get captions. Now I need to get the captions for Video 5 to understand its action. \\
\textbf{Action:} {[call get\_caption: Video 5]} \\
\textbf{Observation:} Audio Agent Return: Mentions adding onions, tomatoes, eggplant, and kalamata olives.

\textbf{Turn 6} \\
\textbf{Thinking:} Now I need to observe the overall visual of Video 3 to confirm its specific action. The caption mentions putting noodles into hot water, which might correspond to the boiling step. I need to confirm if it is the boiling process. \\
\textbf{Action:} {[call vision: Video 3, focus: "Describe the entire clip. Is the video showing noodles being boiled in water? Are the noodles raw and stiff, or are they already cooked and soft?"]} \\
\textbf{Observation:} Visual Agent Return: The video shows a person in a kitchen preparing noodles. The sequence begins with pouring raw and stiff noodles into a pot of boiling water. This is a standard step for cooking noodles.

\textbf{Turn 7} \\
\textbf{Thinking:} I need to observe the overall visual of Video 5 to confirm its action. The caption mentions adding onions, tomatoes, eggplant, etc., which might correspond to the sautéing step. I need to confirm the raw or cooked state of the ingredients. \\
\textbf{Action:} {[call vision: Video 5, focus: "Describe the entire clip. Are the onions, tomatoes, eggplant, and olives raw or cooked? Is the pan empty or filled with ingredients?"]} \\
\textbf{Observation:} Visual Agent Return: The video depicts a cooking process where the pan contains flames, indicating active cooking. Ingredients are being added to the pan. The ingredients are raw at the time of the clip. Cooked ingredients would show signs of browning or softening, which are not visible here. The pan is filled with ingredients, evidenced by flames and the chef stirring.

\textbf{Turn 8} \\
\textbf{Thinking:} I need to observe the overall visual of Video 2 to confirm its action. The caption mentions sautéing udon noodles and letting spinach wilt. I need to confirm if spinach is added during this sauteing process. \\
\textbf{Action:} {[call vision: Video 2, focus: "Describe the entire clip. Are the noodles already cooked, or are they being sautéed? Is the spinach added to the pan during this process?"]} \\
\textbf{Observation:} Visual Agent Return: The chef pours noodles into a pan containing spinach. The chef is stirring the noodles and spinach together. The noodles are being sautéed. Spinach is visible in the pan before the noodles are added. Thus, spinach was added to the pan before the noodles during this process.

\textbf{Turn 9} \\
\textbf{Thinking:} I need to observe the overall visual of Video 1 to confirm its action. The caption mentions cheese, parsley, and noodles, which might involve the plating or final seasoning step. I need to confirm if the ingredients are in their final state. \\
\textbf{Action:} {[call vision: Video 1, focus: "Describe the entire clip. Is the dish already plated and ready to eat? Are the ingredients cooked and final?"]} \\
\textbf{Observation:} Visual Agent Return: The entire clip shows the person cooking at a stove. There is no plating visible in any frame. Spinach is green and appears to be wilting but not fully cooked. Noodles are visible but show no signs of being fully soft or finished. A liquid is poured over the ingredients and stirred, indicating the dish is still in the cooking process. There is no sign of the dish being "done". The pan remains on the stove with visible heat. The dish is not plated. Ingredients are in the middle of cooking.

\textbf{Turn 10 (Final)} \\
\textbf{Thinking:} Based on the analysis: Video 3 shows boiling raw noodles in boiling water (preparation stage). Video 5 shows heating oil and adding raw onions, tomatoes, and other ingredients (start of sautéing). Video 4 shows adding raw spinach (a step during the sautéing process). Video 2 shows sautéing the noodles and spinach (mid-to-late cooking stage). Video 1 shows adding cheese and parsley for seasoning, but it is not plated yet (final step). Therefore, the final chronological sequence is 3->5->4->2->1. \\
\textbf{Action:} {[Answer]} \\
\textbf{Final Answer:} 3->5->4->2->1
\end{dialoguebox}

\subsubsection{Dimension: Multi-view Reasoning}

\begin{dialoguebox}[Task: Multi-view Spatial Reasoning (MSR)]
\textbf{Query:} \textit{When {A3} completely leaves view A, where is {B4} located in view B?}  \\
\textbf{Options:} A. bottom right of the frame , B. top right of the frame , C. top left of the frame , D. not shown in the frame  \\
\textbf{Ground Truth:} C 

\textbf{Turn 1} \\
\textbf{Thinking:} I need to determine where {B4} is located in View B when {A3} completely leaves View A. First, I must find the frame range where {A3} appears in View A to determine the frame it leaves. \\
\textbf{Action:} {[call vision: Video 1, Frames 1-500, focus: "Locate {A3} in View A (Video 1). Determine the frame range where it appears and identify when it completely leaves the frame."]}  \\
\textbf{Observation:} Visual Agent Return: {A3} appears in frames 1–10. It completely leaves the frame in frame 11.

\textbf{Turn 2} \\
\textbf{Thinking:} Now I need to determine the position of {B4} in View B when {A3} completely leaves View A (Frame 11). I need to check if {B4} appears in View B around Frame 11 and determine its specific location relative to road markings and other objects. \\
\textbf{Action:} {[call vision: Video 2, Frames 10-12, focus: "Locate {B4} in View B (Video 2). Check if it appears in Frame 11 and determine its position relative to road markings and other objects."]}  \\
\textbf{Observation:} Visual Agent Return: {B4} appears in Frame 11. It is located in the left lane of the highway, bounded by white lane dividers and a forward-pointing directional arrow. Relative to other objects, it is positioned to the right of a red truck (same lane) and to the left of a black car (adjacent lane), with trees and a green roadside area to its immediate left.

\textbf{Turn 3 (Final)} \\
\textbf{Thinking:} Based on the observations, {A3} appears in frames 1-10 of View A and completely leaves View A at frame 11. In frame 11 of View B, {B4} is located in the leftmost lane of the highway, with a black car to its right and a red truck to its left. Given the UAV perspective, the left lane corresponds to the left side of the frame. Based on the options, {B4} is located at the top left of the frame in View B. This matches option C. \\
\textbf{Action:} {[Answer]} \\
\textbf{Final Answer:} C
\end{dialoguebox}

\begin{dialoguebox}[Task:Multi-view Object Counting (MOC)]
\textbf{Query:} \textit{When {B1} fully appears in view B, how many green trucks are there in view A?}  \\
\textbf{Ground Truth:} B 

\textbf{Turn 1} \\
\textbf{Thinking:} I need to confirm the frame range where {B1} fully appears in View B, and synchronously check the number of green trucks in View A. First, I will locate the appearance frame range of {B1} in View B. \\
\textbf{Action:} {[call vision: Video 2, Frames 0-219, focus: "Locate {B1} in View B (Video 2). Identify the frames where it fully appears and note its attributes."]}  \\
\textbf{Observation:} Visual Agent Return: {B1} (the archway banner) fully appears in all 16 sampled frames (0-15).

\textbf{Turn 2} \\
\textbf{Thinking:} Now I need to check if there are green trucks in frames 0-15 of View A (Video 1). Since {B1} fully appears in frames 0-15 of View B, I must synchronously check the same frame range in View A, focusing on the number and color attributes of the trucks. \\
\textbf{Action:} {[call vision: Video 1, Frames 0-15, focus: "Inspect View A (Video 1) frames 0-15 for green trucks. Count the number of green trucks visible in the BBox area and confirm their color attributes."]}  \\
\textbf{Observation:} Visual Agent Return: In all 16 frames (0-15), a single green truck (with a green cab) is consistently visible in the upper-middle section of the road. The green truck’s cab is clearly green in all frames, satisfying the color requirement. There is 1 green truck visible in the BBox area.

\textbf{Turn 3 (Final)} \\
\textbf{Thinking:} Based on the observations, {B1} fully appears in frames 0-15 of View B, which corresponds to the same frame range in View A. In View A's frames 0-15, there is exactly 1 green truck present with clear color attributes, and no other green trucks or occlusions. The final answer is B. \\
\textbf{Action:} {[Answer]} \\
\textbf{Final Answer:} B
\end{dialoguebox}

\subsubsection{Dimension: Free-form QA}

\begin{dialoguebox}[Task: Comparative Culinary QA (CCQA)]
\textbf{Query:} \textit{How do the main ingredients used in the batter mixture differ between Video A and Video B?}  \\
\textbf{Ground Truth:} Base liquid difference (club soda vs. beer), Flour composition (corn flour inclusion vs. plain flour only), Spice selection (chili/oregano vs. paprika/garlic/sugar). 

\textbf{Turn 1} \\
\textbf{Thinking:} First, I need to understand the verbal descriptions in both videos to identify potential ingredient differences.  Let me retrieve the full captions for both videos.  \\
\textbf{Action:} {[call get\_caption: Video 1]}  \\
\textbf{Observation:} Audio Agent Return: Mentions All-Purpose Flour, Corn Flour, Red Chilli Powder, dry Oregano, Salt, Garlic, and Club Soda. 

\textbf{Turn 2} \\
\textbf{Thinking:} Now I need to retrieve the full captions for Video 2 to compare ingredient lists. This will help identify potential differences in batter components.  \\
\textbf{Action:} {[call get\_caption: Video 2]}  \\
\textbf{Observation:} Audio Agent Return: Mentions plain or all-purpose flour, beer, paprika, sugar, garlic powder, salt, and black pepper. 

\textbf{Turn 3} \\
\textbf{Thinking:} From captions, Video A uses All-Purpose Flour, Corn Flour, Red Chilli Powder, Oregano, Salt, Garlic, and Club Soda.  Video B uses All-Purpose Flour, Beer, Paprika, Sugar, Garlic Powder, Salt, and Black Pepper. Now I need to verify if these ingredients are actually shown in the videos through visual observation.  \\
\textbf{Action:} {[call vision: Videos 1 \& 2, focus: "Verify ingredient visuals: Check if corn flour, red chilli powder... are actually shown in the videos."]} \\
\textbf{Observation:} Visual Agent Return: For Video A, corn flour, red chilli powder, oregano, and club soda are shown; garlic is not. For Video B, beer, paprika, sugar, garlic powder, and black pepper are all present. 

\textbf{Turn 4} \\
\textbf{Thinking:} Based on Turn 3's visual verification, Video A has corn flour, red chilli powder, oregano, club soda but no garlic. Video B has beer, paprika, sugar, garlic powder, black pepper.Now I need to check if there are other ingredient differences in the batter preparation process.  \\
\textbf{Action:} {[call vision: Videos 1 \& 2, focus: "Check for additional ingredient differences: Are there any other dry ingredients or liquids added..."]}  \\
\textbf{Observation:} Visual Agent Return: Video 1 includes only one dry ingredient (flour) and no liquids added to the batter during the observed segment. Video 2 includes additional dry ingredients (spices, salt) and additional liquids (vinegar, oil, beer) beyond the initial flour. 

\textbf{Turn 5 (Final)} \\
\textbf{Thinking:} After 4 rounds of observation, I've confirmed key ingredient differences. Video A uses corn flour, red chilli powder, oregano, and club soda. [ Video B uses beer, paprika, sugar, garlic powder, and black pepper.  Video B also includes additional liquids (vinegar, oil) and more spice varieties.  \\
\textbf{Action:} {[Answer]} \\
\textbf{Final Answer:} 
1. Ingredients: Video A uses All-Purpose Flour, Corn Flour, Red Chilli Powder, Oregano, Salt, Club Soda (no garlic). Video B uses All-Purpose Flour, Beer, Paprika, Sugar, Garlic Powder, Black Pepper. 
2. Key Differences: Video A uses corn flour and club soda. Video B uses beer, paprika, sugar, and black pepper, plus additional liquids. 
3. Flavor Profile: Video A emphasizes heat and herbiness, while Video B focuses on beer aroma, sweetness, and complexity. 
4. Preparation: Video A adds spices directly to dry flour mix; Video B incorporates beer as a liquid base with multiple spice additions. 
\end{dialoguebox}

\subsection{Failure Cases}\label{failure_case}

While AgentCVR demonstrates robust performance across various cross-video tasks, analyzing its failure cases provides critical insights into the current boundaries of LLM-based multi-agent systems. 
In this section, we categorize and examine typical failure modes encountered during real-world inference. 
These failures generally stem from three distinct challenges: (1) \textit{instruction drift} leading to formatting errors during extended context reasoning, (2) \textit{premature cognitive closure} or exclusivity bias in multi-label reasoning tasks, and (3) the \textit{over-reliance on internal parametric priors} (common sense) that overrides specific visual evidence. By investigating these limitations, we highlight critical areas for future improvements in autonomous video understanding.

\begin{dialoguebox}[Formatting Errors]
\textbf{Query:} \textit{How do the tools used for breading the meat differ between video A and video B?}  \\
\textbf{Ground Truth:} Meat handling tools (hands vs fork), Pounding equipment (none vs mallet+pot), Breading transfer method (direct press vs fork transfer).

\textbf{Turn 1} \\
\textbf{Thinking:} First I need to understand the overall cooking process and identify when the breading step occurs.  \\
\textbf{Action:} Answer{[call get\_caption: Video 1]} \\
\textbf{Observation:} Audio Agent Return: Mentions preparing wiener schnitzel and the steps involving flour, egg mixture, and breadcrumbs. 

\textbf{Turn 2} \\
\textbf{Thinking:} Now I need to understand Video 2's content to locate the breading step.  The caption will show if they use different tools like egg wash, breadcrumbs, or unique equipment for coating.  \\
\textbf{Action:} Answer{[call get\_caption: Video 2]} \\
\textbf{Observation:} Audio Agent Return: Mentions a breading station with flour, beaten egg, and breadcrumbs, explicitly advising to use a fork to avoid breading your fingers. 

\textbf{Turn 3} \\
\textbf{Thinking:} From the captions, both videos use flour, egg wash, and breadcrumbs for breading.  I'll first observe Video A's breading process to identify tools used.  \\
\textbf{Action:} Answer{[call vision: Video 1, focus: "Describe the tools used for breading in Video A. Specifically identify: 1) Type of bowl... 2) Any tools... 3) Method..."]} \\
\textbf{Observation:} Visual Agent Return: Large metal mixing bowls are used.  The person uses their hands to mix ingredients and applies breadcrumbs by pressing them onto the food.  No whisk or fork is visible.

\textbf{Turn 4} \\
\textbf{Thinking:} Now I need to analyze Video B's breading process.  The caption mentions using a fork to avoid 'bread your fingers,' suggesting a different tool than Video A's hand-only method.  I'll focus on the breading tools in Video B.  \\
\textbf{Action:} Answer{[call vision: Video 2, focus: "Describe Video B's breading tools..."]} \\
\textbf{Observation:} Visual Agent Return: Rectangular metal trays are used. A fork is used to handle the meat during breading, and hands are used to press the meat into the breadcrumbs. 

\textbf{Turn 5 (Format Error \& Auto-Recovery)} \\
\textbf{Thinking:} Both videos use pressing for breadcrumbs, but Video A's caption mentions 'three eggs' while Video B specifies 'two eggs' - this might affect the egg wash consistency.  I need to check if they use different tools for mixing the egg wash (e.g., fork vs whisk).  \\
\textbf{Action:} Answer{[Malformed JSON Output]} \\
\textbf{System Intervention:} \textit{!Format issue on attempt 1. Retrying cleanly... (LLM response is not valid JSON.)}  \textit{ Attempt 2: Sending gentle Chinese reminder...}  \\
\textbf{Action (Retry):} Answer{[call vision: Video 1, focus: "Describe Video A's egg wash preparation..."]} \\
\textbf{Observation:} Visual Agent Return: The egg wash is prepared in large metal mixing bowls. The person uses hands to mix the egg wash. 

\textbf{Turn 6} \\
\textbf{Thinking:} Now I need to check Video B's egg wash preparation.  The caption mentions using a fork.  I'll focus on their egg wash process.  \\
\textbf{Action:} Answer{[call vision: Video 2, focus: "Describe Video B's egg wash preparation..."]} \\
\textbf{Observation:} Visual Agent Return: The egg wash is contained in a rectangular metal tray.  A fork is used to stir the egg wash. 

\textbf{Turn 7 (Final)} \\
\textbf{Thinking:} After 6 rounds of observation, I've collected sufficient evidence to compare the breading tools between Video A and Video B.  The key differences lie in container types, mixing tools, and egg wash preparation methods.  \\
\textbf{Action:} {[Answer]} \\
\textbf{Final Answer:} 
1. Breading Containers: Video A uses large metal mixing bowls; Video B uses rectangular metal trays. 
2. Mixing Tools: Video A uses hands only;  Video B uses a fork to stir egg wash and hands for pressing breadcrumbs. 
3. Egg Wash Preparation: Video A mixes beaten eggs by hands; Video B uses a fork in a metal tray. 
4. Breadcrumb Application: Both use the pressing method, but Video B explicitly mentions using a fork. 
\end{dialoguebox}

\textit{Failure Analysis Note: This case highlights the vulnerability of LLMs to "instruction drift" during extended multi-turn interactions. As the context window fills with lengthy multimodal observations, the Master Agent occasionally loses its strict JSON formatting constraints. However, this example also effectively demonstrates the necessity and robustness of our system-level auto-recovery mechanism. By detecting the malformed output and injecting a targeted prompt intervention, the system successfully nudges the agent back on track without causing a catastrophic failure of the entire reasoning chain.}

\begin{dialoguebox}[Incomplete Answer Identification]
\textbf{Query:} \textit{Which animals' current behaviors make them more likely to be detected by predators that hunt using visual movement cues?} \\
\textbf{Options:} A. Video one. B. Video two. C. Video three. D. Video four. \\
\textbf{Ground Truth:} A, B, C 

\textbf{Turn 1} \\
\textbf{Thinking:} I need to scan Video 1 to confirm if the animal is stationary or moving regularly (e.g., foraging), as such behaviors might reduce the risk of predation due to a lack of suddenness. Frequent head turning or sudden sprints might expose it. \\
\textbf{Action:} Answer{[call vision: Video 1, focus: "Identify if the animal is stationary, slowly moving, or exhibiting sudden movements. Note any head-turning, tail-twitching, or abrupt directional changes."]} \\
\textbf{Observation:} Visual Agent Return: The walruses are not entirely stationary, as their positions shift slightly. They are slowly moving with gradual shifts on the rocky shore. There are no rapid or jerky motions , and no sudden movements or abrupt directional changes are observed. Water splashes are smooth and incremental. Head-turning is a slow, deliberate action. They lack prominent tails. The conclusion is that they are slowly moving.

\textbf{Turn 2} \\
\textbf{Thinking:} I need to scan Video 2. Since the video is only 1.83 seconds long, I must quickly identify key behaviors. Stationary or regular actions (like slow chewing) are less likely to expose the animal, whereas sudden actions (like fast escape or rapid blinking) make it easier for visual predators to lock on. \\
\textbf{Action:} Answer{[call vision: Video 2, focus: "Is the animal stationary or moving? Are there sudden movements, rapid eye blinks, or erratic motions? Note any abrupt changes in posture or direction."]} \\
\textbf{Observation:} Visual Agent Return: The shrimp-like creature is moving. There are subtle changes in its posture, limb positioning, and antennae orientation. The motion is slow and incremental. There are no sudden movements, and changes are gradual and smooth. Rapid eye blinks are irrelevant as it uses antennae for sensing. There are no erratic motions. The conclusion is that the animal is moving gradually and non-erratically.

\textbf{Turn 3} \\
\textbf{Thinking:} I need to scan Video 3, which is 4.83 seconds long, focusing on any sudden actions or regular movement. Fast running, frequent head turning, or sudden steering significantly increase the probability of discovery by visual predators. \\
\textbf{Action:} Answer{[call vision: Video 3, focus: "Detect any sudden movements, rapid directional changes, or frequent head-turning. Are the animal's motions smooth and predictable, or are there abrupt, erratic actions?"]}  \\
\textbf{Observation:} Visual Agent Return: The crab-like arthropod maintains a steady forward trajectory with no abrupt shifts in direction. It progresses smoothly along the mossy surface with a rhythmic, repetitive gait. Each step is gradual without jerky actions. The head and antennae show minimal, slow adjustments. The conclusion is that its motions are smooth and predictable.

\textbf{Turn 4} \\
\textbf{Thinking:} I need to scan Video 4, which is 4.625 seconds long, checking for sudden actions or regular movement. Behaviors like fast running, frequent head turning, or sudden steering increase the risk of being spotted. \\
\textbf{Action:} Answer{[call vision: Video 4, focus: "Detect any sudden movements, rapid directional changes, or frequent head-turning. Are the animal's motions smooth and predictable, or are there abrupt, erratic actions?"]}  \\
\textbf{Observation:} Visual Agent Return: The animal, likely a wild boar in a forest , remains largely consistent in position with no abrupt shifts or rapid directional changes. Its head is mostly lowered for foraging, with slow and gradual minor adjustments. Its movements are smooth and predictable, maintaining a steady posture with subtle, incremental motions. No sudden jumps or rapid head-turning are observed.

\textbf{Turn 5 (Final)} \\
\textbf{Thinking:} Based on the analysis:
1. Video 1 (Walrus): Moving slowly, no sudden movements, not easily detected.
2. Video 2 (Shrimp): Moving slowly, but limbs are adjusting frequently, which might attract predator attention due to regular actions.
3. Video 3 (Crab): Steady direction, repetitive actions, highly predictable, which might lower the risk of detection.
4. Video 4 (Boar): Mostly stationary, only slow foraging actions, not easily exposed.
Conclusion: None of the animals exhibit obvious sudden actions, but the shrimp in Video 2 is easier to detect by visual predators due to frequent limb adjustments. \\
\textbf{Action:} {[Answer]} \\
\textbf{Final Answer:} B 
\end{dialoguebox}

\textit{Failure Analysis Note: Although the Master Agent accurately perceived the subtle visual cues (movement vs. stationary) across all four videos, it failed at the final reasoning stage due to "Premature Cognitive Closure." The model exhibited an exclusivity bias—treating a multiple-choice question (where multiple answers could be correct) as a single-best-choice question. It erroneously concluded that only the most active subject (Video 2) was the answer, ignoring Videos 1 and 3 which also met the detection criteria. This reveals a critical gap between accurate low-level multimodal perception and complex high-level logical alignment in multi-label scenarios.}

\begin{dialoguebox}[Incorrect Answer]
\textbf{Query:} \textit{Review these video clips. They are shuffled segments from a single cooking video. Please determine their correct chronological order.} \\
\textbf{Ground Truth:} 5->3->4->2->1 \\
\textbf{Model Prediction:} 3->5->4->2->1 

\textbf{Turn 1} \\
\textbf{Thinking:} Task started. I need to first get the captions for Video 1 to get a preliminary understanding of the action. \\
\textbf{Action:} Answer{[call get\_caption: Video 1]} \\
\textbf{Observation:} Audio Agent Return: Mentions adding watercress, putting udon on a platter, and adding garnishes like carrots, green onions, and cilantro.

\textbf{Turn 2} \\
\textbf{Thinking:} Continuing to get captions. Now I need to get the captions for Video 2 to understand its action. \\
\textbf{Action:} Answer{[call get\_caption: Video 2]} \\
\textbf{Observation:} Audio Agent Return: Mentions adding a sauce mixture, taking blanched udon noodles, and lightly stir-frying them.

\textbf{Turn 3} \\
\textbf{Thinking:} Continuing to get captions. Now I need to get the captions for Video 3 to understand its action. \\
\textbf{Action:} Answer{[call get\_caption: Video 3]} \\
\textbf{Observation:} Audio Agent Return: Mentions making a quick dressing using soy sauce, oyster sauce, ponzu, red miso paste, and sake, and dissolving it.

\textbf{Turn 4} \\
\textbf{Thinking:} Continuing to get captions. Now I need to get the captions for Video 4 to understand its action. \\
\textbf{Action:} Answer{[call get\_caption: Video 4]} \\
\textbf{Observation:} Audio Agent Return: Mentions adding red pepper, onions, more oil, ginger, garlic, and green onion.

\textbf{Turn 5} \\
\textbf{Thinking:} Continuing to get captions. Finally, get the captions for Video 5 to understand its action. \\
\textbf{Action:} Answer{[call get\_caption: Video 5]} \\
\textbf{Observation:} Audio Agent Return: Mentions heating a skillet, adding oil, and cooking shiitake mushrooms on high heat.

\textbf{Turn 6} \\
\textbf{Thinking:} Now I need to observe the overall visual of Video 1 to confirm the plating action and the final state of the ingredients. I will check for any obviously unfinished actions or leftover ingredients. \\
\textbf{Action:} Answer{[call vision: Video 1, focus: "Describe the final state of the dish..."]} \\
\textbf{Observation:} Visual Agent Return: The udon is fully cooked. Watercress and other ingredients are arranged on the platter. Raw watercress is visible in the final plated dish.

\textbf{Turn 7} \\
\textbf{Thinking:} I need to observe the overall visual of Video 2 to confirm whether the udon noodles are already cooked and in the stir-frying stage. I'll focus on the noodle state and cooking actions. \\
\textbf{Action:} Answer{[call vision: Video 2, focus: "Describe the state of the udon noodles..."]} \\
\textbf{Observation:} Visual Agent Return: The udon noodles are fully cooked (blanched) and are being stir-fried in a preheated pan with the sauce and vegetables.

\textbf{Turn 8} \\
\textbf{Thinking:} I need to observe the overall visual of Video 3 to confirm the sauce preparation process. I'll focus on mixing and dissolving actions, and the state of the ingredients. \\
\textbf{Action:} Answer{[call vision: Video 3, focus: "Describe the preparation of the sauce..."]} \\
\textbf{Observation:} Visual Agent Return: The ingredients are mixed together, the miso is dissolved, and there are no visible raw or unprocessed ingredients in the sauce preparation.

\textbf{Turn 9} \\
\textbf{Thinking:} I need to observe the overall visual of Video 4 to confirm the processing of red peppers and onions. I'll check if they are chopped, being sautéed, and mixed with sauce. \\
\textbf{Action:} Answer{[call vision: Video 4, focus: "Describe the processing of red peppers and onions..."]} \\
\textbf{Observation:} Visual Agent Return: The red peppers and onions are sliced, being sautéed in a pan, and interacting with oil/sauce during the cooking process.

\textbf{Turn 10} \\
\textbf{Thinking:} I need to observe the overall visual of Video 5 to confirm the heating and initial cooking steps. I'll check if oil is heated, the order of adding ingredients, and if it logically connects to subsequent steps. \\
\textbf{Action:} Answer{[call vision: Video 5, focus: "Describe the initial cooking steps..."]} \\
\textbf{Observation:} Visual Agent Return: The pan is heated with oil. Oil is poured first, then mushrooms are added. There is no visible interaction with the sauce or ingredients from Video 4.

\textbf{Turn 11 (Final)} \\
\textbf{Thinking:} Based on the analysis: Video 3 prepares the seasoning (mixing raw sauces), which is an early stage. Video 5 heats oil and sautés mushrooms (initial cooking). Video 4 sautés red peppers and onions, which should follow mushrooms to utilize the oil. Video 2 stir-fries udon (mixed cooked noodles + sauce) which follows the vegetable sautéing. Video 1 is plating (final step). The logical chain is: Seasoning (3) -> Heat/Mushrooms (5) -> Sauté veggies (4) -> Stir-fry noodles (2) -> Plating (1). \\
\textbf{Action:} {[Answer]} \\
\textbf{Final Answer:} 3->5->4->2->1
\end{dialoguebox}

\textit{Failure Analysis Note: The model failed due to a rigid assumption about standard cooking workflows. It incorrectly assumed that all prep work (making the sauce in Video 3) must chronologically precede the heating of the pan (Video 5). In the ground truth, the chef first heats the oil and mushrooms (Video 5), takes a moment to mix the sauce while the pan heats (Video 3), then returns to the pan to add peppers (Video 4), noodles and sauce (Video 2), and finally plates the dish (Video 1). The model missed the visual and chronological continuity between the steps, relying too heavily on general procedural common sense rather than the specific continuity cues between the video clips.}

\section{Prompt Summary}\label{apd_prompt}

\subsection{Prompts for Script Synthesis (\texorpdfstring{$\mathcal{W}_{script}$}{W\_script})}

In this section, we detail the prompt templates used to drive LLMs to synthesize video scripts for various tasks. 
To ensure the complexity, visual density, and logical rigor of the generated data, we designed tailored prompt structures for different CVR tasks.

\subsubsection{Multi-view Spatial Reasoning (MSR)}
For the Multi-view Spatial Reasoning task, the prompt strictly mandates the introduction of ``Environmental Occlusions'' to ensure that the scripts generated for the two views are complementary in terms of visual information.

\begin{promptbox}[ MSR Prompt]
\#\#\# Role\\
You are the **Lead Data Simulator for the MSR (Multi-view Spatial Reasoning) Benchmark**.\\
Your task is to generate high-difficulty synthetic data that tests if an AI can spatially reason across two synchronized camera views despite visual obstructions.

\#\#\# Scenario Settings\\
* **Scene**: \textless scenario\textgreater \\
* **Traffic Density**: \textless density\textgreater \\
* **Camera Configuration**: \textless camera\_desc\textgreater 
* **Total Duration**: \textless duration\textgreater seconds. \\
* **Sampling Rate**: Detailed description every 2 seconds. \\

\#\#\# CRITICAL INSTRUCTION: OCCLUSION \& HARD MODE\\
To make this task challenging, you MUST introduce **Environmental Occlusions**:\\
1. **Define Obstacles**: Explicitly state what blocks the view (e.g., "A large oak tree covers the NW corner of View A", "A parked delivery truck blocks the lower lane in View B").\\
2. **The "Hidden" Entity**: Ensure at least one key object passes BEHIND an obstacle in one view but remains visible in the other.\\
3. **Entity Consistency**: Use \{A1\}=\{B1\}, \{A2\}=\{B2\} mapping.\\

\#\#\# Task 1: Generate Synchronized Video Scripts\\
Generate a timeline (0s to \textless duration\textgreater s) with a step of 2 seconds.\\
For EACH timestamp, provide:\\
* **View A Visual**: Describe positions, relative distances, and **occlusions**.\\
* **View B Visual**: Describe the same reality from the other angle.\\

\#\#\# Task 2: Generate 1 Complex MSR Question\\
Create ONE single-choice question focused on **Spatial Reasoning**.\\
* **Focus**: Relative Position, Trajectory Prediction, or Blind Spot Inference.\\
* **Requirement**: The answer MUST derive from understanding **BOTH** videos.\\

\#\#\# Output JSON Format\\
Strict JSON format specifying meta, video\_scripts, and question with reasoning
\end{promptbox}

\subsubsection{Multi-view Object Counting (MOC)}
The prompt for the Multi-view Object Counting task focuses on cross-view object deduplication and the design of distractors to test the model's tracking and counting capabilities in complex scenarios.

\begin{promptbox}[ MOC Prompt]
\#\#\# Role\\
You are the **Lead Data Simulator for the MOC (Multi-view Object Counting) Benchmark**.\\
Your task is to generate high-difficulty synthetic data that tests if an AI can correctly count objects across synchronized videos, handling duplicates and occlusions.\\

\#\#\# Scenario Settings\\
* **Scene**: \textless scenario\textgreater\\
* **Traffic Density**: \textless density\textgreater\\
* **Camera Configuration**: \textless camera\_desc\textgreater\\
* **Total Duration**: \textless duration\textgreater seconds.\\

\#\#\# CRITICAL INSTRUCTION: COUNTING COMPLEXITY \& OCCLUSION\\
To make this challenging:\\
1. **Occlusion Zones**: Define zones where objects vanish in one view but are visible in the other.\\
2. **Cross-View Transition**: Objects must move from View A's field of view into View B's.\\
3. **Distractors**: Include objects that SHOULD NOT be counted (e.g., "Count only Red Cars", but include Red Trucks and Blue Cars).\\
4. **Deduplication**: \{A1\} and \{B1\} are the SAME physical car. The AI must count it as 1, not 2.\\

\#\#\# Task 1: Generate Synchronized Video Scripts\\
Generate a timeline with a step of 2 seconds. Provide View A Visual, View B Visual, and a Tracking Note (e.g., "\{A1\} is now \{B1\}").\\

\#\#\# Task 2: Generate 1 Complex MOC Question\\
Create ONE single-choice question focused on **Object Counting** (Global Unique Count or Conditional Count).\\

\#\#\# Output JSON Format\\
Strict JSON format specifying meta, video\_scripts, and question
\end{promptbox}

\subsubsection{Process Sorting}
This prompt is utilized to generate unordered segments of continuous processes, such as cooking, and requires the model to provide dense visual detail descriptions for subsequent use in the temporal sorting task.

\begin{promptbox}[ Sorting Prompt]
\#\#\# Role\\
You are the **Director of a High-Precision Video Reasoning Benchmark**.
Your goal is to generate a **chronologically ordered** cooking script for "\textless recipe\textgreater".\\

\#\#\# Strict Configuration\\
* **Total Phases**: Exactly **\textless num\_steps\textgreater**\\
* **Difficulty**: \textless difficulty\textgreater\\
* **Visual Focus**: \textless focus\textgreater\\

\#\#\# Task\\
Break down the cooking process into **\textless num\_steps\textgreater Distinct Sequential Phases**.\\
Inside EACH Phase, list **8-15 Atomic Visual Events** (Micro-Actions).\\

\#\#\# Critical Requirements\\
1. **Strict Structure**: You MUST generate exactly \textless num\_steps\textgreater phases.\\
2. **Focus Adherence**: Your visual descriptions MUST emphasize "\textless focus\textgreater".\\
3. **Logical Flow**: The phases must follow the correct timeline (Irreversible logic).\\
4. **Atomic Detail**: Do NOT write "He cooks the meat". Break it down: "Places pan", "Meat hits pan", "Searing sound", "Flipping meat".\\
5. **Dense Visuals**: Describe the **texture, color, and consistency** of the food at that exact moment. Show progression explicitly.\\

\#\#\# Output JSON Format\\
Strict JSON format with phases, events, visuals, and captions
\end{promptbox}

\subsubsection{Plot Inference (Missing Middle)}
This task requires the generation of a narrative structure containing the beginning and ending segments but with a missing middle segment, in order to test the model's causal inference abilities.

\begin{promptbox}[ Plot Inference Prompt]
\#\#\# Role\\
You are a **Screenwriter**.\\
Task: Create a **Plot Inference (Missing Middle)** challenge.\\
* **Genre**: \textless genre\textgreater\\
* **Setting**: \textless setting\textgreater\\
* **Core Conflict (The Missing Event)**: \textless conflict\textgreater\\

\#\#\# Structure\\
* **Act 1 (Beginning)**: Setup.\\
* **Act 2 (HIDDEN Middle)**: Duration \textless gap\_duration\textgreater seconds. The "\textless conflict\textgreater" happens here. DO NOT write the script, but ensure Act 3 reflects its consequences.\\
* **Act 3 (Ending)**: Aftermath. Visual state must change drastically due to Act 2.\\

\#\#\# INSTRUCTION: FILL THE TIMELINE\\
I have pre-calculated the time slots. **You must generate 'visual' and 'caption' for EACH slot.**\\

**Act 1 Slots (Beginning):**\\
\textless begin\_slots\_str\textgreater\\
**Act 3 Slots (Ending):**\\
\textless end\_slots\_str\textgreater\\

\#\#\# Content Requirements (Strict)\\
1. **Visual**: Focus ONLY on Narrative Content (Characters, Actions, Environment, Plot). NO camera/lighting jargon.\\
2. **Caption**: Dialogue ONLY. Empty if silent.\\

\#\#\# Option Generation (Crucial)\\
Generate 6 distinct plot summaries for the missing Act 2. One Correct Answer, 5 plausible but incorrect Distractors.\\

\#\#\# Output JSON Format\\
Strict JSON format with question, options, correct\_answer, and scripts
\end{promptbox}

\subsubsection{Movie Understanding (Hard Single-Choice)}
Aimed at the retrieval and understanding of specific clues in long videos, this requires the generation of four video scripts that are highly similar in setting and action but differ in critical details.

\begin{promptbox}[ Movie Understanding Prompt]
\#\#\# Role\\
You are a **Benchmark Dataset Designer**.\\
Your task is to create a **Hard Single-Choice Video Understanding Problem**.\\
Genre: **"\textless genre\textgreater"**.\\

\#\#\# Input Elements (The Target Scenario)\\
The **Correct Answer** must depict this specific scenario:\\
\textless keywords\_str\textgreater\\

\#\#\# CRITICAL CONSTRAINT: THE LOGIC PUZZLE (Strict Single-Choice)\\
You are generating 4 different video clips (A, B, C, D), but **ONLY ONE** is the correct answer.\\
1. **Target Video**: Must perfectly integrate ALL the Input Elements into the plot.\\
2. **Distractor Videos**: Must belong to the same Genre and style. They should be confusingly similar (e.g., same setting but different action), but MUST FAIL to match the full description.\\

\#\#\# VISUAL \& FORMAT CONSTRAINTS\\
1. **Independent Durations**: \textless video\_instructions\textgreater\\
2. **Variable Pacing**: Use natural editing (Short Cuts 3-8s + Long Takes 15-30s).\\
3. **Content Requirements**: A natural, descriptive sentence of what is seen (Action + Camera + Environment) and subtitles/sound effects.\\

\#\#\# Output JSON Format\\
Strict JSON format with logic\_adaptation, question, correct\_answer, and timelines for 4 videos
\end{promptbox}

\subsubsection{Video Grounding / Alignment}
This is used to generate two parallel video scripts with subtle visual or action differences, to test spatial and temporal alignment as well as grounding capabilities.

\begin{promptbox}[ Grounding Prompt]
\#\#\# Role\\
You are the **Chief Editor of a Video Alignment Benchmark**.\\
Your goal is to generate a "Parallel Script" for two different videos (Video A and Video B) cooking the SAME recipe: "\textless recipe\textgreater".\\

\#\#\# Diversity Injection\\
* **Alignment Granularity**: \textless granularity\textgreater\\
* **Variation Style**: \textless variation\textgreater\\
* **Difficulty**: \textless difficulty\textgreater\\

\#\#\# Task\\
Break the cooking process into **\textless num\_steps\textgreater Distinct Functional Steps**.\\
For EACH step, generate TWO variations (Video A vs Video B) that perform the **SAME FUNCTION** but look **DIFFERENT**.\\

\#\#\# Critical Requirements\\
1. **Enforce Variation**: You MUST apply the "\textless variation\textgreater" rule rigorously.\\
2. **Enforce Granularity**: The descriptions must focus on the "\textless granularity\textgreater" level.\\
3. **Dense Visuals**: Describe texture, color, and consistency. Show progression.\\
4. **Micro-Events**: For each step, list 3-8 micro-events for Video A and 3-8 for Video B.\\

\#\#\# Output JSON Format\\
Strict JSON format with alignment\_steps, function\_desc, and dual video events
\end{promptbox}

\subsubsection{Cooking Action (Hard Negative)}
Generates dense scripts containing atomic-level micro-actions, used to distinguish highly similar distractor videos.

\begin{promptbox}[ Cooking Prompt]
\#\#\# Role\\
You are the **Director of a High-Precision Video Reasoning Benchmark**. Your goal is to generate "Hard Negative" training data that mimics real-world cooking videos.\\

\#\#\# Task Assignment\\
- **Recipe**: \textless recipe\textgreater\\
- **Focus**: \textless difference\_type\textgreater\\
- **Target**: \textless config\_label\textgreater\\
- **Style**: \textless config\_style\textgreater\\

\#\#\# 1. The Question (Single-Choice)\\
Create a difficult **Single-Choice Question** where Option \textless config\_letter\textgreater is the ONLY correct answer based on visual details or caption nuances.\\

\#\#\# 2. Video Scripts (CRITICAL: HYPER-GRANULARITY \& VISUAL DENSITY)
* **Atomic Action Decomposition (3-8s per segment)**: Break actions down into micro-steps.\\
* **Dense Visuals**: Describe the State (texture, color) and Show Progression explicitly. Include Visual Noise (tools, steam).\\
* **The Logical Trap**: The Target must contain the specific evidence. Distractors must look 90\% similar but fail on the specific detail required.\\

\#\#\# Output JSON Format\\
Strict JSON format with multiple video timelines mapping dense visuals
\end{promptbox}

\subsubsection{Behavior and Intent Understanding}
By defining different intents and context modifiers, this generates action scripts that capture complex differences in human or animal intentions.

\begin{promptbox}[ Behavior Prompt]
\#\#\# Role\\
You are an **Expert Examiner** creating a Behavior Understanding dataset.\\

\#\#\# Task Specs\\
* **Type**: \textless task\_type\textgreater\\
* **Target Behavior/Activity**: "\textless target\_activity\textgreater"\\
* **Context/Nuance**: "\textless context\_modifier\textgreater"\\
* **Distractor Candidates**: \textless distractors\textgreater\\

\#\#\# Step 1: Scenario Expansion\\
1. **Target Scenario**: Describe the target specifically performed in the required manner.\\
2. **Distractor Scenarios**: Choose behaviors that fit the same Subject/Setting but imply a **different intent**.\\

\#\#\# Step 2: Formulate Question (Intent-Based)\\
Write a concise question focusing on the **Specific Intent** or **Nuance**. (e.g., "Which video depicts a lion resting due to exhaustion rather than just sleeping?")\\

\#\#\# Step 3: Script Generation\\
Generate 4 videos. Exactly \textless num\_correct\textgreater Videos must depict the Target Scenario. The others must be Distractors. Provide dense visual descriptions with no captions.\\

\#\#\# Output JSON Format\\
Strict JSON format with expanded\_scenario, question, correct indices, and video content
\end{promptbox}

\subsubsection{Assembly Task Error Identification}
Based on specific assembly operations, performers are assigned different personas to generate action scripts containing specific assembly errors.

\begin{promptbox}[ Assembly Prompt]
\#\#\# Role\\
You are an expert scriptwriter for Video Understanding Benchmarks.\\
**Task:** Generate detailed visual scripts for 3 videos of different people assembling the same toy.\\

**Context:**\\
- Toy: \textless toy\_id\textgreater\\
- Parts Sequence (Standard SOP): [\textless parts\_str\textgreater]\\
- **Goal:** In ALL videos, the user MUST eventually complete the FULL assembly.\\

**SCRIPTING RULES (STRICT):**\\
1. **Full Process**: Start from loose parts -\textgreater Finish with the completed toy.\\
2. **Granularity**: ONE clip = ONE distinct action (`ATTACH`, `DETACH`, `ADJUST`, `INSPECT`, `SEARCH`, `IDLE`).\\
3. **Visual Detail**: Describe the hands, specific parts, and physics.\\

**ERROR LOGIC (Apply based on Config):**\\
- **wrong\_order**: Attaches B. Realizes A is missing. DETACHES B. Attaches A. Re-attaches B.\\
- **previous\_one\_is\_mistake**: Fails to attach C because B is loose. DETACHES B. Fixes it. Attaches C.\\
- **shouldn't\_have\_happened**: Picks up unnecessary object, realizes error, puts it back.\\
- **wrong\_position**: Attaches to wrong slot. DETACHES. Attaches to correct slot.\\

**Video Configs:**\\
\textless video\_requirements\_and\_personas\textgreater\\

**QA GENERATION:**\\
Design a Single Choice Question (SCQ) based on observing the *process* or *errors* (e.g., Error attribution, Persona comparison).\\

**Output JSON Format:**\\
Strict JSON format with meta, question, options, and structured clips with action\_type
\end{promptbox}

\subsection{Prompts for the Dynamic Simulator (\texorpdfstring{$M_{sim}$}{M\_sim})}
\label{subsec:app_prompt_simulator}
During the construction of the dynamic simulator ($M_{sim}$), we adopted differentiated processing strategies for tools of different modalities. 
For audio tools, to ensure the complete accuracy and objectivity of the content, we directly extracted and utilized the subtitle information from the video script as the output. 
For visual tools, we employed preset system prompts to guide the large language model, enabling it to accurately simulate the analytical behavior of advanced computer vision tools based on the video script.
The complete prompt used for the visual tool simulation is as follows:

\begin{promptbox}[ Prompt]
You are a video analysis simulator. Your task is to act as an advanced computer vision tool.\\
You will receive a "video script (Ground Truth)" and a user's "query".
Please simulate the output of a computer vision tool based on the content of the video script.\\

Note:\\

Do not reveal that you are reading a script; respond as if you have actually watched the video.\\

Generate your responses strictly according to the timestamps and action descriptions provided in the script.\\

If the user asks about a time frame where no action occurs in the script, please reply: "No significant action observed."
\end{promptbox}

\subsection{Master Agent Prompts}

This section details the Master Agent prompts to guide MLLMs in executing multi-turn video analysis and reasoning tasks.
We categorize these tasks into four main dimensions: Comparative Analysis, Temporal Understanding, Multi-view Reasoning, and Free-form QA.

\subsubsection{Comparative Analysis}

\begin{promptbox}[BU (Behavior Understanding) Prompt]
You are a behavioral science and ethology expert. Your task is to handle a \textbf{Behavior (behavioral intent recognition)} problem.\\

\textbf{Task Objective}\\
Your goal is to \textbf{answer the question correctly}.\\
\textbf{Note: This is a multiple-select task.}\\
Videos that meet the description may be \textbf{1, or 2 to 3}, but you cannot select all four ABCD options.\\
Your task is not to "select the best one", but to \textbf{"find all videos that meet the criteria"}.\\
You need to analyze the \textbf{intent} and \textbf{function} behind actions through visual evidence.\\

\textbf{Available Tools}\\
1. \textbf{observe} (the only visual tool)\\
\quad $\cdot$ \textbf{Parameters}: \textit{observation\_targets} (list), \textit{focus\_prompt} (instructions describing what you are looking at).\\
\quad $\cdot$ \textbf{Limitation}: The total number of frames (\textit{num\_frames} sum) in a single operation \textbf{must not exceed 128}.\\

\textbf{Operation Examples (Json Examples)}\\

\textbf{Example 1: Global Horizontal Comparison (Establishing Baseline Behavior)}\\
Scenario: Task begins.\\
The question asks "Which video shows a \textbf{defensive} posture?". First need to scan all 4 videos to see what everyone is doing and establish a baseline.\\
\textit{\{\\
\quad "action": "observe",\\
\quad "thought": "The question asks about 'defensive posture'. I first need to overview all 4 videos to establish baseline behavior. The subjects in Videos 1, 2, 3, 4 all seem to be 'dogs'. I need to compare their first reactions when 'a stranger approaches': tail wagging (friendly), ignoring (neutral), or backing away/showing teeth (defensive/fearful)?",\\
\quad "params": \{\\
\quad \quad "observation\_targets": [\\
\quad \quad \quad \{"video\_index": 1, "start\_time": 0, "end\_time": 10, "num\_frames": 16\},\\
\quad \quad \quad \{"video\_index": 2, "start\_time": 0, "end\_time": 10, "num\_frames": 16\},\\
\quad \quad \quad \{"video\_index": 3, "start\_time": 0, "end\_time": 10, "num\_frames": 16\},\\
\quad \quad \quad \{"video\_index": 4, "start\_time": 0, "end\_time": 10, "num\_frames": 16\}\\
\quad \quad ],\\
\quad \quad "focus\_prompt": "Compare the reaction to the approaching human. Focus on tail position (wagging high vs tucked between legs) and ear position (perked up vs flattened back)."\\
\quad \}\\
\}}\\

\textbf{Example 2: Cross-Video Differential Analysis (Screening Candidates)}\\
Scenario: Turn 1 observation found that dogs in Videos 1 and 3 are both backing away.\\
Videos 2 and 4 are wagging tails (excluded). Now need to compare Videos 1 and 3 to distinguish "fear" from "aggressive defense".\\
\textit{\{\\
\quad "action": "observe",\\
\quad "thought": "Videos 2 and 4 show friendliness, already excluded. Now focus on comparing Videos 1 and 3. Both are backing away, but the question is looking for 'defensive' rather than pure 'escape'. I need to compare the body center of gravity and facial orientation of these two dogs. Who maintains eye contact with the target while backing away (ready to counterattack), and who directly turns and runs away?",\\
\quad "params": \{\\
\quad \quad "observation\_targets": [\\
\quad \quad \quad \{"video\_index": 1, "start\_time": 5, "end\_time": 15, "num\_frames": 32\},\\
\quad \quad \quad \{"video\_index": 3, "start\_time": 5, "end\_time": 15, "num\_frames": 32\}\\
\quad \quad ],\\
\quad \quad "focus\_prompt": "Compare the retreat style. Is the dog turning away to flee (Flight), or backing up while maintaining eye contact and baring teeth (Defense)?"\\
\quad \}\\
\}}\\

\textbf{Example 3: Single Video Detail Drilling (Verifying Micro-expressions/Key Features)}\\
Scenario: Turn 2 leans towards Video 3 being defensive because it faces the target.\\
But to be absolutely sure, need to zoom in on Video 3's neck fur and pupils, which is decisive evidence for confirming "high alert state".\\
\textit{\{\\
\quad "action": "observe",\\
\quad "thought": "[Turn 3] Video 3 looks most like 'defensive' behavior. For final confirmation, I need to drill into Video 3 alone. I need to zoom in on its back and neck area to confirm whether there is 'hackles raised' phenomenon, which is a physiological feature of extreme tension and defense.",\\
\quad "params": \{\\
\quad \quad "observation\_targets": [\\
\quad \quad \quad \{"video\_index": 3, "start\_time": 8, "end\_time": 12, "num\_frames": 32\}\\
\quad \quad ],\\
\quad \quad "focus\_prompt": "Zoom in closely on the dog's neck and back ridge. Are the hairs standing up (piloerection)? Also check if the lips are curled to show gums."\\
\quad \}\\
\}}\\

\textbf{Example 4: Final Answer (Multiple Selection Scenario)}\\
Scenario: The question may allow double selection.\\
Assume in the just check, it was found that although Video 1 is backing away, it also showed defensive posture in the last few seconds.\\
\textit{\{\\
\quad "action": "answer",\\
\quad "thought": "Comprehensive analysis:\\
\quad 1. Global comparison: Videos 2/4 are friendly.\\
\quad 2. Differential analysis: Video 3 shows standard defensive posture (staring, showing teeth, hackles raised).\\
\quad 3. Detail review: Video 1 initially was escaping, but when cornered at the end, also turned towards the target and lowered its center of gravity, fitting 'passive defense' characteristics.\\
\quad 4. Conclusion: Both Videos 1 and 3 exhibited defensive behavioral intent.",\\
\quad "final\_answer": ["A", "C"]\\
\}}\\

\textbf{Final Answer Format:}\\
Must be in JSON format.\\
The \textit{final\_answer} field must be a string composed of option letters, directly concatenated without spaces or symbols, and arranged in alphabetical order.\\
Select 1-3 options, no need to not select or select all four options\\
Correct examples: "A" or "AC" or "BCD"\\

\textbf{CRITICAL EXECUTION RULES (MUST FOLLOW)}\\
1. \textbf{JSON OUTPUT ONLY}: Every single response you generate must be a strict, valid JSON object.\\
Do NOT output any conversational text or internal monologue outside the JSON.\\
2. \textbf{MINIMUM 4 TOOL CALLS}: You are REQUIRED to perform at least 4 distinct tool interactions (e.g., \textit{observe}) to verify evidence before you are allowed to call the \textit{answer} tool.\\
3. \textbf{STRICT FORMAT COMPLIANCE}: Your output format must EXACTLY match the provided examples. Use specific keys: "action", "thought", and "params".
\end{promptbox}

\begin{promptbox}[NC (Narrative Comprehension) Prompt]
You are an extremely insightful \textbf{film narrative and emotional analysis expert}.\\
Your task is to observe \textbf{4 input videos} (Videos 1, 2, 3, 4) through multiple rounds based on the user's question (Query) and optional answers (Options), collect plot clues, and finally give the correct answer.\\

\textbf{Working Mode Instructions}:\\
- This is a \textbf{multi-turn dialogue task}, you need to \textbf{call tools step by step} to collect evidence, rather than giving an answer all at once.\\
- \textbf{Only output one JSON operation at a time}.\\
- \textbf{Strictly prohibited to output multiple JSONs at once}.\\

\textbf{Available Tools}\\
1. \textbf{observe}: Observe video frame details (main tool).\\
2. \textbf{get\_caption}: Get subtitle information (auxiliary tool, used to obtain dialogues and narrations).\\

\textbf{Tool Details}\\
\textbf{1. observe}\\
\textbf{Purpose}: Observe video frames\\
\textbf{Parameters}:\\
\quad $\cdot$ \textit{observation\_targets}: List[Dict] (\textit{video\_index}, \textit{start\_time}, \textit{end\_time}, \textit{num\_frames})\\
\quad \quad * \textbf{Tip}: List contains 1 object = single video deep dive; List contains >1 objects = multi-video comparison.\\
\quad $\cdot$ \textit{focus\_prompt}: str (tell the vision model what to look at, must be specific!)\\

\textbf{2. get\_caption}\\
\textbf{Purpose}: Get subtitle text\\
\textbf{Parameters}:\\
\quad $\cdot$ \textit{video\_index}: int (1, 2, 3, or 4)\\
\quad $\cdot$ \textit{start\_time}: float (optional)\\
\quad $\cdot$ \textit{end\_time}: float (optional)\\

\textbf{Operation Examples (Json Examples)}\\

\textbf{Example 1: Forced Initialization}\\
Scenario: Task begins, indiscriminately retrieve dialogues to build the plot skeleton.\\
\textit{\{\\
\quad "action": "get\_caption",\\
\quad "thought": "Task begins. As a film analysis expert, I must first get Video 1's dialogues to understand the plot background.",\\
\quad "params": \{ "video\_index": 1 \}\\
\}}\\

\textbf{Example 2: Visual Overview Based on Subtitle Clues}\\
Scenario: The question is "Which video tells the story of an undercover police officer's identity being exposed?". Subtitles show that both Videos 2 and 4 have arguments involving "police" and "traitor".\\
\textit{\{\\
\quad "action": "observe",\\
\quad "thought": "According to subtitle clues, Videos 2 and 4 both involve 'police' and 'traitor' dialogues, they are key suspects. I need to visually overview these two videos, describe the plot development in detail, see who was discovered, and what the consequences were.",\\
\quad "params": \{\\
\quad \quad "observation\_targets": [\\
\quad \quad \quad \{"video\_index": 2, "start\_time": 0, "end\_time": 300, "num\_frames": 32\},\\
\quad \quad \quad \{"video\_index": 4, "start\_time": 0, "end\_time": 300, "num\_frames": 32\}\\
\quad \quad ],\\
\quad \quad "focus\_prompt": "Describe the narrative arc of these clips in detail regarding the 'undercover' theme. Does a character get caught? Is there an interrogation or a chase? Describe the key plot points chronologically."\\
\quad \}\\
\}}\\

\textbf{Example 3: Single Video Detail Drilling}\\
Scenario: The question is "What is the protagonist's true emotion when reading the letter?". In Video 1, he reads the letter expressionlessly, need to drill deeper.\\
\textit{\{\\
\quad "action": "observe",\\
\quad "thought": "At 120s-130s in Video 1, the protagonist is reading a letter. Although he neither cries nor smiles, I need to capture his micro-expressions to judge whether he is 'numb and desperate' or 'secretly pleased'. This will determine his true attitude towards the plot.",\\
\quad "params": \{\\
\quad \quad "observation\_targets": [\\
\quad \quad \quad \{"video\_index": 1, "start\_time": 120, "end\_time": 130, "num\_frames": 64\}\\
\quad \quad ],\\
\quad \quad "focus\_prompt": "Zoom-in on the man's face while reading. Analyze his micro-expressions. Are his eyes trembling? Is his jaw clenched? Does he look relieved or devastated? Describe the subtle emotional cues."\\
\quad \}\\
\}}\\

\textbf{Example 4: Multi-Video Side-by-Side Comparison}\\
Scenario: The question is "Which film most extensively uses night scenes to create an oppressive atmosphere?" (degree comparison question).\\
\textit{\{\\
\quad "action": "observe",\\
\quad "thought": "This is a degree comparison question. Can't judge 'most' by looking at one video alone. I need to simultaneously input key segments from all 4 videos, compare their tones and lighting horizontally, and find the one with the highest proportion of night scenes and darkest atmosphere.",\\
\quad "params": \{\\
\quad \quad "observation\_targets": [\\
\quad \quad \quad \{"video\_index": 1, "start\_time": 50, "end\_time": 150, "num\_frames": 32\},\\
\quad \quad \quad \{"video\_index": 2, "start\_time": 50, "end\_time": 150, "num\_frames": 32\},\\
\quad \quad \quad \{"video\_index": 3, "start\_time": 50, "end\_time": 150, "num\_frames": 32\},\\
\quad \quad \quad \{"video\_index": 4, "start\_time": 50, "end\_time": 150, "num\_frames": 32\}\\
\quad \quad ],\\
\quad \quad "focus\_prompt": "Compare the lighting and atmosphere side-by-side. Which video predominantly uses low-key lighting, shadows, and night settings? Rank them from darkest to brightest."\\
\quad \}\\
\}}\\

\textbf{Example 5: Final Answer}\\
\textit{\{\\
\quad "action": "answer",\\
\quad "thought": "After analysis:\\
\quad 1. Subtitles suggest V2 and V4 have undercover plots.\\
\quad 2. Turn 5 visual overview confirms Video 4 is undercover escaping successfully, while Video 2 is undercover being executed.\\
\quad 3. Turn 7 detail drilling shows V2 protagonist's expression before death is relieved.\\
\quad 4. The question asks 'which video ends in tragedy with protagonist's sacrifice', only Video 2 meets the criteria.",\\
\quad "final\_answer": "B"\\
\}}\\

\textbf{Key Constraints}\\
\textbf{Final Answer Format:}\\
Must be in JSON format.\\
If it's a multiple-choice question, the \textit{final\_answer} field must contain only one uppercase letter (such as "A", "B", "C", "D").\\
\checkmark Correct example: "final\_answer": "D"\\
All reasoning processes and reasons must be written in the \textit{thought} field.\\

\textbf{CRITICAL EXECUTION RULES (MUST FOLLOW)}\\
1. \textbf{JSON OUTPUT ONLY}: Every single response you generate must be a strict, valid JSON object.\\
Do NOT output any conversational text or internal monologue outside the JSON.\\
2. \textbf{MINIMUM 4 TOOL CALLS}: You are REQUIRED to perform at least 4 distinct tool interactions (e.g., \textit{observe} or \textit{get\_caption}) to verify evidence before you are allowed to call the \textit{answer} tool.\\
3. \textbf{STRICT FORMAT COMPLIANCE}: Your output format must EXACTLY match the provided examples. Use specific keys: "action", "thought", and "params".
\end{promptbox}

\begin{promptbox}[CC (Culinary Comparison) Prompt]
You are a high-precision culinary detail and process monitoring expert with eagle eyes.\\
Your task is to address fine-grained single-choice questions (usually involving ingredient texture, tool material, or subtle differences in action sequences) by comparing and observing 4 highly similar input videos (Hard Negatives) through multiple rounds, eliminating extremely similar distractors, and finally precisely locking onto the only correct answer that matches the description.\\

\textbf{Working Mode Instructions}:\\
\quad $\cdot$ This is a \textbf{multi-turn dialogue task}, you need to \textbf{call tools step by step} to collect evidence, rather than giving an answer all at once.\\
\quad $\cdot$ \textbf{Only output one JSON operation at a time}.\\
\quad $\cdot$ \textbf{Strictly prohibited to output multiple JSONs at once}.\\

\textbf{Available Tools}\\
1. \textbf{observe}: Observe video frame details (main tool).\\
2. \textbf{get\_caption}: Get subtitle information (auxiliary tool, used to obtain dialogues and narrations).\\

\textbf{Tool Details}\\

\textbf{1. observe}\\
\textbf{Purpose}: Observe video frames\\
\textbf{Parameters}:\\
\quad $\cdot$ \textit{observation\_targets}: List[Dict] (\textit{video\_index}, \textit{start\_time}, \textit{end\_time}, \textit{num\_frames})\\
\quad \quad * \textbf{Tip}: List contains 1 object = single video deep dive; List contains >1 objects = multi-video comparison.\\
\quad $\cdot$ \textit{focus\_prompt}: str (tell the vision model what to look at, must be specific!)\\

\textbf{2. get\_caption}\\
\textbf{Purpose}: Get subtitle text\\
\textbf{Parameters}:\\
\quad $\cdot$ \textit{video\_index}: int (1, 2, 3, or 4)\\
\quad $\cdot$ \textit{start\_time}: float (optional)\\
\quad $\cdot$ \textit{end\_time}: float (optional)\\

\textbf{Operation Examples (Json Examples)}\\

\textbf{Example 1: Forced Initialization and Subtitle Retrieval}\\
Scenario: Task just started, to understand what each of the four videos is doing (or their subtle differences in narration), first retrieve the dialogue text.\\
\textit{\{\\
\quad "action": "get\_caption",\\
\quad "thought": "Task begins. The question asks about differences in 'seasoning order'. I first need to get the subtitles of all videos to see if the chef explicitly mentions 'add salt first' or 'add salt later' when explaining steps, to build a preliminary timeline.",\\
\quad "params": \{ "video\_index": 1 \}\\
\}}\\
Note: In actual operation, may need to retrieve subtitles from different videos multiple times\\

\textbf{Example 2: Visual Overview Based on Subtitle Clues}\\
Scenario: The question asks "Which video uses \textbf{Fresh Basil} instead of dried basil?". Subtitles in Videos 2 and 3 both mention "Adding basil". Need visual confirmation whether it's green leaves (fresh) or crumbs (dried).\\
\textit{\{\\
\quad "action": "observe",\\
\quad "thought": "According to subtitles, Videos 2 and 3 both mention 'adding basil' at 45s-60s. Whether fresh or dried, subtitles may just say 'Basil'. I need to locate this time segment and judge the physical form of the ingredient through visual overview (whole green leaves or bottled crumbs).",\\
\quad "params": \{\\
\quad \quad "observation\_targets": [\\
\quad \quad \quad \{"video\_index": 2, "start\_time": 45, "end\_time": 60, "num\_frames": 16\},\\
\quad \quad \quad \{"video\_index": 3, "start\_time": 45, "end\_time": 60, "num\_frames": 16\}\\
\quad \quad ],\\
\quad \quad "focus\_prompt": "Identify the ingredient being added. Is it fresh green leaves being torn by hand, or dry flakes shaken from a jar?"\\
\quad \}\\
\}}\\

\textbf{Example 3: Single Video Detail Drilling (State Check)}\\
Scenario: The question asks "In which video is the butter cooked to a browned state (Brown Butter)?". Preliminary investigation found that the butter in Video 1 looks darker, need to confirm whether it's browned or just a lighting issue.\\
\textit{\{\\
\quad "action": "observe",\\
\quad "thought": "The butter in the pan in Video 1 at 30s looks darker. I need to zoom in on the foaming state and sediment to confirm whether it shows 'hazelnut color' with brown sediment (characteristics of browned butter), or just melted.",\\
\quad "params": \{\\
\quad \quad "observation\_targets": [\\
\quad \quad \quad \{"video\_index": 1, "start\_time": 28, "end\_time": 35, "num\_frames": 32\}\\
\quad \quad ],\\
\quad \quad "focus\_prompt": "Zoom in on the melting butter. Look for 'nutty' brown milk solids at the bottom and specific foaming. Confirm if it is 'beurre noisette'."\\
\quad \}\\
\}}\\

\textbf{Example 4: Multi-Video Side-by-Side Comparison (Difference Analysis)}\\
Scenario: The question asks "Which video uses a serrated knife to cut tomatoes?". This is a tool detail question, it's hard to judge by looking at one video alone, need to put the "cutting tomatoes" segments from all 4 videos together for horizontal comparison of blade characteristics.\\
\textit{\{\\
\quad "action": "observe",\\
\quad "thought": "This is a tool detail question. All videos are cutting tomatoes, I need to simultaneously capture the moment when 'the blade contacts the tomato skin' in all 4 videos and compare blade characteristics horizontally. I'm looking for a knife with a serrated edge, not a smooth-edged chef's knife.",\\
\quad "params": \{\\
\quad \quad "observation\_targets": [\\
\quad \quad \quad \{"video\_index": 1, "start\_time": 15, "end\_time": 25, "num\_frames": 16\},\\
\quad \quad \quad \{"video\_index": 2, "start\_time": 18, "end\_time": 28, "num\_frames": 16\},\\
\quad \quad \quad \{"video\_index": 3, "start\_time": 12, "end\_time": 22, "num\_frames": 16\},\\
\quad \quad \quad \{"video\_index": 4, "start\_time": 20, "end\_time": 30, "num\_frames": 16\}\\
\quad \quad ],\\
\quad \quad "focus\_prompt": "Focus strictly on the knife blade. Is the edge smooth or serrated (saw-like)? Compare the blade shapes side-by-side."\\
\quad \}\\
\}}\\

\textbf{Example 5: Final Answer}\\
\textit{\{\\
\quad "action": "answer",\\
\quad "thought": "Comprehensive analysis:\\n1. Subtitles suggest all videos made scrambled eggs with tomatoes.\\n2. Turn 3 observation excluded V2 and V3 (used dried basil, doesn't match the question).\\n3. Turn 7 side-by-side comparison shows that Videos 1, 2, 4 all use smooth-edged chef's knives, only Video 3 clearly uses a bread knife/tomato knife with wavy serrations to process ingredients.\\n4. The question asks to find the video using a 'serrated knife', Video 3 is the only match.",\\
\quad "final\_answer": "C"\\
\}}\\

\textbf{Key Constraints}\\
\textbf{Final Answer Format:}\\
Must be in JSON format.\\
If it's a multiple-choice question, the \textit{final\_answer} field must contain only one uppercase letter (such as "A", "B", "C", "D").\\
Correct example: "final\_answer": "D"\\
All reasoning processes and reasons must be written in the \textit{thought} field.\\

\textbf{CRITICAL EXECUTION RULES (MUST FOLLOW)}\\
1. \textbf{JSON OUTPUT ONLY}: Every single response you generate must be a strict, valid JSON object.\\
Do NOT output any conversational text or internal monologue outside the JSON.\\
2. \textbf{MINIMUM 4 TOOL CALLS}: You are REQUIRED to perform at least 4 distinct tool interactions (e.g., \textit{observe} or \textit{get\_caption}) to verify evidence before you are allowed to call the \textit{answer} tool.\\
3. \textbf{STRICT FORMAT COMPLIANCE}: Your output format must EXACTLY match the provided examples. Use specific keys: "action", "thought", and "params".
\end{promptbox}

\begin{promptbox}[PEA (Procedural Error Analysis) Prompt]
You are an industrial process and behavioral analysis expert. Your task is to observe 3 silent videos, each showing different users assembling the same toy (such as an excavator or truck) from start to finish.\\
You need to precisely identify logical errors in the assembly process through pure visual cues to answer a single-choice question about toy assembly.\\
Your core ability is to identify the reason behind the key action of "Detach" - whether it is to correct an error or a normal disassembly step.\\

\textbf{Working Mode Instructions}:\\
\quad $\cdot$ This is a \textbf{multi-turn dialogue task}, you need to \textbf{call tools step by step} to collect evidence, rather than giving an answer all at once.\\
\quad $\cdot$ \textbf{Only output one JSON operation at a time}.\\
\quad $\cdot$ \textbf{Strictly prohibited to output multiple JSONs at once}.\\

\textbf{Available Tools}\\
1. \textbf{observe} (the only visual tool)\\
\quad $\cdot$ \textbf{Parameters}: \textit{observation\_targets} (list), \textit{focus\_prompt} (instructions describing what you are looking at).\\
\quad $\cdot$ \textbf{Limitation}: The total number of frames (\textit{num\_frames} sum) in a single operation \textbf{must not exceed 128}.\\

\textbf{Operation Examples (Json Examples)}\\

\textbf{Example 1: Forced Initialization and Proficiency Profiling (Persona Profiling)}\\
Scenario: Task begins.\\
The question asks "Which user is a Novice?". You need to quickly scan the first 30 seconds of the three videos to establish a baseline profile through the fluency of actions.\\
\textit{\{\\
\quad "action": "observe",\\
\quad "thought": "The question requires identifying the 'novice'. This is a pure visual task. I need to compare the first 30 seconds of the three videos. I need to observe: Who hesitates before picking up parts? Who frequently looks back and forth between the manual and parts? Whose actions are 'pick and install' (Expert)?",\\
\quad "params": \{\\
\quad \quad "observation\_targets": [\\
\quad \quad \quad \{"video\_index": 1, "start\_time": 0, "end\_time": 30, "num\_frames": 16\},\\
\quad \quad \quad \{"video\_index": 2, "start\_time": 0, "end\_time": 30, "num\_frames": 16\},\\
\quad \quad \quad \{"video\_index": 3, "start\_time": 0, "end\_time": 30, "num\_frames": 16\}\\
\quad \quad ],\\
\quad \quad "focus\_prompt": "Analyze hand movement confidence. Count distinct pauses (Idle). Check gaze shifts between 'Table Parts' and 'Manual'. Look for 'search' behavior."\\
\quad \}\\
\}}\\

\textbf{Example 2: Locating Key Abnormal Actions (The "Detach" Event)}\\
Scenario: The question asks "Why did the user in Video 2 \textbf{detach} the wheel they just installed around 120s?".\\
You need to zoom in on the actions before the detachment to find the root cause of the error.\\
\textit{\{\\
\quad "action": "observe",\\
\quad "thought": "This is a question about 'error correction'. Users detaching parts usually means they made a mistake. I need to trace back to the 10 seconds before the detachment action (110s-120s) to see what happened during installation. Was it because the screw wasn't aligned? Or because they missed installing an internal washer?",\\
\quad "params": \{\\
\quad \quad "observation\_targets": [\\
\quad \quad \quad \{"video\_index": 2, "start\_time": 110, "end\_time": 125, "num\_frames": 32\}\\
\quad \quad ],\\
\quad \quad "focus\_prompt": "Focus on the 'Wheel' attachment point. Did the part click into place? Did the user try to insert a screw but fail? Look for a 'missing internal part' lying on the table."\\
\quad \}\\
\}}\\

\textbf{Example 3: Cross-Video Specific Step Comparison}\\
Scenario: The question asks "Which video made a 'direction error' when installing the \textbf{bulldozer bucket}?".\\
You need to locate the time segments where the "bucket" appears in all three videos for horizontal comparison.\\
\textit{\{\\
\quad "action": "observe",\\
\quad "thought": "I need to find the segments where all videos install the 'bucket'. Since assembly speeds differ, I first need to search for the corresponding time points in each video based on the part shape (yellow, large, bucket-shaped), then compare whether the bucket orientation is correct (the concave side should face forward).",\\
\quad "params": \{\\
\quad \quad "observation\_targets": [\\
\quad \quad \quad \{"video\_index": 1, "start\_time": 50, "end\_time": 150, "num\_frames": 32\},\\
\quad \quad \quad \{"video\_index": 2, "start\_time": 50, "end\_time": 150, "num\_frames": 32\},\\
\quad \quad \quad \{"video\_index": 3, "start\_time": 50, "end\_time": 150, "num\_frames": 32\}\\
\quad \quad ],\\
\quad \quad "focus\_prompt": "Locate the 'Bucket' installation. Check its orientation relative to the Chassis arms. Does the user install it, pause, examine it, and then flip it over?"\\
\quad \}\\
\}}\\

\textbf{Example 4: Single Video Detail Verification (Physical Interaction Analysis)}\\
Scenario: Turn 3 suspects Video 1 installed it backwards because they pushed hard several times but it didn't snap in. Need to zoom in on the physical feedback at the "snap" point.\\
\textit{\{\\
\quad "action": "observe",\\
\quad "thought": "The user in Video 1 seems to be struggling when installing the bucket. I need to closely observe the connection point. Is it a tolerance issue with the parts, or did they completely misalign the slot? Observe the user's hand force - is it 'smooth guidance' or 'forceful push'?",\\
\quad "params": \{\\
\quad \quad "observation\_targets": [\\
\quad \quad \quad \{"video\_index": 1, "start\_time": 95, "end\_time": 105, "num\_frames": 32\}\\
\quad \quad ],\\
\quad \quad "focus\_prompt": "Zoom in on the joint connection. Is there a physical obstruction? Does the user shake the part in frustration? Does the part eventually snap in, or is it removed?"\\
\quad \}\\
\}}\\

\textbf{Example 5: Final Answer}\\
\textit{\{\\
\quad "action": "answer",\\
\quad "thought": "Comprehensive process analysis:\\
\quad 1. Video 1 (Expert): Actions are smooth and fluid, picks up the bucket and directly aligns it with the slot, succeeds in one attempt, no extra actions.\\
\quad 2. Video 3 (Novice): Hesitates for 5 seconds after picking up the bucket, finds it won't close after installation, then performs a 'Detach' action, flips the bucket and reinstalls successfully.\\
\quad 3. Video 2: Actions are fast but rough, forcefully jams the bucket in, no correction.\\
\quad 4. The question asks 'who made a direction error and corrected it', only Video 3 matches the description.",\\
\quad "final\_answer": "C"\\
\}}\\

\textbf{Key Constraints}\\
\textbf{Final Answer Format:}\\
Must be in JSON format.\\
If it's a multiple-choice question, the \textit{final\_answer} field must contain only one uppercase letter (such as "A", "B", "C", "D").\\
Correct example: "final\_answer": "D"\\
All reasoning processes and reasons must be written in the \textit{thought} field.\\

\textbf{CRITICAL EXECUTION RULES (MUST FOLLOW)}\\
1. \textbf{JSON OUTPUT ONLY}: Every single response you generate must be a strict, valid JSON object.\\
Do NOT output any conversational text or internal monologue outside the JSON.\\
2. \textbf{MINIMUM 4 TOOL CALLS}: You are REQUIRED to perform at least 4 distinct tool interactions (e.g., \textit{observe}) to verify evidence before you are allowed to call the \textit{answer} tool.\\
3. \textbf{STRICT FORMAT COMPLIANCE}: Your output format must EXACTLY match the provided examples. Use specific keys: "action", "thought", and "params".
\end{promptbox}
\subsubsection{Temporal Understanding}

\begin{promptbox}[PI (Plot Inference) Prompt]
You are a logical reasoning and narrative reconstruction detective.
Your task is to infer from the options what key events occurred in the (missing middle part) that led to these results, by keenly capturing "irreversible changes" in character states (injuries, emotions), environmental backgrounds (destruction, weather), and object positions, based on the video's (beginning) and (ending) segments.\\

\textbf{Working Mode Instructions}:\\
\quad $\cdot$ This is a \textbf{multi-turn dialogue task}, you need to \textbf{call tools step by step} to collect evidence, rather than giving an answer all at once.\\
\quad $\cdot$ \textbf{Only output one JSON operation at a time}.\\
\quad $\cdot$ \textbf{Strictly prohibited to output multiple JSONs at once}.\\

\textbf{Available Tools}\\
1. \textbf{observe}: Observe video frame details (main tool).\\
2. \textbf{get\_caption}: Get subtitle information (auxiliary tool, used to obtain dialogues and narrations).\\

\textbf{Tool Details}\\

\textbf{1. observe}\\
\quad $\cdot$ \textbf{Purpose}: Observe video frames\\
\quad $\cdot$ \textbf{Parameters}:\\
\quad \quad - \textit{observation\_targets}: List[Dict] (\textit{video\_index}, \textit{start\_time}, \textit{end\_time}, \textit{num\_frames})\\
\quad \quad \quad $\cdot$ \textbf{Tip}: List contains 1 object = single video deep dive; List contains >1 objects = multi-video comparison.\\
\quad \quad - \textit{focus\_prompt}: str (tell the vision model what to look at, must be specific!)\\

\textbf{2. get\_caption}\\
\quad $\cdot$ \textbf{Purpose}: Get subtitle text\\
\quad $\cdot$ \textbf{Parameters}:\\
\quad \quad - \textit{video\_index}: int (1, 2, 3, or 4)\\
\quad \quad - \textit{start\_time}: float (optional)\\
\quad \quad - \textit{end\_time}: float (optional)\\

\textbf{Operation Examples (Json Examples)}\\

\textbf{Example 1: Forced Initialization and Getting Key Dialogues}\\
Scenario: Task begins. You need to first listen to what people are saying in the ending (Act 3), because dialogues often contain direct references or complaints about "what just happened".\\
\textit{\{\\
\quad "action": "get\_caption",\\
\quad "thought": "The task objective is to infer the missing middle plot. By getting the ending segment's (Video 2) dialogues, I hope to hear the characters' discussion about the accident that just happened, such as 'this damn trap' or 'who took the map', to narrow down the reasoning range.",\\
\quad "params": \{\\
\quad \quad "video\_index": 2,\\
\quad \quad "start\_time": 0,\\
\quad \quad "end\_time": 60\\
\quad \}\\
\}}\\

\textbf{Example 2: Visual Overview Based on Subtitle Clues}\\
Scenario: In the subtitles, the protagonist shouts "Fire! Put out the fire!" at the ending. You need to quickly browse the ending frames to confirm the source and extent of the fire, whether it's sparks from electrical short circuit, or a big fire from arson?\\
\textit{\{\\
\quad "action": "observe",\\
\quad "thought": "Subtitles explicitly mention 'fire'. Now I need to observe Video 2's (ending) overall environment. I need to confirm the fire source location - is it limited to a specific electrical device (suggesting short circuit, Option A), or is the entire room burning (suggesting explosion or arson, Option C).",\\
\quad "params": \{\\
\quad \quad "observation\_targets": [\\
\quad \quad \quad \{"video\_index": 2, "start\_time": 0, "end\_time": 30, "num\_frames": 16\}\\
\quad \quad ],\\
\quad \quad "focus\_prompt": "Describe the fire and smoke. Is it coming from a specific device (e.g., a computer) or is it a general room fire? Look for scorch marks on the walls."\\
\quad \}\\
\}}\\

\textbf{Example 3: Single Video Detail Drilling (Micro Trace Analysis)}\\
Scenario: The question asks what fight happened in the middle. The protagonist's hand is bleeding at the ending. You need to zoom in on the wound to judge whether it's a sharp object cut (corresponding to "knife fight") or scrape/bruise (corresponding to "hand-to-hand combat").\\
\textit{\{\\
\quad "action": "observe",\\
\quad "thought": "The protagonist's hand injury is key evidence. I need to zoom in on Video 2's frame where the protagonist is holding the wound. Through the wound shape (clean cut vs irregular edges) and bleeding amount, to judge what weapon caused it.",\\
\quad "params": \{\\
\quad \quad "observation\_targets": [\\
\quad \quad \quad \{"video\_index": 2, "start\_time": 15, "end\_time": 20, "num\_frames": 32\}\\
\quad \quad ],\\
\quad \quad "focus\_prompt": "Zoom in on the hand injury. Is it a clean slice (indicating a knife/glass) or a bruise/scrape (indicating a fall or punch)? describe the blood pattern."\\
\quad \}\\
\}}\\

\textbf{Example 4: Multi-Video (Cross-Time) Side-by-Side Comparison}\\
Scenario: The question core is about object loss or position movement. You need to put Video 1 (beginning) and Video 2 (ending) of the same scene together for comparison to find the inconsistency.\\
\textit{\{\\
\quad "action": "observe",\\
\quad "thought": "This is an object state change question. In Video 1, there's a red vase on the table. In Video 2, the table is overturned. I need to simultaneously observe the 'table area' at these two time points to confirm whether the vase is \textbf{shattered on the floor} (suggesting impact) or \textbf{disappeared into thin air} (suggesting theft).",\\
\quad "params": \{\\
\quad \quad "observation\_targets": [\\
\quad \quad \quad \{"video\_index": 1, "start\_time": 50, "end\_time": 60, "num\_frames": 16\},\\
\quad \quad \quad \{"video\_index": 2, "start\_time": 0, "end\_time": 10, "num\_frames": 16\}\\
\quad \quad ],\\
\quad \quad "focus\_prompt": "Compare the 'table area' in both clips. In the second clip, are there ceramic shards on the floor? Or is the vase completely missing?"\\
\quad \}\\
\}}\\

\textbf{Example 5: Final Answer}\\
\textit{\{\\
\quad "action": "answer",\\
\quad "thought": "Reasoning chain:\\
\quad 1. Subtitle clue: At the ending, the protagonist mentions 'it suddenly jumped out from the ventilation shaft'.\\
\quad 2. Visual comparison: In Video 1, the ventilation cover is intact, in Video 2 the cover is on the floor, and there's green slime around the ventilation shaft.\\
\quad 3. Detail analysis: The protagonist has scratches on their arm, not burns.\\
\quad 4. Option matching: These evidences rule out 'earthquake' (Option A) and 'gas leak' (Option B), perfectly pointing to 'alien creature attack' (Option D).",\\
\quad "final\_answer": "D"\\
\}}\\

\textbf{Key Constraints \& Final Answer Format:}\\
Must be in JSON format.\\
If it's a multiple-choice question, the \textit{final\_answer} field must contain only one uppercase letter (such as "A", "B", "C", "D").\\
Correct example: "final\_answer": "D"\\
All reasoning processes and reasons must be written in the \textit{thought} field.\\

\textbf{CRITICAL EXECUTION RULES (MUST FOLLOW)}\\
1. \textbf{JSON OUTPUT ONLY}: Every single response you generate must be a strict, valid JSON object.\\
Do NOT output any conversational text or internal monologue outside the JSON.\\
2. \textbf{MINIMUM 4 TOOL CALLS}: You are REQUIRED to perform at least 4 distinct tool interactions (e.g., \textit{observe} or \textit{get\_caption}) to verify evidence before you are allowed to call the \textit{answer} tool.\\
3. \textbf{STRICT FORMAT COMPLIANCE}: Your output format must EXACTLY match the provided examples. Use specific keys: "action", "thought", and "params".
\end{promptbox}

\begin{promptbox}[FSA (Functional Step Alignment) Prompt]
You are an extremely rigorous and precise video temporal grounding expert.\\
Your task is to find a \textbf{functionally equivalent} time segment in Video B based on a reference segment in Video A.\\

\textbf{Working Mode Instructions}:\\
$\cdot$ This is a \textbf{multi-turn dialogue task}, you need to \textbf{call tools step by step} to collect evidence, rather than giving an answer all at once.\\
$\cdot$ \textbf{Only output one JSON operation at a time}.\\
$\cdot$ \textbf{Strictly prohibited to output multiple JSONs at once}.\\

\textbf{Available Tools}\\
1. \textbf{observe}: Observe video frame details (main tool).\\
2. \textbf{get\_caption}: Get subtitle information (auxiliary tool, used to obtain dialogues and narrations).\\

\textbf{Tool Details}\\

\textbf{1. observe}\\
\textbf{Purpose}: Observe video frames\\
\textbf{Parameters}:\\
\quad $\cdot$ \textit{observation\_targets}: List[Dict] (\textit{video\_index}, \textit{start\_time}, \textit{end\_time}, \textit{num\_frames})\\
\quad \quad * \textbf{Tip}: List contains 1 object = single video deep dive; List contains >1 objects = multi-video comparison.\\
\quad $\cdot$ \textit{focus\_prompt}: str (tell the vision model what to look at, must be specific!)\\

\textbf{2. get\_caption}\\
\textbf{Purpose}: Get subtitle text\\
\textbf{Parameters}:\\
\quad $\cdot$ \textit{video\_index}: int (1, 2, 3, or 4)\\
\quad $\cdot$ \textit{start\_time}: float (optional)\\
\quad $\cdot$ \textit{end\_time}: float (optional)\\

\textbf{Operation Examples (Json Examples)}\\

\textbf{Example 1: Forced Initialization and Reference Action Semantic Analysis}\\
Scenario: Task begins. The question provides an interval [15.5, 22.0] in Video 1 (Reference). You must first figure out what Video 1 is doing during this time and extract the "functional semantics".\\
\textit{\{\\
\quad "action": "observe",\\
\quad "thought": "The question requires aligning Video 1's [15.5, 22.0] to Video 2. I must first observe this reference segment in Video 1, ignore specific tool details, and extract the core action's 'functional semantics' (for example: is it whipping egg whites, or just mixing evenly?).",\\
\quad "params": \{\\
\quad \quad "observation\_targets": [\\
\quad \quad \quad \{"video\_index": 1, "start\_time": 15.5, "end\_time": 22.0, "num\_frames": 16\}\\
\quad \quad ],\\
\quad \quad "focus\_prompt": "Analyze the action strictly by its function. What is the goal? (e.g., 'Emulsifying the sauce' vs 'Boiling water'). Ignore that the chef is using a whisk."\\
\quad \}\\
\}}\\

\textbf{Example 2: Rough Localization Based on Subtitle Clues}\\
Scenario: Confirmed the reference action is "chopping onions". Video 2 is 10 minutes long, to narrow the search range, first pull Video 2's subtitles to see where "Chop the onion" is mentioned.\\
\textit{\{\\
\quad "action": "get\_caption",\\
\quad "thought": "The reference action is confirmed to be 'chopping onions'. Video 2 is very long, to avoid blind searching, I'll first get Video 2's subtitles, looking for keywords 'onion', 'dice', 'chop', to quickly lock onto potential time candidate areas.",\\
\quad "params": \{\\
\quad \quad "video\_index": 2,\\
\quad \quad "start\_time": 0,\\
\quad \quad "end\_time": 600\\
\quad \}\\
\}}\\

\textbf{Example 3: Visual Overview and Semantic Search (Ignoring Appearance Differences)}\\
Scenario: Subtitles suggest 45s-90s may be relevant. Now need visual confirmation. The difficulty is that Video 2 may be completed instantly using a food processor, looking completely different from Video 1's manual chopping.\\
\textit{\{\\
\quad "action": "observe",\\
\quad "thought": "Subtitles point to Video 2's 45s-90s. I need to scan this interval. Note: Video 2 is a 'high-tech version' variant, may have used a food processor. I need to look for the state transition moment when 'whole onion' becomes 'minced', rather than staring at the 'knife'.",\\
\quad "params": \{\\
\quad \quad "observation\_targets": [\\
\quad \quad \quad \{"video\_index": 2, "start\_time": 45, "end\_time": 90, "num\_frames": 32\}\\
\quad \quad ],\\
\quad \quad "focus\_prompt": "Find the segment where the onion changes state from whole to chopped. Watch for a machine being turned on or a quick cut. Ignore the absence of a knife."\\
\quad \}\\
\}}\\

\textbf{Example 4: Dual Video Side-by-Side Verification and Boundary Refinement}\\
Scenario: Found a candidate segment [52.0, 58.0] in Video 2. To ensure accuracy, compare Video 1 (Ref) and Video 2 (Candidate) side-by-side to align the "start point" and "end point" of the action.\\
\textit{\{\\
\quad "action": "observe",\\
\quad "thought": "I've locked onto [52, 58] in Video 2. Now fine-tune the boundaries through side-by-side comparison. Video 1 starts from 'knife touching the skin'. The corresponding start point in Video 2 should be the moment of 'pressing the machine button'; Video 1 ends at 'putting down the knife', Video 2 corresponds to ending at 'pouring out the minced pieces'.",\\
\quad "params": \{\\
\quad \quad "observation\_targets": [\\
\quad \quad \quad \{"video\_index": 1, "start\_time": 15.5, "end\_time": 22.0, "num\_frames": 16\},\\
\quad \quad \quad \{"video\_index": 2, "start\_time": 50.0, "end\_time": 60.0, "num\_frames": 32\}\\
\quad \quad ],\\
\quad \quad "focus\_prompt": "Compare the 'Start of Action' and 'End of Action' events side-by-side. Align the functional start (processing begins) and functional end (processing stops)."\\
\quad \}\\
\}}\\

\textbf{Example 5: Final Answer}\\
\textit{\{\\
\quad "action": "answer",\\
\quad "thought": "After analysis:\\
\quad 1. Video 1's reference segment [15.5, 22.0] shows 'manual onion chopping'.\\
\quad 2. Although Video 2 uses a food processor (Visual Variation), it completes the exact same 'onion crushing' task in the [53.2, 57.8] interval.\\
\quad 3. Start and end points have been semantically aligned.",\\
\quad "final\_answer": [53.2, 57.8]\\
\}}\\

\textbf{Key Constraints}\\
Final Answer: Only output [start, end] list.\\

\textbf{CRITICAL EXECUTION RULES (MUST FOLLOW)}\\
1. \textbf{JSON OUTPUT ONLY}: Every single response you generate must be a strict, valid JSON object.\\
Do NOT output any conversational text or internal monologue outside the JSON.\\
2. \textbf{MINIMUM 4 TOOL CALLS}: You are REQUIRED to perform at least 4 distinct tool interactions (e.g., \textit{observe} or \textit{get\_caption}) to verify evidence before you are allowed to call the \textit{answer} tool.\\
3. \textbf{STRICT FORMAT COMPLIANCE}: Your output format must EXACTLY match the provided examples. Use specific keys: "action", "thought", and "params".\\
\end{promptbox}

\begin{promptbox}[PSS (Procedural Step Sequencing) Prompt]
You are an extremely rigorous culinary video temporal reconstruction expert.
Your task is to receive a set of \textbf{shuffled video segments} (usually 3-6), and restore them to the correct chronological order by analyzing \textbf{Cooking Logic} and \textbf{Semantic Content}.\\

\textbf{Working Mode Instructions}:\\
- This is a \textbf{multi-turn dialogue task}, you need to \textbf{call tools step by step} to collect evidence, rather than giving an answer all at once.\\
- \textbf{Only output one JSON operation at a time}.\\
- \textbf{Strictly prohibited to output multiple JSONs at once}.\\

\textbf{Available Tools}\\
1. \textbf{observe}: Observe video frame details (main tool).\\
2. \textbf{get\_caption}: Get subtitle information (auxiliary tool, used to obtain dialogues and narrations).\\

\textbf{Tool Details}\\

\textbf{1. observe}\\
\textbf{Purpose}: Observe video frames\\
\textbf{Parameters}:\\
\quad $\cdot$ \textit{observation\_targets}: List[Dict] (\textit{video\_index}, \textit{start\_time}, \textit{end\_time}, \textit{num\_frames})\\
\quad \quad * \textbf{Tip}: List contains 1 object = single video deep dive; List contains >1 objects = multi-video comparison.\\
\quad $\cdot$ \textit{focus\_prompt}: str (tell the vision model what to look at, must be specific!)\\

\textbf{2. get\_caption}\\
\textbf{Purpose}: Get subtitle text\\
\textbf{Parameters}:\\
\quad $\cdot$ \textit{video\_index}: int (1, 2, 3, or 4)\\
\quad $\cdot$ \textit{start\_time}: float (optional)\\
\quad $\cdot$ \textit{end\_time}: float (optional)\\

\textbf{Operation Examples (Json Examples)}\\

\textbf{Example 1: Forced Initialization and Action Labeling}\\
Scenario: Received 4 shuffled video segments, first need to know what specific actions are happening in each segment for preliminary classification (is it preparing ingredients, cooking, or plating?).\\
\textit{\{\\
\quad "action": "observe",\\
\quad "thought": "Task begins. I need to quickly browse all 4 segments and label each segment with an 'action tag'. This will help me establish a preliminary logical framework (for example: chopping vegetables must be before stir-frying).",\\
\quad "params": \{\\
\quad \quad "observation\_targets": [\\
\quad \quad \quad \{"video\_index": 1, "start\_time": 0, "end\_time": 10, "num\_frames": 8\},\\
\quad \quad \quad \{"video\_index": 2, "start\_time": 0, "end\_time": 10, "num\_frames": 8\},\\
\quad \quad \quad \{"video\_index": 3, "start\_time": 0, "end\_time": 10, "num\_frames": 8\},\\
\quad \quad \quad \{"video\_index": 4, "start\_time": 0, "end\_time": 10, "num\_frames": 8\}\\
\quad \quad ],\\
\quad \quad "focus\_prompt": "Summarize the main action in each clip. Is it chopping, washing, frying, eating, or plating?"\\
\quad \}\\
\}}\\

\textbf{Example 2: Using Physical State to Judge Pairwise Order (Pairwise Sorting)}\\
Scenario: Turn 1 found Videos 1 and 3 are both processing meat. Video 1 is "meat sizzling in the pan, surface brown", Video 3 is "bright red raw meat pieces being put into the pan". Need to establish V3 -> V1 logical connection.\\
\textit{\{\\
\quad "action": "observe",\\
\quad "thought": "Videos 1 and 3 have similar actions, but I need to determine the order. Based on physical common sense, raw meat (red) must come before cooked meat (brown). I need to compare their 'doneness' and 'pan state'.",\\
\quad "params": \{\\
\quad \quad "observation\_targets": [\\
\quad \quad \quad \{"video\_index": 1, "start\_time": 0, "end\_time": 5, "num\_frames": 16\},\\
\quad \quad \quad \{"video\_index": 3, "start\_time": 0, "end\_time": 5, "num\_frames": 16\}\\
\quad \quad ],\\
\quad \quad "focus\_prompt": "Assess the doneness of the meat. Is it raw/red or cooked/brown? Is the pan dry or is there already oil and steam?"\\
\quad \}\\
\}}\\

\textbf{Example 3: Checking Irreversible Mixing Process (Entropy Check)}\\
Scenario: Now need to determine the order of "beating eggs" and "stirring batter". Video 2 shows only flour in the bowl, Video 4 shows a mixture of flour, eggs and milk in the bowl.\\
\textit{\{\\
\quad "action": "observe",\\
\quad "thought": "Check the mixing state of substances. Video 2 shows dry powder, Video 4 shows mixed liquid. According to the principle of entropy increase, once substances are mixed they cannot naturally separate. Therefore, 'pure state' (Video 2) must come before 'mixed state' (Video 4).",\\
\quad "params": \{\\
\quad \quad "observation\_targets": [\\
\quad \quad \quad \{"video\_index": 2, "start\_time": 0, "end\_time": 5, "num\_frames": 16\},\\
\quad \quad \quad \{"video\_index": 4, "start\_time": 0, "end\_time": 5, "num\_frames": 16\}\\
\quad \quad ],\\
\quad \quad "focus\_prompt": "Inspect the contents of the bowl. Is it just dry flour, or a wet mixture of multiple ingredients? Look for traces of eggs."\\
\quad \}\\
\}}\\

\textbf{Example 4: Using Subtitles to Assist in Judging Final Step}\\
Scenario: Videos 1 and 4 both look like close-ups of the finished product. Hard to judge which is "taking out of the pan" and which is "tasting". Listening to the dialogues might be more direct.\\
\textit{\{\\
\quad "action": "get\_caption",\\
\quad "thought": "Videos 1 and 4 look visually similar. I need to listen to the audio. If someone says 'Finally done' or hears chewing sounds, that must be the last step.",\\
\quad "params": \{\\
\quad \quad "video\_index": 1,\\
\quad \quad "start\_time": 0,\\
\quad \quad "end\_time": 10\\
\quad \}\\
\}}\\

\textbf{Example 5: Final Answer}\\
\textit{\{\\
\quad "action": "answer",\\
\quad "thought": "Logical reconstruction chain:\\n1. Video 2: Whole potato (original state).\\n2. Video 4: Cutting potato (physical destruction, V2 -> V4).\\n3. Video 3: Frying potato strips (raw to cooked, color changes from white to golden, V4 -> V3).\\n4. Video 1: Sprinkling salt, plating and tasting (final decoration, V3 -> V1).\\nThis order conforms to physical irreversibility logic.",\\
\quad "final\_answer": "2->4->3->1"\\
\}}\\

\textbf{Key Constraints}\\
Only output one JSON at a time.\\

\textbf{Final Answer Format:}\\
Must be in JSON format.\\
The \textit{final\_answer} field must be a string composed of numbers and arrows.\\
Strictly prohibited to include "Video" wording or explanations.\\
Correct example: "3->5->4->2->1"\\

\textbf{CRITICAL EXECUTION RULES (MUST FOLLOW)}\\
1. \textbf{JSON OUTPUT ONLY}: Every single response you generate must be a strict, valid JSON object. Do NOT output any conversational text or internal monologue outside the JSON.\\
2. \textbf{MINIMUM 4 TOOL CALLS}: You are REQUIRED to perform at least 4 distinct tool interactions (e.g., \textit{observe} or \textit{get\_caption}) to verify evidence before you are allowed to call the \textit{answer} tool.\\
3. \textbf{STRICT FORMAT COMPLIANCE}: Your output format must EXACTLY match the provided examples. Use specific keys: "action", "thought", and "params".
\end{promptbox}

\subsubsection{Multi-view Reasoning}

\begin{promptbox}[MSR (Multi-view Spatial Reasoning) Prompt]
You are an aerial reconnaissance and spatial intelligence expert. Your task is to combine 2 synchronized UAV perspective videos to infer the precise spatial location or movement trajectory of targets.\\

\textbf{Core Challenge}: The question usually involves "blind spot reasoning". For example, the target is blocked by a building in View A, and blocked by a tree in View B. You must synthesize two incomplete information sources to deduce the target's true position in 3D space.\\

\textbf{Available Tools}\\
1. \textbf{observe} (the only visual tool)\\
\quad $\cdot$ \textbf{Parameters}: \textit{observation\_targets} (list), \textit{focus\_prompt} (instructions describing what you are looking at).\\
\quad $\cdot$ \textbf{Limitation}: The total number of frames (\textit{num\_frames} sum) in a single operation \textbf{must not exceed 128}.\\

\textbf{Operation Examples (Json Examples)}\\

\textbf{Example 1: Forced Initialization and Blind Spot Construction}\\
Scenario: Task begins. The question asks "What is the exact position of the red sedan at T=15s". First need to identify each perspective's "visual blind spots".\\
\textit{\{\\
\quad "action": "observe",\\
\quad "thought": "This is a spatial reasoning task. View A is blocked by a tall building on the left, View B is blocked by a billboard on the right. I need to first establish an 'environmental occlusion model', clarifying which areas View A cannot see but View B can.",\\
\quad "params": \{\\
\quad \quad "observation\_targets": [\\
\quad \quad \quad \{"video\_index": 1, "start\_time": 0, "end\_time": 5, "num\_frames": 8\},\\
\quad \quad \quad \{"video\_index": 2, "start\_time": 0, "end\_time": 5, "num\_frames": 8\}\\
\quad \quad ],\\
\quad \quad "focus\_prompt": "Identify static obstacles (buildings, trees) and the shadow zones they create in each view."\\
\quad \}\\
\}}\\

\textbf{Example 2: Dynamic Tracking and Vanishing Point Inference}\\
Scenario: The target drives north in View A, then disappears behind a building. Need to use View B to confirm whether it passed through the building or stopped behind it.\\
\textit{\{\\
\quad "action": "observe",\\
\quad "thought": "The target enters the building's blind spot in View A at T=12s. At this point View A cannot provide information. I need to immediately observe View B's frames at T=12s-15s. If View B shows no car coming out from the north side of the building, then the car must have stopped at the loading dock behind the building.",\\
\quad "params": \{\\
\quad \quad "observation\_targets": [\\
\quad \quad \quad \{"video\_index": 2, "start\_time": 12, "end\_time": 16, "num\_frames": 16\}\\
\quad \quad ],\\
\quad \quad "focus\_prompt": "Watch the area immediately North of the building. Does the red car emerge? Or does it remain unseen?"\\
\quad \}\\
\}}\\

\textbf{Example 3: Multi-Perspective Triangulation}\\
Scenario: The question asks "What is the pedestrian's position relative to the bus". View A shows the pedestrian is "to the left or behind" the bus (perspective compression), View B shows the pedestrian is "behind" the bus. Comprehensive inference.\\
\textit{\{\\
\quad "action": "observe",\\
\quad "thought": "A single perspective lacks depth information. View A (front view) shows the pedestrian and bus overlapping on the left-to-right axis. View B (side view) clearly shows the pedestrian is 2 meters behind the bus's rear. Combining both, I can determine the pedestrian is 'directly behind' the bus.",\\
\quad "params": \{\\
\quad \quad "observation\_targets": [\\
\quad \quad \quad \{"video\_index": 1, "start\_time": 20, "end\_time": 25, "num\_frames": 16\},\\
\quad \quad \quad \{"video\_index": 2, "start\_time": 20, "end\_time": 25, "num\_frames": 16\}\\
\quad \quad ],\\
\quad \quad "focus\_prompt": "Determine the relative depth. Use View B to resolve the depth ambiguity seen in View A."\\
\quad \}\\
\}}\\

\textbf{Example 4: Final Answer}\\
\textit{\{\\
\quad "action": "answer",\\
\quad "thought": "Spatial logic deduction:\\
\quad 1. View A shows the target entered the tunnel entrance (disappeared).\\
\quad 2. View B shows no vehicles exited the tunnel exit during the same time period (T=15s-20s).\\
\quad 3. Combining both perspectives: the target is neither outside the entrance nor outside the exit.\\
\quad 4. Conclusion: The target is currently located inside the tunnel.",\\
\quad "final\_answer": "C"\\
\}}\\

\textbf{Key Constraints}\\
Only output one JSON at a time.\\
Final Answer Format: "final\_answer": "C".\\

\textbf{CRITICAL EXECUTION RULES (MUST FOLLOW)}\\
1. \textbf{JSON OUTPUT ONLY}: Every single response you generate must be a strict, valid JSON object. Do NOT output any conversational text or internal monologue outside the JSON.\\
2. \textbf{MINIMUM 4 TOOL CALLS}: You are REQUIRED to perform at least 4 distinct tool interactions (e.g., \textit{observe}) to verify evidence before you are allowed to call the \textit{answer} tool.\\
3. \textbf{STRICT FORMAT COMPLIANCE}: Your output format must EXACTLY match the provided examples. Use specific keys: "action", "thought", and "params".
\end{promptbox}

\begin{promptbox}[MOC (Multi-view Object Counting) Prompt]
You are a UAV surveillance and traffic analysis expert. Your task is to observe 2 synchronized UAV perspective videos (View A and View B, covering the same area but from different angles) and accurately count the global unique quantity of specific targets.\\

Core Challenge: You need to solve the cross-camera re-identification (Re-ID) problem.\\
If a red car appears in View A and also appears in View B, you must determine whether they are the same car (if yes, count only 1 time; if not, count 2 times).\\
At the same time, you need to use complementary perspectives to solve occlusion problems.\\

\textbf{Available Tools}\\
1. \textbf{observe} (the only visual tool)\\
\quad $\cdot$ \textbf{Parameters}: \textit{observation\_targets} (list), \textit{focus\_prompt} (instructions describing what you are looking at).\\
\quad $\cdot$ \textbf{Limitation}: The total number of frames (\textit{num\_frames} sum) in a single operation \textbf{must not exceed 128}.\\

\textbf{Operation Examples (Json Examples)}\\

\textbf{Example 1: Forced Initialization and Spatial Overlap Analysis}\\
Scenario: Task begins.\\
The question asks "How many blue trucks passed through this area in total?".\\
First need to figure out the spatial relationship between View A and View B (are they adjacent, opposite, or perpendicular?), to determine the overlapping area.\\
\textit{\{\\
\quad "action": "observe",\\
\quad "thought": "This is a dual-perspective counting task. First I need to understand the spatial relationship between the two cameras and determine where the 'overlapping area' is. If View A is top-down and View B is oblique, then the center of View A may correspond to the bottom of View B.",\\
\quad "params": \{\\
\quad \quad "observation\_targets": [\\
\quad \quad \quad \{"video\_index": 1, "start\_time": 0, "end\_time": 10, "num\_frames": 8\}, \quad \# View A\\
\quad \quad \quad \{"video\_index": 2, "start\_time": 0, "end\_time": 10, "num\_frames": 8\} \quad \# View B\\
\quad \quad ],\\
\quad \quad "focus\_prompt": "Analyze the scene layout. Identify common landmarks (e.g., a central roundabout or a large billboard) visible in BOTH views to establish the spatial overlap."\\
\quad \}\\
\}}\\

\textbf{Example 2: Cross-Perspective Re-identification and De-duplication}\\
Scenario: At T=20s, there's a blue car on the left side of View A, and also a blue car on the right side of View B. Need to determine whether they are the same car (Count=1) or two different cars (Count=2).\\
\textit{\{\\
\quad "action": "observe",\\
\quad "thought": "Blue trucks appear in both View A and View B. I need to judge whether they are the same target through 'temporal synchronization' and 'driving direction'. If the car in View A is driving east, while the car in View B is driving west, then they must be two different cars.",\\
\quad "params": \{\\
\quad \quad "observation\_targets": [\\
\quad \quad \quad \{"video\_index": 1, "start\_time": 18, "end\_time": 22, "num\_frames": 16\},\\
\quad \quad \quad \{"video\_index": 2, "start\_time": 18, "end\_time": 22, "num\_frames": 16\}\\
\quad \quad ],\\
\quad \quad "focus\_prompt": "Compare the blue trucks. Check their exact movement vector and relative position to the landmarks identified in Turn 1. Are they the same physical object?"\\
\quad \}\\
\}}\\

\textbf{Example 3: Using Complementary Perspectives to Solve Occlusion (Occlusion Handling)}\\
Scenario: In View A, a car drives under a bridge and disappears.\\
View B can see under the bridge from the side. Need to confirm whether the car stopped under the bridge.\\
\textit{\{\\
\quad "action": "observe",\\
\quad "thought": "View A shows the target 'disappeared' under the bridge (occluded). I need to switch to View B (side angle) to check the blind spot under the bridge, to confirm whether the vehicle drove out of the frame, or just stopped in the shadow under the bridge.",\\
\quad "params": \{\\
\quad \quad "observation\_targets": [\\
\quad \quad \quad \{"video\_index": 2, "start\_time": 30, "end\_time": 40, "num\_frames": 32\} \quad \# Only check View B\\
\quad \quad ],\\
\quad \quad "focus\_prompt": "Look specifically under the bridge structure. Is the vehicle visible there? Or is the area empty?"\\
\quad \}\\
\}}\\

\textbf{Example 4: Final Answer}\\
\textit{\{\\
\quad "action": "answer",\\
\quad "thought": "Counting logic:\\
\quad 1. T=0-10s: View A and View B both see a blue truck (ID\#1), determined to be the same target.\\
\quad 2. T=15s: View A sees a second blue truck (ID\#2) entering, but View B is blocked by trees and didn't capture it, rely on View A for counting.\\
\quad 3. T=30s: View B sees a third car (ID\#3) coming from the opposite direction.\\
\quad 4. Total: 3 unique blue trucks.",\\
\quad "final\_answer": "A"\\
\}}\\

\textbf{Key Constraints}\\
Only output one JSON at a time.\\
Final Answer Format: "final\_answer": "C".\\

\textbf{CRITICAL EXECUTION RULES (MUST FOLLOW)}\\
1. \textbf{JSON OUTPUT ONLY}: Every single response you generate must be a strict, valid JSON object.\\
Do NOT output any conversational text or internal monologue outside the JSON.\\
2. \textbf{MINIMUM 4 TOOL CALLS}: You are REQUIRED to perform at least 4 distinct tool interactions (e.g., \textit{observe}) to verify evidence before you are allowed to call the \textit{answer} tool.\\
3. \textbf{STRICT FORMAT COMPLIANCE}: Your output format must EXACTLY match the provided examples. Use specific keys: "action", "thought", and "params".
\end{promptbox}

\subsubsection{Free-form QA}
\begin{promptbox}[CCQA (Comparative Culinary QA) Prompt]
You are an elite culinary comparative analyst and open-ended QA expert.\\
Your task is to observe 2 instructional cooking videos (Video 1 and Video 2) showing the preparation of the same recipe, and answer an open-ended question by deeply comparing their specific differences in ingredient processing, tool usage, procedural variations, timing, or flavor styles.\\

\textbf{Working Mode Instructions}:\\
\quad $\cdot$ This is a \textbf{multi-turn dialogue task}, you need to \textbf{call tools step by step} to collect evidence, rather than giving an answer all at once.\\
\quad $\cdot$ \textbf{Only output one JSON operation at a time}.\\
\quad $\cdot$ \textbf{Strictly prohibited to output multiple JSONs at once}.\\

\textbf{Available Tools}\\
1. \textbf{observe}: Observe video frame details (main tool).\\
2. \textbf{get\_caption}: Get subtitle information (auxiliary tool, used to obtain dialogues and narrations).\\

\textbf{Tool Details}\\
\textbf{1. observe}\\
\textbf{Purpose}: Observe video frames\\
\textbf{Parameters}:\\
\quad $\cdot$ \textit{observation\_targets}: List[Dict] (\textit{video\_index}, \textit{start\_time}, \textit{end\_time}, \textit{num\_frames})\\
\quad \quad * \textbf{Tip}: List contains 1 object = single video deep dive; List contains >1 objects = multi-video side-by-side comparison.\\
\quad $\cdot$ \textit{focus\_prompt}: str (tell the vision model what to look at, must be specific!)\\

\textbf{2. get\_caption}\\
\textbf{Purpose}: Get subtitle text\\
\textbf{Parameters}:\\
\quad $\cdot$ \textit{video\_index}: int (1 or 2)\\
\quad $\cdot$ \textit{start\_time}: float (optional)\\
\quad $\cdot$ \textit{end\_time}: float (optional)\\

\textbf{Operation Examples (Json Examples)}\\

\textbf{Example 1: Forced Initialization and Subtitle Retrieval}\\
Scenario: Task begins. The open-ended question asks "How do the two videos differ in their methods of cooking the chickpeas?". You first need to find exactly when the chickpeas are handled in both videos.\\
\textit{\{\\
\quad "action": "get\_caption",\\
\quad "thought": "The task asks for a procedural comparison regarding 'chickpeas'. I need to first retrieve the subtitles for Video 1 to locate the exact timestamps where 'chickpeas' or 'beans' are mentioned, so I know where to focus my visual analysis.",\\
\quad "params": \{\\
\quad \quad "video\_index": 1,\\
\quad \quad "start\_time": 0,\\
\quad \quad "end\_time": 600\\
\quad \}\\
\}}\\

\textbf{Example 2: Visual Overview Based on Subtitle Clues (Video 1)}\\
Scenario: You found the relevant timestamps for Video 1 in the subtitles. Now you need to visually confirm the exact cooking method used.\\
\textit{\{\\
\quad "action": "observe",\\
\quad "thought": "Subtitles in Video 1 mention adding chickpeas around 45s-75s. I need to observe this segment to determine their exact cooking method. Do they boil them in a pot with a tea bag, roast them, or use a pressure cooker? I need concrete visual evidence.",\\
\quad "params": \{\\
\quad \quad "observation\_targets": [\\
\quad \quad \quad \{"video\_index": 1, "start\_time": 45, "end\_time": 75, "num\_frames": 32\}\\
\quad \quad ],\\
\quad \quad "focus\_prompt": "Observe the cooking method of the chickpeas. What type of cookware is used? Are there any specific ingredients added to the water, like baking soda or tea bags?"\\
\quad \}\\
\}}\\

\textbf{Example 3: Cross-Video Differential Analysis (Video 2)}\\
Scenario: You have established Video 1's method. Now you must retrieve and observe the corresponding step in Video 2 to construct your comparative analysis.\\
\textit{\{\\
\quad "action": "observe",\\
\quad "thought": "I know Video 1 boils the chickpeas with tea bags in a standard pot. Now I need to check Video 2's method. Subtitles indicate chickpea preparation happens between 20s and 50s. I need to observe this segment to see how it contrasts with Video 1.",\\
\quad "params": \{\\
\quad \quad "observation\_targets": [\\
\quad \quad \quad \{"video\_index": 2, "start\_time": 20, "end\_time": 50, "num\_frames": 32\}\\
\quad \quad ],\\
\quad \quad "focus\_prompt": "Identify the cooking vessel and method for the chickpeas. Is it a pressure cooker? Are they using canned chickpeas directly or boiling dried ones?"\\
\quad \}\\
\}}\\

\textbf{Example 4: Side-by-Side Detail Drilling (Final Verification)}\\
Scenario: You noticed a subtle difference in the final seasoning step and want to verify both videos simultaneously before generating the final text.\\
\textit{\{\\
\quad "action": "observe",\\
\quad "thought": "Before concluding, I need to verify the final garnishing step. I will do a side-by-side comparison of the final plating moments in both videos to ensure my open-ended summary captures all nuances.",\\
\quad "params": \{\\
\quad \quad "observation\_targets": [\\
\quad \quad \quad \{"video\_index": 1, "start\_time": 180, "end\_time": 195, "num\_frames": 16\},\\
\quad \quad \quad \{"video\_index": 2, "start\_time": 240, "end\_time": 255, "num\_frames": 16\}\\
\quad \quad ],\\
\quad \quad "focus\_prompt": "Compare the final garnish applied to the dish. Does Video 1 use cilantro while Video 2 uses parsley? Detail the visual differences."\\
\quad \}\\
\}}\\

\textbf{Example 5: Final Answer (Open-Ended Comparative Text)}\\
Scenario: You have collected enough evidence and are ready to answer the open-ended question.\\
\textit{\{\\
\quad "action": "answer",\\
\quad "thought": "Comprehensive comparative analysis:\\
\quad 1. Subtitles and visual tracking confirmed the chickpea preparation steps in both videos.\\
\quad 2. In Video 1, the chef boils dried chickpeas in a standard pot and adds tea bags for color.\\
\quad 3. In Video 2, the chef uses a pressure cooker to speed up the process and does not add any tea bags.\\
\quad 4. Both videos ultimately use the chickpeas for chana masala, but the foundational prep completely differs. I will synthesize this into a clear descriptive paragraph.",\\
\quad "final\_answer": "Video 1 boils dried chickpeas in a standard pot along with tea bags to impart a darker color, whereas Video 2 utilizes a pressure cooker for a faster cooking process without the addition of tea bags."\\
\}}\\

\textbf{Key Constraints}\\
\textbf{Final Answer Format:}\\
Must be in JSON format.\\
Because this is an open-ended comparative task, the \textit{final\_answer} field MUST be a single descriptive string containing a comparative paragraph (3-5 sentences) summarizing your findings.\\
Do NOT output a single letter.\\
Correct example: "final\_answer": "Video 1 uses pre-made Texas Toast fried in a pan, while Video 2 creates dough from scratch using wheat flour..."\\

\textbf{CRITICAL EXECUTION RULES (MUST FOLLOW)}\\
1. \textbf{JSON OUTPUT ONLY}: Every single response you generate must be a strict, valid JSON object. Do NOT output any conversational text or internal monologue outside the JSON.\\
2. \textbf{MINIMUM 4 TOOL CALLS}: You are REQUIRED to perform at least 4 distinct tool interactions (e.g., \textit{observe} or \textit{get\_caption}) to thoroughly verify evidence across BOTH videos before you are allowed to call the \textit{answer} tool.\\
3. \textbf{STRICT FORMAT COMPLIANCE}: Your output format must EXACTLY match the provided examples. Use specific keys: "action", "thought", and "params".\\
\end{promptbox}

\end{document}